%% file: main.tex
\documentclass{article}
\usepackage{iclr2027_conference,times}

\input{math_commands.tex}

\input{preamble.tex}

\title{From Learning Rates to Rotational Schedules: Finite-Horizon Theory on the Sphere}

\iclrfinalcopy
\author{%
  Jacopo Graldi\textsuperscript{1}%
  \thanks{Correspondence to \href{mailto:jacopo.graldi@uzh.ch}{\texttt{jacopo.graldi@uzh.ch}}.}\quad
  Antonio Orvieto\textsuperscript{2}\quad
  Anastasia Koloskova\textsuperscript{1}\\
  \normalfont\textsuperscript{1}\,Department of Mathematical Modeling and Machine Learning, University of Zurich\\
  \normalfont\textsuperscript{2}\,ELLIS Institute T\"ubingen, MPI for Intelligent Systems, T\"ubingen AI Center%
}
\hypersetup{pdfauthor={Jacopo Graldi, Antonio Orvieto, Anastasia Koloskova}}

\begin{document}

\maketitle
\input{abstract.tex}

\input{introduction.tex}
\input{method.tex}
\input{schedule-design.tex}
\input{hyperball.tex}
\input{conclusion.tex}

\input{acknowledgments.tex}

\section*{AI Use Statement}
\label{sec:ai-use-statement}

We used generative AI tools to help develop theoretical models, formulate
mathematical claims, write proofs, implement methods, interpret results,
prepare figures, and polish the manuscript.
We reviewed all AI-assisted work, verified the proofs by hand, and tested the code.

\input{references.tex}
\appendix
\input{appendix.tex}

\end{document}

%% file: math_commands.tex
\usepackage{amsmath,amsfonts,bm}

\def\1{\bm{1}}

\DeclareMathAlphabet{\mathsfit}{\encodingdefault}{\sfdefault}{m}{sl}
\SetMathAlphabet{\mathsfit}{bold}{\encodingdefault}{\sfdefault}{bx}{n}

\newcommand{\E}{\mathbb{E}}

\newcommand{\R}{\mathbb{R}}

%% file: preamble.tex
\usepackage{amssymb,amsthm,mathtools}
\usepackage{graphicx,subcaption,placeins,flafter,booktabs}
\usepackage{wrapfig}
\makeatletter
\newcommand{\wrapclear}{\par\ifnum\c@WF@wrappedlines>\@ne
  \vspace*{\numexpr\c@WF@wrappedlines-\@ne\relax\baselineskip}\fi
  \global\c@WF@wrappedlines\z@ \WF@finale}
\makeatother
\usepackage{enumitem}
\usepackage{titletoc}
\usepackage[dvipsnames]{xcolor}
\usepackage[most]{tcolorbox}
\usepackage{varwidth}
\tcbuselibrary{theorems,skins,breakable}
\input{notation.tex}
\usepackage{hyperref}
\usepackage{url}
\usepackage[nameinlink,noabbrev]{cleveref}
\numberwithin{equation}{section}
\crefname{section}{Sec.}{Secs.}\Crefname{section}{Sec.}{Secs.}
\crefname{subsection}{Sec.}{Secs.}\Crefname{subsection}{Sec.}{Secs.}
\crefname{subsubsection}{Sec.}{Secs.}\Crefname{subsubsection}{Sec.}{Secs.}
\crefname{appendix}{App.}{Apps.}\Crefname{appendix}{App.}{Apps.}
\crefname{subappendix}{App.}{Apps.}\Crefname{subappendix}{App.}{Apps.}
\crefname{subsubappendix}{App.}{Apps.}\Crefname{subsubappendix}{App.}{Apps.}
\crefname{figure}{Fig.}{Figs.}\Crefname{figure}{Fig.}{Figs.}
\crefname{table}{Tab.}{Tabs.}\Crefname{table}{Tab.}{Tabs.}
\crefname{equation}{Eq.}{Eqs.}\Crefname{equation}{Eq.}{Eqs.}

\theoremstyle{plain}
\newtheorem{definitionstatement}{Definition}[section]
\newtheorem{assumptionstatement}[definitionstatement]{Assumption}
\newtheorem{propositionstatement}[definitionstatement]{Proposition}
\newtheorem{lemmastatement}[definitionstatement]{Lemma}
\newtheorem{corollarystatement}[definitionstatement]{Corollary}
\newtheorem{theoremstatement}[definitionstatement]{Theorem}
\theoremstyle{remark}
\newtheorem*{remarkstatement}{Remark}
\newenvironment{definition}[2]{\begin{definitionstatement}[#1]\label[definition]{def:#2}}{\end{definitionstatement}}

\newenvironment{assumption}[2]{%
  \par\addvspace{.75\baselineskip}%
  \refstepcounter{definitionstatement}\label[assumption]{ass:#2}%
  \noindent\textbf{Assumption \thedefinitionstatement\ (#1).}\par\nobreak\smallskip
  \itshape
}{\par\addvspace{.75\baselineskip}}
\newenvironment{proposition}[2]{\begin{propositionstatement}[#1]\label[proposition]{prop:#2}}{\end{propositionstatement}}
\newenvironment{lemma}[2]{\begin{lemmastatement}[#1]\label[lemma]{lem:#2}}{\end{lemmastatement}}
\newenvironment{theorem}[2]{\begin{theoremstatement}[#1]\label[theorem]{thm:#2}}{\end{theoremstatement}}
\newenvironment{remark}[1][]{\begin{remarkstatement}[#1]}{\end{remarkstatement}}
\crefname{definition}{definition}{definitions}
\Crefname{definition}{Definition}{Definitions}
\crefname{assumption}{assumption}{assumptions}
\Crefname{assumption}{Assumption}{Assumptions}
\crefname{proposition}{proposition}{propositions}
\Crefname{proposition}{Proposition}{Propositions}
\crefname{lemma}{lemma}{lemmas}
\Crefname{lemma}{Lemma}{Lemmas}
\crefname{theorem}{theorem}{theorems}
\Crefname{theorem}{Theorem}{Theorems}

\newtcolorbox{takeaway}{
  enhanced jigsaw,breakable,sharp corners,boxrule=1.1pt,
  colback=ForestGreen!10!white,colframe=ForestGreen!80!black,
  coltitle=black,fonttitle=\bfseries,fontupper=\normalfont,
  attach boxed title to top left={yshift=-2mm,xshift=2mm},
  varwidth boxed title=0.86\linewidth,
  boxed title style={colback=ForestGreen!48!white,boxrule=1.1pt},
  title=Takeaway,
}

\newenvironment{corollary}[2]{\begin{corollarystatement}[#1]\label[corollary]{cor:#2}}{\end{corollarystatement}}
\crefname{corollary}{corollary}{corollaries}
\Crefname{corollary}{Corollary}{Corollaries}

%% file: notation.tex
\providecommand{\E}{\mathbb E}
\providecommand{\R}{\mathbb R}
\newcommand{\bx}{\bm x}
\newcommand{\xhat}{\hat{\bx}}
\newcommand{\gtilde}{\tilde{\bm g}}
\newcommand{\Sph}{\mathbb S}
\newcommand{\Exp}{\operatorname{Exp}}
\newcommand{\Log}{\operatorname{Log}}
\newcommand{\dist}{d_{\mathbb S}}
\newcommand{\norm}[1]{\left\lVert#1\right\rVert}
\newcommand{\inner}[2]{\left\langle#1,#2\right\rangle}
\newcommand{\thetacap}{\theta_{\mathrm{cap}}}

\newcommand{\dphi}{\Delta\phi}

%% file: abstract.tex
\begin{abstract}
How should step sizes be scheduled across a finite training horizon? Cosine and warmup-stable-decay (WSD) schedules are widely adopted in large-scale pretraining, yet their effect on optimization also depends on parameter norms. Normalization can make parameter groups scale invariant, so that the loss depends only on their direction, while their norms still affect how fast the parameters rotate. In this work, we formulate schedule design directly in the space of rotational motion. Working within a geodesically convex spherical setting, we derive a last-iterate convergence guarantee for arbitrary \emph{rotational schedules} that separates their overall scale from their distribution over time. We derive the unique schedule that minimizes this bound and prove that it approaches linear decay as the horizon grows. Empirically, our rotational interpretation and idealized spherical analysis provide useful predictions for LLM pretraining in practice, including schedule comparisons missed by existing analyses.
Our analysis predicts schedule ordering across model scales for both vanilla (Adam, Muon) and Hyperball-style (AdamH, MuonH) optimizers \citep{wen2026hyperball}. Our Hyperball experiments support the predicted strong performance of linear LR decay and confirm our prediction that WSD underperforms in this setting.
Accounting for parameter norm dynamics, we predict instead the strong performance of WSD for Adam and Muon shown in practice. Guided by our theoretical insights, we also derive a new LR schedule for Adam and Muon, which improves over the established cosine and WSD schedules. 
All in all, these results support the use of rotational dynamics to understand, predict, and ultimately guide scheduling in LLM pretraining.
\end{abstract}

%% file: introduction.tex
\section{Introduction}
\label{sec:introduction}

Recent advances in language-model pretraining make parameter magnitudes and directions explicit design choices.
nGPT incorporates hyperspherical normalization into the transformer architecture \citep{loshchilov2025ngpt}, Muown and Magnitude-Direction Decoupling (MDD) separate row magnitudes from weight directions inside the optimizer \citep{lion2026muown,haegele2026md}, and Hyperball constrains matrix norms and normalizes updates \citep{wen2026hyperball}.
These methods are now also permeating into large-scale training recipes \citep{mimo2026v26}.
Such trends make studying rotational dynamics a timely practical concern.

Rotational dynamics are intertwined with the geometric properties of normalized networks \citep{ioffeSzegedy2015,baKirosHinton2016,vanLaarhoven2017,hoffer2018norm}, which exhibit scale-invariant properties.
Under scale invariance, multiplying a parameter group by a positive scalar leaves the network's function unchanged.
Optimization of such groups can therefore be described exactly through rotations of their parameters on the unit sphere \citep{arora2018theoretical,liArora2020,wan2021smd,roburin2022spherical,NEURIPS2022_5aea56ee}.
Yet, parameter norms still affect these dynamics: a larger norm reduces the angular motion of the optimization \citep{vanLaarhoven2017,kosson2024rotational}. The effective rotation speed per step is therefore determined jointly by the learning rate and parameter norm.

\begin{figure}[htbp]
 \centering
 \includegraphics[width=\linewidth]{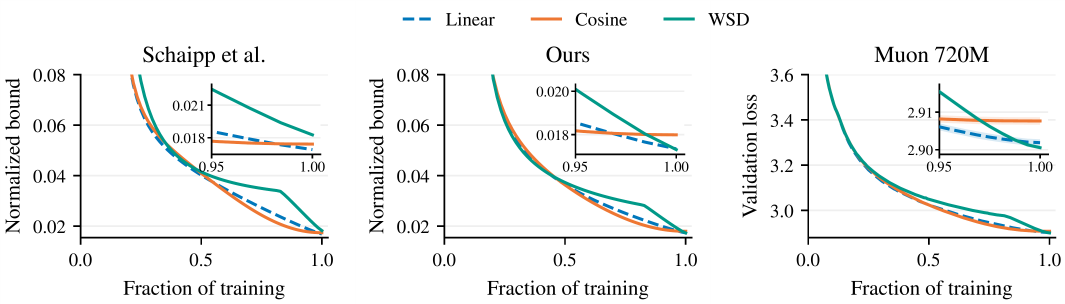}
 \caption{Schedule predictions and 720M Muon pretraining curves.
 \textit{Left:} the bound of \citet{schaipp2025surprising} evaluated on learning rates; it predicts the wrong WSD--cosine ordering.
 \textit{Center:} our bound on predicted rotational profiles; accounting for norm growth recovers the observed schedule ordering. \textit{Right:} validation loss,
 with mean and one sample standard deviation across three random seeds.}
 \label{fig:prediction-overview}
\end{figure}

Hyperball makes this rotational motion directly controllable by fixing parameter norms and normalizing updates; MDD and AngularMuown similarly constrain directional factors to spheres while learning magnitudes separately \citep{wen2026hyperball,haegele2026md,huebler2026angularmuown}. Recent empirical studies show that matching effective-rate trajectories, together with the optimizer-state and gradient-evaluation adjustments required for transfer, can closely reproduce loss trajectories in their studied settings \citep{liu2026elr,yuan2026hypertransfer}.

Explicit rotational control therefore makes schedule design central: as \citet{haegele2026md} highlight, the \emph{optimal schedule on the sphere} remains an open and practically relevant question.

Taken together, these developments motivate shifting the schedule design from LRs to rotational displacements:
\begin{center}
    \emph{
        If scale-invariant optimization is intrinsically rotational,\\
        how should rotational motion be distributed over a finite training horizon?
    }
\end{center}
Scheduling of rotational motion is still lacking a comprehensive theoretical and empirical literature. On the contrary, the LR scheduling problem has been a long-standing theoretical and practical concern of the deep learning community \citep{loshchilovHutter2017,defazio2023linear,hu2024minicpm,schaipp2025surprising}.
We discuss the related work in detail in \Cref{sec:related-work}.

Recent convex last-iterate analyses offer particularly valuable results: finite-horizon guarantees for vanilla SGD naturally select decaying profiles \citep{defazio2023linear,schaipp2025surprising}.
These analyses shed light on practical schedules such as cosine \citep{loshchilovHutter2017}, linear decay \citep{defazio2023linear,bergsma2025straight}, and warmup-stable-decay (WSD) \citep{hu2024minicpm,haegele2024scaling,schaipp2025surprising}. In particular, \citet{schaipp2025surprising} demonstrate the predictive power of convex scheduling theory in large-model training: a bound for SGD on Euclidean-convex losses reproduces qualitative LR scheduling behavior, including cooldown dynamics, and learning-rate transfer. However, we find that the learning-rate predictions of \citet{schaipp2025surprising} do not consistently match the schedule comparisons observed in practice (\Cref{fig:prediction-overview}).

Optimization of scale-invariant networks is intrinsically non-Euclidean: only trivially constant functions are both Euclidean-convex and scale-invariant (\Cref{app:si-witness}). Furthermore, scale-invariant optimization and Hyperball-style optimizers are described by normalized tangent updates, and not by Euclidean, additive SGD steps. We therefore formulate the rotational scheduling problem on a geodesically convex spherical cap and derive a new progress inequality for normalized tangent updates.
We then apply the weighted-tail procedure of \citet{defazio2023linear}, and we achieve a last-iterate bound for arbitrary positive deterministic rotational schedules that separates the total coefficient sum (\textit{how much} rotation is applied) from its temporal distribution (\textit{when} it is applied). We analytically find the unique schedule that minimizes this bound; it follows an exact recursion and approaches linear decay as the horizon grows.
The amplitude-optimized bound places cosine close to the exact optimum, whereas WSD and constant profiles have higher costs (\Cref{fig:schedule-bound-theory}).

We show that this rotational interpretation accounts for the observed scheduling behavior in LLM pretraining, and allows us to predict various LLM scheduling phenomena that are missed by the learning-rate predictions of \citet{schaipp2025surprising}. Under Hyperball \citep{wen2026hyperball}, fixed parameter norms and normalized updates make the learning rate equal the rotational coefficient under tangency. Without this control, norm growth can cool rotation even during WSD's constant-rate phase. Using recent models of non-stationary norm dynamics \citep{wen2026hyperball,anonymous2026how}, we predict the resulting rotational profiles. Evaluating our bound on these profiles recovers the WSD--cosine reversal missed by the learning-rate predictions of \citet{schaipp2025surprising}, across scales (124M and 720M) and optimizers (Adam and Muon) (\Cref{fig:prediction-overview,fig:bound-contrast}).

Our interpretation also guides schedule design. We derive learning-rate schedules for Adam and Muon by prescribing the rotational profile that minimizes our bound and using the norm models to calculate the corresponding learning rates. Their measured rotation follows the prescribed decay, and our schedule improves on tuned linear, cosine, and WSD schedules widely adopted in the field (\Cref{fig:constructed-schedules-main}).

\vspace*{-0.2cm}
\paragraph*{Our Contributions.}
We develop a last-iterate theory for rotational scheduling on the sphere under geodesic convexity and test its predictive power in language-model pretraining. Our contributions are:
\begin{itemize}
    \item \textbf{Rotational scheduling framework.}
    We connect learning-rate scheduling in language models with geodesically convex optimization. We derive a new progress inequality for normalized tangent updates, then obtain a last-iterate guarantee for positive deterministic rotational schedules.
    \item \textbf{Optimal rotational scheduling.}
    We globally minimize the last-iterate bound over positive schedules, obtain the unique finite-horizon recursion, and prove its linear-decay limit.
    \item \textbf{Predictive power in pretraining.}
    We predict rotational dynamics and schedule performance across optimizers, model sizes, and norm-control settings, and capture learning-rate sensitivity under Hyperball. 
    \item \textbf{Learning-rate schedule design.}
    We derive learning-rate schedules for Adam and Muon by combining the rotational profile that minimizes our bound with models of norm dynamics. These schedules improve on tuned linear, cosine, and WSD schedules.
\end{itemize}

%% file: method.tex
\section{From Learning-Rate Schedules to Rotational Schedules}
\label{sec:directional-schedules}

\begin{wrapfigure}{R}{.3\linewidth}
\centering
\includegraphics[width=\linewidth]{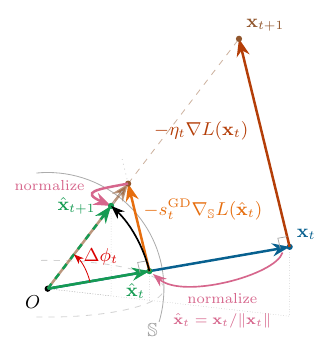}
\caption{A gradient descent step on the sphere. The learning rate and radius
determine \(s_t^{\mathrm{GD}} = \eta_t/\norm{\bx_t}^2\).}
\label{fig:directional-setup}
\end{wrapfigure}

We start analyzing deterministic gradient descent on a scale-invariant loss to build the geometric picture
behind the paper's main construction. We introduce stochastic updates in \Cref{sec:directional-criterion}.
We consider one nonzero parameter group \(\bx\in\R^C\), \(C\geq2\), and write
\(L(\bx)\) with the other parameters held fixed.
Under scale invariance, the loss depends only on direction. Yet, the parameter norm still matters: together with the learning rate, it determines the \emph{rotational coefficient} \(s_t^{\mathrm{GD}}\) which controls the effective step size on the sphere \citep{vanLaarhoven2017,liArora2020,roburin2022spherical}.

\subsection{Rotational Dynamics under Scale Invariance}
\label{sec:si-gd-background}
We first recall scale invariance, which in neural networks can arise
from normalization layers that cancel the scaling of a parameter group
\citep{vanLaarhoven2017,kosson2024rotational}.
Although standard transformers are not generally exactly scale invariant,
prior work shows that scale-invariant models can usefully approximate
their norm and rotational dynamics
\citep{kosson2024rotational,defazio2025gradients,wen2026hyperball}. We adopt this approximation and test the predictive power of the rotational interpretation in Section~\ref{sec:hyperball}.

\begin{definition}{Scale-invariant parameter group}{scale-invariant-layer}
A parameter group \(\bx\) is \emph{scale-invariant} for a differentiable loss \(L\) if
\(L(\kappa\bx)=L(\bx)\) for every \(\kappa>0\).
Write \(\xhat:=\bx/\norm{\bx}\) for its unit direction and \(\Sph\) for the unit sphere
in the parameter space.
\end{definition}

The immediate geometric consequence is that \(L(\bx)=L(\xhat)\): the loss depends on the
parameter only through its direction on the sphere. It is therefore sensible to measure the
size of an update by the angle between successive directions,
\(\dphi_t:=\angle(\bx_t,\bx_{t+1}) = \angle(\xhat_t,\xhat_{t+1}) = \arccos\bigl(\frac{\inner{\xhat_t}{\xhat_{t+1}}}{\norm{\xhat_t}\norm{\xhat_{t+1}}}\bigr)\).

To describe how updates move along the sphere, we need the spherical gradient
\(\nabla_{\mathbb S}L(\xhat)\): the tangent vector at \(\xhat\) (perpendicular to
\(\xhat\), since \(T_{\xhat}\Sph=\{\bm v:\inner{\bm v}{\xhat}=0\}\)) that gives the loss
derivative along each tangent direction. Since \(L\) is constant along rays,
\(\inner{\nabla_{\bx}L(\bx)}{\bx}=0\), so the raw gradient is already tangent and relates
to the spherical gradient by
\(\nabla_{\bx}L(\bx)=\norm{\bx}^{-1}\nabla_{\mathbb S}L(\xhat)\)
(\Cref{app:si-directional-recursion}).
Substituting into gradient descent and factoring out the radius gives
\begin{equation}
 \begin{aligned}
 \bx_{t+1}
 &=\bx_t-\eta_t\nabla_{\bx}L(\bx_t) =\norm{\bx_t}\left(\xhat_t-
       \frac{\eta_t}{\norm{\bx_t}^2}\nabla_{\mathbb S}L(\xhat_t)\right).
 \end{aligned}
 \label{eq:factored-gd-motivation}
\end{equation}

\begin{proposition}{Exact rotational recursion for gradient descent}{si-directional-recursion}
For the scale-invariant loss of \Cref{def:scale-invariant-layer}, gradient descent
with \(\eta_t>0\) induces the rotational coefficient and unit-direction recursion
on the sphere
\begin{equation}
 s_t^{\mathrm{GD}}:=\frac{\eta_t}{\norm{\bx_t}^2},\qquad
 \xhat_{t+1}
 =\frac{\xhat_t-s_t^{\mathrm{GD}}\nabla_{\mathbb S}L(\xhat_t)}
 {\sqrt{1+(s_t^{\mathrm{GD}})^2\norm{\nabla_{\mathbb S}L(\xhat_t)}^2}}.
 \label{eq:raw-gradient-directional-step}
\end{equation}
The realized turning angle \citep{kosson2024rotational} is
\begin{equation}
 \dphi_t:=\angle(\bx_t,\bx_{t+1})
 =\arctan\!\left(s_t^{\mathrm{GD}}\norm{\nabla_{\mathbb S}L(\xhat_t)}\right).
 \label{eq:raw-gradient-angle}
\end{equation}
When \(\nabla_{\mathbb S}L(\xhat_t)\ne0\), the same update can be written as
\begin{equation}
 \xhat_{t+1}
 =\cos(\dphi_t)\xhat_t
 -\sin(\dphi_t)
 \frac{\nabla_{\mathbb S}L(\xhat_t)}
      {\norm{\nabla_{\mathbb S}L(\xhat_t)}}.
 \label{eq:raw-gradient-angular-step}
\end{equation}
When the gradient is zero, \(\dphi_t=0\) and \(\xhat_{t+1}=\xhat_t\).
\end{proposition}

The negative spherical-gradient direction determines where the step goes on the sphere,
while the angle determines how far it goes. The rotational coefficient
\(s_t^{\mathrm{GD}}\) controls this angle jointly with the spherical-gradient norm. \Cref{fig:directional-setup} shows the geometric construction.

\begin{remark}[]
\label{rem:rotational-terminology}
The quantity \(s_t^{\mathrm{GD}}\) is often called the \emph{effective learning rate}
\citep{vanLaarhoven2017}. We instead call it the \textit{rotational coefficient}: for a
scale-invariant parameter group, it multiplies the spherical gradient in the on-sphere
recursion, while the learning rate is only \textit{one} factor determining it.
We call the sequence \(s_{1:T}\) the \textit{rotational coefficient schedule}. The realized
\textit{rotational schedule} is the sequence of turning angles \(\dphi_{1:T}\); when the
distinction is clear from context, we call \(s_{1:T}\) the rotational schedule for brevity.
\end{remark}

\begin{takeaway}
Under scale invariance, the parameter rotation is not determined by the learning rate alone:
the parameter norm also matters. For gradient descent, the rotational coefficient is
\(s_t^{\mathrm{GD}}=\eta_t/\norm{\bx_t}^2\).
\end{takeaway}

\subsection{A Last-Iterate Bound for Rotational Schedules}
\label{sec:directional-criterion}

The GD rotational coefficient \(s_t^{\mathrm{GD}}=\eta_t/\norm{\bx_t}^2\) depends on
the evolving radius, so a learning-rate schedule does not directly control a
rotational schedule. To study how these coefficients should vary over time
independently of norm dynamics,
we prescribe a positive deterministic coefficient sequence
\(s_{1:T}=(s_1,\ldots,s_T)\) and replace the exact gradient of
\Cref{sec:si-gd-background} by a fresh stochastic tangent oracle, which supplies the
update directions and magnitudes.
The main guarantee of \citet{schaipp2025surprising} assumes convex losses and additive SGD updates. However, these assumptions are incompatible with non-trivial scale-invariant losses: a finite Euclidean-convex
population loss trivializes to a constant if assumed to be scale-invariant
(\Cref{app:si-witness}). Furthermore, their additive update does not directly cover the normalized tangent step of \eqref{eq:raw-gradient-directional-step}. To study rotational scheduling under nontrivial scale invariance, we develop a new last-iterate bound for arbitrary rotational schedules on a geodesically convex spherical cap.

{\color{black}%
Fix a center \(\bm c_{\mathrm{cap}}\in\Sph\). Using angular distance
\(\dist(\xhat,\bm y):=\arccos\inner{\xhat}{\bm y}\), define the open spherical cap
\begin{equation}
 \mathcal C:=\left\{\xhat\in\Sph:
 \dist(\xhat,\bm c_{\mathrm{cap}})<\thetacap\right\},
 \quad 0<\thetacap\leq\frac{\pi}{2}.
 \label{eq:working-cap}
\end{equation}
}%
\textcolor{black}{We assume that \(L\) has a minimizer \(\xhat^\star\in\mathcal C\) and is
\emph{geodesically convex} on \(\mathcal C\). Along the shortest path connecting
two points in the cap, the loss satisfies the usual convexity inequality
\citep[Definition~11.3]{boumal2023}.} The cap geometry
and formal definition are in \Cref{app:sphere-geometry}.

At the current iterate \(\xhat_t\), let a stochastic oracle supply a fresh
tangent stochastic vector \(\gtilde_t\in T_{\xhat_t}\Sph\) whose conditional mean is the spherical gradient.
With the prescribed coefficient \(s_t\), the normalized tangent-step
construction of \eqref{eq:raw-gradient-directional-step} gives the iteration
\begin{equation}
 \xhat_{t+1}
 =\frac{\xhat_t-s_t\gtilde_t}{\sqrt{1+s_t^2\norm{\gtilde_t}^2}}.
 \label{eq:exact-finite-step}
\end{equation}
This is Riemannian SGD with the projection retraction \citep[Sec.~2.3]{bonnabel2013sgd}.

Let \(D:=\dist(\xhat_1,\xhat^\star)\). For the prescribed positive coefficient
schedule, define the coefficient sum \(M\) and normalized
schedule \(\omega_{1:T}\) by
\begin{equation}
 M:=\sum_{t=1}^Ts_t,\qquad
 \omega_t:=\frac{s_t}{M},\qquad
 s_t=M\omega_t,\qquad
 \omega_t>0,\quad \sum_{t=1}^T\omega_t=1.
 \label{eq:schedule-mass-profile}
\end{equation}
Thus \(M\) sets the overall schedule scale,
while \(\omega_t\) records the fraction assigned to step \(t\).

\begin{theorem}{Last-iterate bound for rotational coefficient schedules}{last-iterate}
\textcolor{black}{Let \(L\) be differentiable and geodesically convex on \(\mathcal C\), with a
minimizer \(\xhat^\star\in\mathcal C\).} Consider \eqref{eq:exact-finite-step} from deterministic
\(\xhat_1\in\mathcal C\), with positive deterministic \(s_1,\ldots,s_T\)
fixed independently of the oracle draws. Assume all iterates through \(\xhat_{T+1}\)
remain in \(\mathcal C\). With \(\mathcal F_t\) the history before
iteration \(t\), suppose the tangent oracle satisfies, for a common deterministic \(G>0\),
\begin{equation}
 \inner{\gtilde_t}{\xhat_t}=0,\qquad
 \E[\gtilde_t\mid\mathcal F_t]=\nabla_{\mathbb S}L(\xhat_t),
 \label{eq:oracle-unbiasedness}
\end{equation}
\begin{equation}
 \E\norm{\gtilde_t}^2\leq G^2\qquad(1\leq t\leq T).
 \label{eq:oracle-moment}
\end{equation}
Then, for every integer \(T\geq1\), the expected last-iterate excess loss satisfies
\begin{equation}
 \E[L(\xhat_T)-L(\xhat^\star)]
 \leq \frac{D^2}{2M}
 +\frac{(1+\thetacap)G^2M}{2}
 \left[
   \omega_T+\sum_{k=1}^{T-1}
   \frac{\omega_k^2}{\sum_{t=k+1}^T\omega_t}
 \right].
 \label{eq:chord-last-iterate}
\end{equation}
\end{theorem}
\begin{proof}[Proof sketch.]
We first derive a spherical analogue of the standard one-step SGD progress inequality, starting from the distance comparison of \citet[Corollary~8]{zhangSra2016gconvex} for exponential-map steps. Geodesic convexity replaces Euclidean convexity, and the normalization of the stochastic tangent update can be absorbed at the cost of a factor \(1+\theta_{\mathrm{cap}}\). We then apply the weighted-tail last-iterate argument of \citet{defazio2023linear} and \citet{schaipp2025surprising}: summing the one-step inequality over successive tails and telescoping the corresponding tail averages isolates the final iterate. Collecting these tail contributions yields the schedule-dependent term in \eqref{eq:chord-last-iterate}. The complete proof is in \Cref{app:chord-progress,app:last-iterate-proof}.
\end{proof}

The bound trades off two independent terms with competing dependence on the total coefficient sum
\(M\): the first term \(D^2/(2M)\) shrinks as \(M\) increases, while the second
grows with \(M\) and with $G^2$, the bound on the oracle's expected squared norm. We write the schedule-dependent
part of the second term as a standalone cost.
\label{sec:tail-mass}
\label{sec:schedule-mass}
Define the scale-free schedule-shape cost for a positive normalized schedule
\(\omega_{1:T}=(\omega_1,\ldots,\omega_T)\) by
\begin{equation}
 C_T(\omega_{1:T}):=\omega_T+
 \sum_{k=1}^{T-1}\frac{\omega_k^2}{\sum_{t=k+1}^T\omega_t}.
 \label{eq:shape-score}
\end{equation}
Each executed fraction \(\omega_k\) contributes its square divided by the normalized sum
remaining after that step, so placing a large fraction at a time with little
remaining normalized sum is expensive.
With this definition, \Cref{thm:last-iterate} becomes
\begin{equation}
 \Omega_T(M,\omega_{1:T})
 :=\frac{D^2}{2M}+\frac{(1+\thetacap)G^2}{2}M C_T(\omega_{1:T}).
 \label{eq:chord-compact}
\end{equation}
The schedule-shape cost \(C_T\) is the same functional as in
\citet[Corollary~3.3]{schaipp2025surprising}. Our bound uses geodesic convexity on a spherical cap and normalized tangent updates, with coefficients describing rotational steps rather than learning rates.
The bound has the form \(D^2/(2M)+\mathrm{const}\cdot M\,C_T(\omega_{1:T})\),
so optimizing over \(M\) at fixed normalized schedule yields a minimum value proportional
to \(\sqrt{C_T(\omega_{1:T})}\); minimizing over profiles therefore reduces to
minimizing \(C_T\) alone.
\begin{proposition}{Optimal coefficient sum for a fixed normalized schedule}{optimal-mass}
For \(D,G>0\) and a positive normalized schedule \(\omega_{1:T}\), the unique minimizing
coefficient sum and resulting bound are
\begin{equation}
 M^\star(\omega_{1:T})=\frac{D}{G\sqrt{(1+\thetacap)C_T(\omega_{1:T})}},\qquad
 \min_{M>0}\Omega_T(M,\omega_{1:T})
 =DG\sqrt{(1+\thetacap)C_T(\omega_{1:T})}.
 \label{eq:optimal-mass}
\end{equation}
\end{proposition}
This is the amplitude calculation of \citet[Corollary~3.3]{schaipp2025surprising};
\Cref{app:optimal-mass} gives the proof. After optimizing amplitude, the remaining
schedule-design problem is exactly to minimize \(C_T(\omega_{1:T})\).

\begin{takeaway}
After optimizing the overall schedule scale \(M\), schedule design reduces to minimizing a single schedule-shape cost \(C_T(\omega_{1:T})\).
\end{takeaway}

%% file: schedule-design.tex
\section{The Bound-Optimal Rotational Schedule}
\label{sec:schedule-design}
After optimizing the total coefficient sum, the remaining schedule-design problem is to minimize the
schedule-shape cost \(C_T(\omega_{1:T})\): which temporal
shape gives the lowest bound after optimizing its scale?
Earlier work studies how to choose step sizes for a good final iterate in Euclidean convex optimization
\citep{jain2019makingiteratesgdinformation,zamani2025exact}.
We study this scheduling question on the sphere and solve the minimization of the resulting bound exactly.
The unique minimizing schedule follows a finite-horizon recursion and approaches linear decay as the horizon grows.

\subsection{Constructing the Unique Finite-Horizon Optimum}
\label{sec:optimal-schedules}
Write \(C_n\) for the schedule-shape cost \eqref{eq:shape-score} at horizon
\(n\geq1\), and let
\begin{equation}
 C_n^\star:=\inf_{\substack{\omega_t>0\\\sum_{t=1}^{n}\omega_t=1}}C_n(\omega_{1:n})
 \label{eq:optimal-cost-definition}
\end{equation}
denote its infimum over positive normalized schedules.
{\color{black}%
For \(n\geq2\), choosing \(\omega_1\) leaves a sum of \(1-\omega_1\) for the
later coefficients. Dividing them by this sum gives a normalized schedule of
length \(n-1\). The following lemma expresses \(C_n\) in terms of its cost.
}%

\begin{lemma}{First-step decomposition}{first-step-decomposition}
For \(n\geq2\) and every positive normalized schedule
\(\omega_{1:n}=(\omega_1,\ldots,\omega_n)\), define the normalized remainder by
\(\widetilde\omega_t:=\omega_t/(1-\omega_1)\) for \(2\leq t\leq n\).
Its coefficients are positive and satisfy
\(\sum_{t=2}^n \widetilde\omega_t=1\). Then
\begin{equation}
 C_n(\omega_{1:n})=\frac{\omega_1^2}{1-\omega_1}
 +(1-\omega_1)\,C_{n-1}(\widetilde\omega_{2:n}).
 \label{eq:normalized-tail-cost-scaling}
\end{equation}
\end{lemma}
\begin{proof}[Proof sketch.]
Separate \(\omega_1\)'s contribution in \eqref{eq:shape-score} and substitute
\(\omega_t=(1-\omega_1)\widetilde\omega_t\). Each later ratio \(\omega_k^2/\sum_{t>k}\omega_t\)
has \((1-\omega_1)^2\) in its numerator and \((1-\omega_1)\) in its denominator;
factoring the remaining \((1-\omega_1)\) from every term, including the terminal
entry, gives \((1-\omega_1)\,C_{n-1}(\widetilde\omega_{2:n})\).
\Cref{app:exact-schedule-optimization} gives the full proof.
\end{proof}
{\color{black}%
For each fixed \(\omega_1\in(0,1)\), the normalized remainder \(\widetilde\omega_{2:n}\)
can be any positive schedule with unit sum: multiplying its entries by
\(1-\omega_1\) gives the corresponding coefficients \(\omega_2,\ldots,\omega_n\).
Since \(1-\omega_1>0\), taking the infimum over these remainders in
\eqref{eq:normalized-tail-cost-scaling} replaces \(C_{n-1}(\widetilde\omega_{2:n})\) by
\(C_{n-1}^\star\). We are left with
}%
\begin{equation}
 C_n^\star=\inf_{0<\omega_1<1}
 \left\{\frac{\omega_1^2}{1-\omega_1}
                 +(1-\omega_1)C_{n-1}^\star\right\},\qquad n\geq2.
 \label{eq:optimal-scalar-problem}
\end{equation}
The objective in \eqref{eq:optimal-scalar-problem} is strictly convex in
\(\omega_1\), so each step has a unique closed-form minimizer; iterating from
\(C_1^\star=1\) gives both the optimal cost and the optimal schedule.

\begin{theorem}{Optimal cost and uniqueness}{optimal-schedule}
For every integer \(n\geq1\), \(C_n^\star>0\) is attained by a unique positive
normalized schedule and satisfies
\begin{equation}
 C_1^\star=1,\qquad
 C_n^\star=2\bigl(\sqrt{1+C_{n-1}^\star}-1\bigr)\quad(n\geq2).
 \label{eq:optimal-cost-recursion}
\end{equation}
\end{theorem}
\begin{proof}[Proof sketch.]
{\color{black}%
For fixed \(\omega_1\), \eqref{eq:normalized-tail-cost-scaling} is minimized
by the shorter optimizer. Minimizing the remaining expression in
\eqref{eq:optimal-scalar-problem} determines \(\omega_1\) uniquely.
Induction from \(C_1^\star=1\) then gives the cost recursion and a unique
schedule attaining it.
}%
\Cref{app:exact-schedule-optimization} gives the full proof.
\end{proof}
The theorem gives the minimum cost and shows that exactly one schedule achieves it.
The following corollary gives a recursion for computing each entry of that schedule.

\begin{corollary}{Construction of the optimal normalized schedule}{optimal-allocation}
For every integer \(T\geq1\), the unique optimal normalized schedule
\(\omega_{1:T}^\star\) is constructed using the costs from
\Cref{thm:optimal-schedule} as
\begin{equation}
 \begin{aligned}
 \omega_k^\star
 &=\left(1-\sum_{j=1}^{k-1}\omega_j^\star\right)
   \left(1-\frac{1}{\sqrt{1+C_{T-k}^\star}}\right),
 \quad 1\leq k<T, \qquad
 \omega_T^\star=1-\sum_{j=1}^{T-1}\omega_j^\star.
 \end{aligned}
 \label{eq:optimal-allocation-recursion}
\end{equation}
\end{corollary}
\begin{proof}[Proof sketch.]
{\color{black}%
For \(k<T\), the remaining coefficients divided by their sum form the
unique optimizer of length \(T-k+1\). Minimizing
\eqref{eq:optimal-scalar-problem} at this shorter length determines its first
entry. Multiplying that entry by the remaining sum gives
\(\omega_k^\star\) in \eqref{eq:optimal-allocation-recursion}.
The final entry receives the remaining sum.
}%
\Cref{app:exact-schedule-optimization} gives the full proof.
\end{proof}
This construction also provides an analytic solution to the constant-envelope
functional optimized numerically by
\citet[App.~C.3]{schaipp2025surprising} for the Euclidean case.
Together with \Cref{prop:optimal-mass}, the optimal normalized schedule determines the unique coefficient schedule that minimizes the bound:
\(s_t^\star=M^\star(\omega_{1:T}^\star)\omega_t^\star\) for \(D,G>0\).

\subsection{The Optimal Schedule Converges to Linear Decay}
\label{sec:optimal-profile}
The recursion in \Cref{cor:optimal-allocation} gives the exact optimizer at each
finite horizon; we now ask what shape it approaches as the horizon grows.
For a fraction \(0\leq\tau<1\) of training completed, write
\(\omega_{1+\lfloor \tau(T-1)\rfloor}^\star\) for the coefficient at that position.

\begin{proposition}{Asymptotic cost and linear schedule}{optimal-profile-limit}
As \(T\to\infty\), the optimal cost decays as
\(C_T^\star\sim4/T\). At every fixed training fraction \(0\leq\tau<1\),
the corresponding coefficient satisfies
\begin{equation}
 \omega_{1+\lfloor \tau(T-1)\rfloor}^\star
 \sim\frac{2(1-\tau)}{T}.
 \label{eq:optimal-profile-limit}
\end{equation}
\end{proposition}
\begin{proof}[Proof sketch.]
{\color{black}%
The cost recursion \eqref{eq:optimal-cost-recursion} gives
\(C_T^\star\sim4/T\). We then compare the remaining coefficient sums
generated by \eqref{eq:optimal-allocation-recursion} with the optimal
costs at the corresponding horizons. A telescoping argument gives the
linear-decay limit.
}%
\Cref{app:optimal-profile-proof} gives the full proof.
\end{proof}

Since the optimum approaches linear decay, a natural question is whether the
closed-form linear schedule itself achieves the same leading bound value. Define its positive normalized
schedule by
\(\omega_t^{\mathrm{lin}}:=2(T+1-t)/[T(T+1)]\) for \(1\leq t\leq T\).
The exact calculation in \Cref{app:linear-benchmark} gives
\begin{equation}
 C_T(\omega_{1:T}^{\mathrm{lin}})=\frac4T+
 O\!\left(\frac{\log T}{T^2}\right).
 \label{eq:linear-profile-leading-cost}
\end{equation}
After optimizing the amplitude of each schedule, \eqref{eq:optimal-mass} yields
\begin{equation}
 \frac{\min_{M>0}\Omega_T(M,\omega_{1:T}^{\mathrm{lin}})}
      {\min_{M>0,\,\omega_{1:T}}\Omega_T(M,\omega_{1:T})}
 =\sqrt{\frac{C_T(\omega_{1:T}^{\mathrm{lin}})}{C_T^\star}}
 \longrightarrow1.
 \label{eq:linear-bound-ratio-main}
\end{equation}
For large horizons, the bound penalty for replacing the exact finite-horizon
optimum with the closed-form linear schedule vanishes.

{\color{black}%
\Cref{fig:schedule-bound-theory} evaluates the bound in \Cref{thm:last-iterate}
for schedules chosen for \(T=20000\). Each schedule uses its full-horizon
optimal amplitude from \Cref{prop:optimal-mass}. The normalized endpoint is
\(\sqrt{C_T(\omega_{1:T})}\), so \(C_T\) orders the final values. Earlier values
use prefixes of the same schedule at that amplitude.
}%
The near-coincidence of the exact optimum with linear decay shows a small bound penalty at this horizon. Cosine is also close, while the specified WSD plateau and constant profile incur larger penalties.
\Cref{app:schedule-comparison-protocol} gives the full calculation and extends
the comparison across horizons and amplitudes.

\begin{takeaway}
Optimizing analytically over all positive rotational coefficient schedules yields a unique exact finite-horizon
solution. Its profile approaches linear decay as the training horizon increases.
\end{takeaway}

%% file: hyperball.tex
\section{Rotational Scheduling in LLM Pretraining}
\label{sec:hyperball}
The last-iterate analysis is derived for geodesically convex losses on a spherical cap. We use its rotational interpretation to predict scheduling behavior and construct learning-rate schedules in nonconvex LLM pretraining, with and without norm control. Hyperball \citep{wen2026hyperball} makes rotational motion directly controllable by fixing each matrix radius and normalizing its update.

\subsection{Hyperball as a Rotational Control}
\label{sec:hyperball-control}
Recent spherical optimizers expose the learning rate as a control on directional motion, but the best schedule for these on-the-sphere trajectories is unknown \citep{haegele2026md,huebler2026angularmuown,xiao2026hyperballfreelunch}. A prominent example is Hyperball, which fixes each matrix radius and normalizes its update. For a matrix \(\mathbf X_t\) with \(R=\|\mathbf X_1\|_F>0\), write \(\mathbf V_t:=\mathbf U_t/\|\mathbf U_t\|_F\) for the normalized base-optimizer update. The update rule is
\begin{equation}
 \mathbf X_{t+1}=R\frac{\mathbf X_t-\eta_tR\mathbf V_t}
 {\|\mathbf X_t-\eta_tR\mathbf V_t\|_F},
 \qquad \|\mathbf V_t\|_F=1.
 \label{eq:hyperball-wrapper}
\end{equation}
Vectorizing and dividing by \(R\) gives \(\xhat_{t+1}=(\xhat_t-\eta_t\bm v_t)/\|\xhat_t-\eta_t\bm v_t\|\), where \(\bm v_t\) is the vectorized direction. If \(\langle\xhat_t,\bm v_t\rangle=0\), its coefficient in \eqref{eq:exact-finite-step} and its turning angle are
\begin{equation}
 s_t^{\mathrm{HB}}=\eta_t,\qquad \dphi_t=\arctan s_t^{\mathrm{HB}}.
 \label{eq:hyperball-rotation}
\end{equation}
Under tangency, scheduling \(\eta_t\) therefore schedules rotation directly, without any intervening norm dynamics. \Cref{app:hyperball-geometry} gives the angle derivation.

We train 124M-parameter Llama models on FineWeb for 20,000 updates using AdamH and MuonH, which combine Hyperball with Adam \citep{kingma2014adam} and Muon \citep{jordan2024muon} directions. We compare linear decay, the asymptotic form of the bound optimum, with cosine, WSD, and constant schedules. For each schedule, we tune the peak learning rate separately on one random seed and report three fresh-seed runs at that fixed tuned rate.
Training details are in \Cref{app:hyperball-protocol}. The measured MuonH and AdamH angles follow \(\arctan\eta_t\) across all four schedules (\Cref{fig:hyperball-rotation}), showing that the learning rate closely controls rotation as in \eqref{eq:hyperball-rotation}.
\Cref{app:hyperball-containment,app:local-scale-sensitivity} report the observed hemispherical containment and local scale sensitivity along training.

\subsection{Predicting Schedule Performance}
\label{sec:hyperball-performance}
We first test whether the framework accounts for schedule performance under direct rotational control. Under the spherical bound of \Cref{thm:last-iterate}, the schedule-shape cost $C_T$ is lower for linear and cosine than for WSD and constant profiles (\Cref{fig:schedule-bound-theory}).

For these four profiles, Hyperball pretraining shows the ordering favored by the spherical bound. WSD's stable phase leaves rotation flat, while linear and cosine anneal it throughout the post-warmup horizon; \(C_T\) favors the latter schedules. Separately tuned linear and cosine schedules reach similar final losses, both lower than those of WSD and the constant schedule (\Cref{fig:hyperball-validation}).

The bound also captures the relative preferred learning rates and sensitivity curves in the Hyperball sweeps (\Cref{fig:hyperball-lr-sensitivity}). Linear and cosine have similar loss minima and nearly overlapping curves, while WSD and constant prefer smaller peak rates and end at higher losses, matching the bound.

Without Hyperball, WSD becomes competitive with linear and cosine (\Cref{fig:prediction-overview,fig:hyperball-unconstrained-loss,fig:hyperball-lr-sensitivity}). Thus its disadvantage under rotational control does not carry over to these runs. This is consistent with \citet[Sec.~4.1.5, Fig.~9]{haegele2026md}, who find linear decay better than the tested WSD variants on the sphere and explain that norm growth supplies implicit annealing during the stable phase of training without a fixed radius.

\begin{figure}[htbp]
 \centering
 \includegraphics[width=0.65\linewidth]{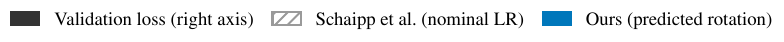}
 \includegraphics[width=.65\linewidth]{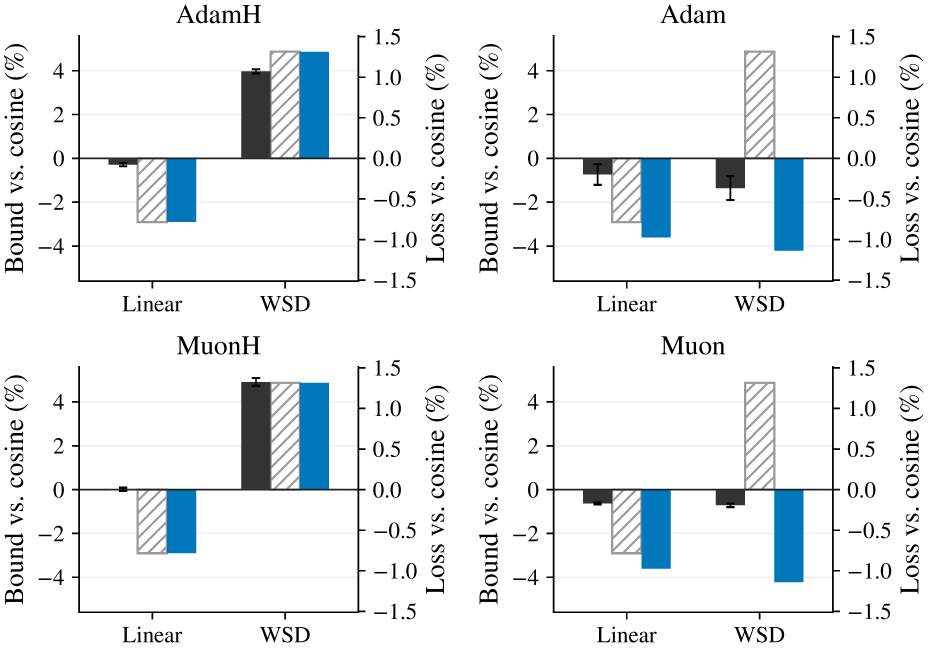}
 \vspace*{-0.2cm}
 \caption{124M schedule comparisons relative to cosine, with Hyperball on the
 left and without it on the right. Dark bars show final validation loss differences (averaged over three random seeds), with one sample standard deviation (right axis).
 Hatched bars show the bound on learning rates of \citet{schaipp2025surprising}; blue bars show our bound
 on predicted rotations (left axis). Each profile receives its optimal amplitude
 over updates 2,001--20,000.}
 \label{fig:bound-contrast}
\end{figure}

For an approximately tangent optimizer update, \(\tan\dphi_t\approx\eta_t\|\mathbf U_t\|_F/\|\mathbf X_t\|_F\). We adapt existing models \citep{wen2026hyperball,anonymous2026how} to predict the update magnitude and norm growth, and hence the normalized rotational profiles, from schedule shapes and fixed optimizer settings. The predicted profiles capture the measured rotational cooling; \Cref{app:rotation-diagnostic} gives the construction and comparison with measurements. Evaluating our rotational bound on these fully predicted shapes, with each shape's amplitude optimized analytically, recovers the observed WSD--cosine reversal (\Cref{fig:prediction-overview,fig:bound-contrast}). Under Hyperball, the learning-rate and rotational predictions coincide. Without Hyperball, \citet{schaipp2025surprising}'s bound on learning rates favors cosine over WSD, while our bound on predicted rotations favors WSD, in agreement with the validation loss of real runs across model scales (124M and 720M) and optimizers (Adam and Muon).

\vspace{-0.2cm}
We repeat the linear, cosine, and WSD comparison at 720M parameters with MuonH and Muon, training on FineWeb for 16,000 updates, or about 16.25B tokens, with 2,000 warmup updates and the same tuning and three-seed evaluation procedure. The larger runs reproduce the comparison: linear and cosine end at similar losses under Hyperball, both lower than the loss for WSD; without Hyperball, WSD ends at a lower loss than cosine and a slightly lower loss than linear. \Cref{fig:prediction-overview} shows the unconstrained prediction and training curves; \Cref{app:hyperball-720m} gives the full comparison and the predicted and measured profiles.

\subsection{Constructing Learning-Rate Schedules}
\label{sec:constructed-schedules}
To test the framework as a guide for schedule design, we use the norm models to construct learning rates targeting the rotational profile that minimizes our bound. We prescribe the exact normalized profile of \Cref{cor:optimal-allocation} over the post-warmup horizon, then calculate the learning rates needed to produce it as each matrix's norm evolves. Compensating for norm growth can require learning rates to keep rising even while the target rotation decreases. The construction recovers learning-rate shapes with prolonged increases, similar to those found effective empirically in prior works \citep{geiping2023cramming,topollai2026warmup}. The relative rates across matrices follow from the model and we tune one overall peak per optimizer by grid search. All candidate schedules are computed \textit{before} training from initialization, matrix dimensions, and optimizer settings. \Cref{app:norm-aware-schedule} gives the construction and its shared-rate ablation.

\begin{figure}[htbp]
 \centering
 \includegraphics[width=0.8\linewidth]{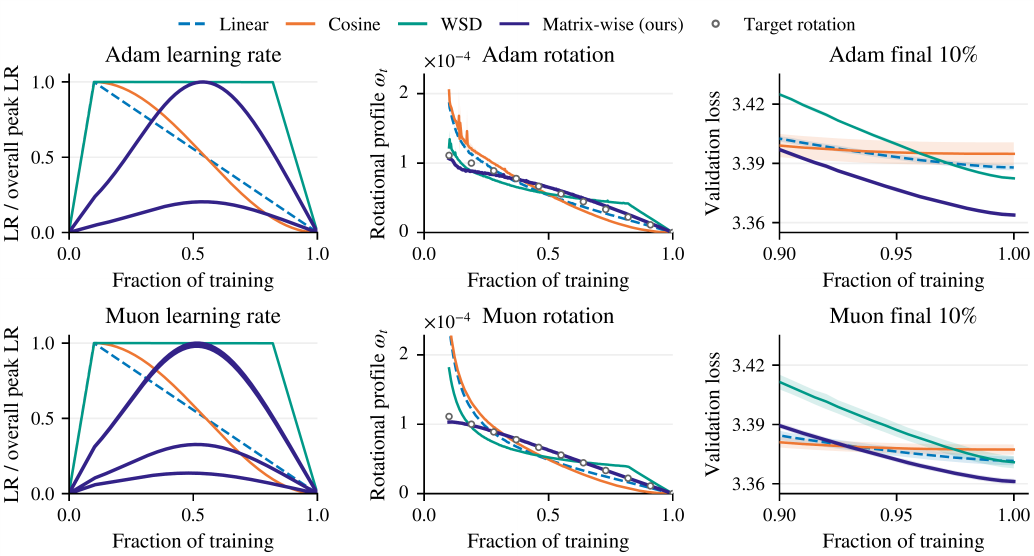}
 \vspace*{-0.2cm}
 \caption{Constructed schedules for Adam (top) and Muon (bottom), compared with
 linear, cosine, and WSD. Left: learning rates divided by each schedule's overall
 peak; purple curves represent distinct matrix groups. Center: measured unit-sum
 rotational profiles, with the theoretical target shown by the circles.
 Right: validation loss over the final 10\%.
 Measured curves and bands show means and standard deviation across
 three evaluation seeds.}
 \label{fig:constructed-schedules-main}
\end{figure}

\Cref{fig:constructed-schedules-main} compares the matrix-wise construction with independently tuned linear, cosine, and WSD for Adam and Muon without Hyperball or weight decay. Measured rotational profiles follow the prescribed linear decay. The constructed schedule improves final validation loss over the best tested standard schedule, WSD, and achieves a lower loss than each standard schedule in all three evaluation seeds. \Cref{fig:norm-aware-schedule} shows the full comparison for both optimizers.

\vspace{-0.4cm}

\FloatBarrier

These results demonstrate the predictive power of our rotational interpretation despite the simplifying assumptions of the analysis: we study normalized tangent SGD under geodesic convexity with a fresh, conditionally unbiased tangent oracle. The pretraining experiments, instead, use nonconvex losses and transformed momentum updates from Adam and Muon. We leave to future work a general study to understand how and why such assumptions still yield useful predictions in practice. 

\begin{takeaway}
Our rotational interpretation predicts motion and schedule performance across norm-control settings, recovering the WSD--cosine reversal missed by the learning-rate predictions of \citet{schaipp2025surprising}. Schedules derived from the theory improve on the tested standard profiles for both Adam and Muon.
\end{takeaway}

%% file: conclusion.tex
\vspace*{-0.1cm}
\section{Conclusion}
\label{sec:conclusion}
\vspace*{-0.1cm}
We showed that for stochastic tangent updates renormalized onto the sphere within a geodesically convex spherical cap, a new progress estimate gives a last-iterate bound for arbitrary positive rotational schedules. We analytically determine the unique schedule that minimizes this bound: it converges asymptotically to linear decay. We combine the spherical scheduling analysis with models of norm dynamics into a rotational interpretation that predicts motion and schedule performance with -- Hyperball -- and without -- vanilla Adam and Muon -- norm control in 124M and 720M pretraining. This interpretation recovers the WSD--cosine reversal missed by the learning-rate predictions of \citet{schaipp2025surprising}. From the same interpretation, we derive Adam and Muon schedules that follow the prescribed rotational decay and improve on the tested standard schedules cosine and WSD. Agreement with training dynamics and improved schedule performance support this rotational interpretation as a useful guide to LLM pretraining.

%% file: acknowledgments.tex
\section*{Acknowledgments}

Antonio Orvieto acknowledges the financial support of the Hector Foundation.

%% file: references.tex
\bibliography{main}

%% file: appendix.tex
\clearpage
\startcontents[appendices]
\section*{Appendix Contents}
\begingroup
\footnotesize
\printcontents[appendices]{}{1}{\setcounter{tocdepth}{2}}
\endgroup
\bigskip
\input{related-work.tex}
\section{Geometry and Proof of the Rotational Coefficient Scheduling Bound}
\label[appendix]{sec:appendix}
\label[appendix]{app:directional-schedules}
We give the geometric derivation and the complete proof of \Cref{thm:last-iterate}.
The notation and assumptions are those of \Cref{sec:directional-schedules}.

For reference throughout the proofs, we restate the assumptions of
\Cref{thm:last-iterate} in full.

\begin{assumption}{Spherical stochastic optimization model}{oracle}
\begin{enumerate}[label=(\roman*),leftmargin=2.4em,labelsep=.6em,
  topsep=0pt,itemsep=.35\baselineskip,parsep=0pt]
\item \textbf{\upshape Loss and initialization.}
The population loss \(L\) is differentiable and geodesically convex on \(\mathcal C\),
{\color{black}%
with a minimizer \(\xhat^\star\in\mathcal C\).
}%
The initial point \(\xhat_1\in\mathcal C\) is deterministic.
\item \textbf{\upshape Coefficients.}
The sequence \(s_1,\ldots,s_T\) is positive, deterministic, and fixed independently
of the oracle draws.
\item \textbf{\upshape Fresh tangent oracle.}
Let \(\mathcal F_t\) be the history before draw \(t\), including the current point
and earlier samples. Almost surely,
\[
 \inner{\gtilde_t}{\xhat_t}=0,\qquad
 \E[\gtilde_t\mid\mathcal F_t]=\nabla_{\mathbb S}L(\xhat_t).
\]
\item \textbf{\upshape Second moment.}
For a deterministic \(G>0\) and every \(1\leq t\leq T\),
\[
 \E\norm{\gtilde_t}^2\leq G^2.
\]
\item \textbf{\upshape Trajectory.}
The iterates of \eqref{eq:exact-finite-step}, including \(\xhat_{T+1}\), remain
in \(\mathcal C\) almost surely.
\end{enumerate}
\end{assumption}

\paragraph{Proof outline.}
\begin{enumerate}
 \item \textbf{From parameters to directions (\Cref{app:si-directional-recursion}).}
 Scale invariance makes the raw gradient tangent and gives its inverse-radius scaling.
 Normalizing the gradient-descent update identifies the induced rotational coefficient
 \(s_t^{\mathrm{GD}}\) and proves \Cref{prop:si-directional-recursion}.
 \item \textbf{Geometry on the cap (\Cref{app:sphere-geometry}).}
 We construct the exponential and logarithm maps, verify that shortest arcs remain in the cap,
 and derive the first-order inequality for a geodesically convex loss.
 \item \textbf{One-step progress (\Cref{app:chord-progress}).}
 We prove the exponential representation of the normalized update and a deterministic
 distance inequality. Conditional unbiasedness and geodesic convexity then give
 expected one-step progress.
 \item \textbf{The last iterate (\Cref{app:last-iterate-proof}).}
 We sum that bound over tails and compare successive tail averages. Their telescoping identity
 isolates the final loss; collecting the remaining terms gives the normalized schedule-shape cost
 \(C_T(\omega_{1:T})\)
 and completes the proof of \Cref{thm:last-iterate}.
 \item \textbf{Separating coefficient sum and normalized schedule (\Cref{app:optimal-mass}).}
 The bound decomposes into coefficient sum and normalized schedule.
 Optimizing the coefficient sum at fixed normalized schedule proves \Cref{prop:optimal-mass}
 and reduces the schedule-design problem to minimizing the schedule-shape cost alone.
\end{enumerate}

\subsection{Scale-Invariant Gradients and the Exact Rotational Update}
\label[appendix]{app:si-directional-recursion}
Let the parameter space be \(\R^C\), \(C\geq2\), with unit sphere \(\Sph\).
We use the differentiable, positively scale-invariant loss of \Cref{def:scale-invariant-layer}. Choosing \(\kappa=1/\norm{\bx}\) gives
\(L(\bx)=L(\xhat)\), which identifies the loss with its restriction to unit directions.

\paragraph{Tangent directions and the raw gradient.}
At \(\xhat\in\Sph\), the tangent space is
\[
 T_{\xhat}\Sph
 :=\left\{\bm v\in\R^C:\inner{\bm v}{\xhat}=0\right\}.
\]
Indeed, any differentiable sphere curve \(\gamma\) through \(\gamma(0)=\xhat\) satisfies
\(\gamma(a)\in\Sph\), and hence \(\norm{\gamma(a)}^2=1\), for every value of its
curve parameter. Differentiating this constant identity gives
\[
 \begin{aligned}
 0
 &=\frac{\mathrm d}{\mathrm da}\inner{\gamma(a)}{\gamma(a)}\\
 &=\inner{\gamma'(a)}{\gamma(a)}+\inner{\gamma(a)}{\gamma'(a)}
 =2\inner{\gamma(a)}{\gamma'(a)}.
 \end{aligned}
\]
Here \(\gamma'\) denotes differentiation with respect to the curve parameter. Evaluating
at \(a=0\) yields \(\inner{\xhat}{\gamma'(0)}=0\). Thus, a sphere curve has tangent
velocity. Conversely,
every tangent vector is the velocity of a sphere curve, as the explicit construction below
verifies.

Stretching the weight along its ray gives
\[
 L(\bx+h\bx)=L((1+h)\bx)=L(\bx)\qquad(h>-1).
\]
The left-hand side is therefore constant as a function of \(h\). Since
\(\frac{\mathrm d}{\mathrm dh}(\bx+h\bx)=\bx\), the ordinary chain rule, following
\citet[App.~B]{kosson2024rotational}, gives
\[
 \begin{aligned}
 0
 &=\left.\frac{\mathrm d}{\mathrm dh}L(\bx+h\bx)\right|_{h=0}\\
 &=\left.\inner{\nabla_{\bx}L(\bx+h\bx)}{\bx}\right|_{h=0}
 =\inner{\nabla_{\bx}L(\bx)}{\bx}.
 \end{aligned}
\]
This zero inner product says that the raw gradient is orthogonal to the radial direction \(\bx\).
Since \(\norm{\bx}>0\), division by the radius gives
\[
 \inner{\nabla_{\bx}L(\bx)}{\xhat}
 =\frac{\inner{\nabla_{\bx}L(\bx)}{\bx}}{\norm{\bx}}=0.
\]
Hence the raw gradient belongs to \(T_{\xhat}\Sph\).

\begin{definition}{Tangent space and spherical gradient}{tangent-space-and-gradient}
At a unit direction \(\xhat\), the tangent space is
\(T_{\xhat}\Sph:=\{\bm v:\inner{\bm v}{\xhat}=0\}\).
Its vectors are the possible instantaneous velocities of curves on the sphere through
\(\xhat\). The spherical gradient \(\nabla_{\mathbb S}L(\xhat)\) is the unique tangent
vector whose inner product with each such velocity gives the corresponding directional
derivative of the loss along the sphere.
\end{definition}

\paragraph{The intrinsic spherical gradient.}
Let \(f\) be differentiable on an open subset of the sphere. Its spherical gradient at
\(\xhat\) is the tangent vector characterized by
\begin{equation}
 \begin{gathered}
 \left.\frac{\mathrm d}{\mathrm da}f(\gamma(a))\right|_{a=0}
 =\inner{\nabla_{\mathbb S}f(\xhat)}{\bm v},\\
 \gamma(0)=\xhat,\qquad\gamma'(0)=\bm v\in T_{\xhat}\Sph.
 \end{gathered}
 \label{eq:spherical-gradient-definition}
\end{equation}

For any \(\bm v\in T_{\xhat}\Sph\), expanding a tangent proposal gives
\[
 \norm{\xhat+a\bm v}^2
 =\norm{\xhat}^2+2a\inner{\xhat}{\bm v}+a^2\norm{\bm v}^2
 =1+a^2\norm{\bm v}^2>0.
\]
Consequently the curve
\[
 \gamma(a)=\frac{\xhat+a\bm v}{\sqrt{1+a^2\norm{\bm v}^2}}
\]
lies on the sphere and stays in the open domain of \(f\) for sufficiently small
\(\lvert a\rvert\). Differentiating the numerator and denominator gives
\[
 \gamma'(a)
 =\frac{\bm v}{\sqrt{1+a^2\norm{\bm v}^2}}
 -\frac{a\norm{\bm v}^2(\xhat+a\bm v)}{(1+a^2\norm{\bm v}^2)^{3/2}},
 \qquad \gamma'(0)=\bm v.
\]
This supplies a curve for every tangent velocity, including \(\bm v=0\).

\begin{lemma}{Scale-invariant gradient and \mbox{spherical gradient}}{si-gradient-bridge}
For a differentiable scale-invariant loss on the open domain closed under positive
rescaling specified above, every \(\bx\) in that domain satisfies
\begin{equation}
 \nabla_{\bx}L(\bx)=\frac{1}{\norm{\bx}}\nabla_{\mathbb S}L(\xhat),
 \qquad \xhat=\frac{\bx}{\norm{\bx}}.
 \label{eq:si-gradient-bridge-proof}
\end{equation}
\end{lemma}
\begin{proof}[Proof of \Cref{lem:si-gradient-bridge}]
Fix \(\bx\) and any \(\bm v\in T_{\xhat}\Sph\). Choose a sphere curve in the loss
domain with \(\gamma(0)=\xhat\) and \(\gamma'(0)=\bm v\), as constructed above.
The radius \(\norm{\bx}\) is constant as the curve parameter varies, and
\(\norm{\bx}\gamma(0)=\bx\). Scale invariance gives
\[
 L(\norm{\bx}\gamma(a))=L(\gamma(a)).
\]
The factor \(\norm{\bx}\) is constant along the curve, so
\[
 \left.\frac{\mathrm d}{\mathrm da}\bigl(\norm{\bx}\gamma(a)\bigr)\right|_{a=0}
 =\norm{\bx}\gamma'(0)=\norm{\bx}\bm v.
\]
The intrinsic gradient definition and the ordinary chain rule now yield
\[
 \begin{aligned}
 \inner{\nabla_{\mathbb S}L(\xhat)}{\bm v}
 &=\left.\frac{\mathrm d}{\mathrm da}L(\gamma(a))\right|_{a=0}\\
 &=\left.\frac{\mathrm d}{\mathrm da}L(\norm{\bx}\gamma(a))\right|_{a=0}\\
 &=\inner{\nabla_{\bx}L(\bx)}{\norm{\bx}\bm v}\\
 &=\norm{\bx}\inner{\nabla_{\bx}L(\bx)}{\bm v}.
 \end{aligned}
\]
Since this holds for every \(\bm v\in T_{\xhat}\Sph\), the tangent vector
\(\norm{\bx}\nabla_{\bx}L(\bx)\) satisfies the characterization
\eqref{eq:spherical-gradient-definition} of the spherical gradient, so
\(\nabla_{\mathbb S}L(\xhat)=\norm{\bx}\nabla_{\bx}L(\bx)\).
Dividing by \(\norm{\bx}>0\) gives \eqref{eq:si-gradient-bridge-proof}.
\end{proof}

Positive rescaling preserves the unit direction, since
\(\norm{\kappa\bx}=\kappa\norm{\bx}\) and
\(\kappa\bx/\norm{\kappa\bx}=\xhat\). Applying the bridge at \(\kappa\bx\) gives
\[
 \nabla_{\bx}L(\kappa\bx)
 =\frac{\nabla_{\mathbb S}L(\xhat)}{\kappa\norm{\bx}}
 =\kappa^{-1}\nabla_{\bx}L(\bx).
\]
This is the inverse-radius raw-gradient scaling used in the scale-invariant analysis
of \citet{vanLaarhoven2017,liArora2020}.

\begin{proof}[Completion of the proof of \Cref{prop:si-directional-recursion}]
\emph{The exact rotational gradient step.}
Choose Euclidean gradient descent with a finite step size \(\eta_t>0\). Substituting
\eqref{eq:si-gradient-bridge-proof} and factoring the radius gives
\begin{align}
 \bx_{t+1}&=\bx_t-\eta_t\nabla_{\bx}L(\bx_t)\label{eq:plain-gd-update}\\
 &=\norm{\bx_t}\xhat_t-\frac{\eta_t}{\norm{\bx_t}}\nabla_{\mathbb S}L(\xhat_t)\nonumber\\
 &=\norm{\bx_t}\left(\xhat_t-\frac{\eta_t}{\norm{\bx_t}^2}
                                  \nabla_{\mathbb S}L(\xhat_t)\right).\nonumber
\end{align}
Define the rotational coefficient
\begin{equation}
 s_t^{\mathrm{GD}}:=\frac{\eta_t}{\norm{\bx_t}^2}.
 \label{eq:raw-gradient-coefficient-proof}
\end{equation}
Tangency removes the cross term in the remaining proposal:
\[
 \begin{aligned}
 \norm{\xhat_t-s_t^{\mathrm{GD}}\nabla_{\mathbb S}L(\xhat_t)}^2
 &=\norm{\xhat_t}^2-2s_t^{\mathrm{GD}}\inner{\xhat_t}{\nabla_{\mathbb S}L(\xhat_t)}
     +(s_t^{\mathrm{GD}})^2\norm{\nabla_{\mathbb S}L(\xhat_t)}^2\\
 &=1+(s_t^{\mathrm{GD}})^2\norm{\nabla_{\mathbb S}L(\xhat_t)}^2>0.
 \end{aligned}
\]
Taking the norm of the factored weight step therefore gives
\[
 \norm{\bx_{t+1}}=\norm{\bx_t}
 \sqrt{1+(s_t^{\mathrm{GD}})^2\norm{\nabla_{\mathbb S}L(\xhat_t)}^2}>0.
\]
Define \(\xhat_{t+1}=\bx_{t+1}/\norm{\bx_{t+1}}\). Substitution displays the positive factor
that cancels:
\[
 \xhat_{t+1}
 =\frac{\norm{\bx_t}\bigl(\xhat_t-s_t^{\mathrm{GD}}\nabla_{\mathbb S}L(\xhat_t)\bigr)}
 {\norm{\bx_t}\sqrt{1+(s_t^{\mathrm{GD}})^2\norm{\nabla_{\mathbb S}L(\xhat_t)}^2}}.
\]
Thus the exact rotational recursion is
\begin{equation}
 \xhat_{t+1}
 =\frac{\xhat_t-s_t^{\mathrm{GD}}\nabla_{\mathbb S}L(\xhat_t)}
 {\sqrt{1+(s_t^{\mathrm{GD}})^2\norm{\nabla_{\mathbb S}L(\xhat_t)}^2}},
 \qquad s_t^{\mathrm{GD}}=\frac{\eta_t}{\norm{\bx_t}^2}.
 \label{eq:raw-gradient-directional-step-proof}
\end{equation}

The coefficient \eqref{eq:raw-gradient-directional-step} is the geometric effective learning
rate of \citet[Sec.~3.1]{vanLaarhoven2017}.
One inverse-radius factor comes from the raw-gradient bridge; the other comes from
factoring the weight radius before normalization.

Now let us recall the turning angle, defined
\(\dphi_t:=\angle(\bx_t,\bx_{t+1})=\angle(\xhat_t,\xhat_{t+1})\); positive normalization
preserves the angle between the two vectors. Tangency makes \(\xhat_t\) and
\(-\nabla_{\mathbb S}L(\xhat_t)/\norm{\nabla_{\mathbb S}L(\xhat_t)}\)
an orthonormal pair when the gradient is nonzero. The normalized endpoint has coordinates
\[
 \cos\dphi_t=\frac{1}{\sqrt{1+(s_t^{\mathrm{GD}})^2\norm{\nabla_{\mathbb S}L(\xhat_t)}^2}},\qquad
 \sin\dphi_t=\frac{s_t^{\mathrm{GD}}\norm{\nabla_{\mathbb S}L(\xhat_t)}}
 {\sqrt{1+(s_t^{\mathrm{GD}})^2\norm{\nabla_{\mathbb S}L(\xhat_t)}^2}}.
\]
The radial coordinate is positive, so
\(\dphi_t=\arctan(s_t^{\mathrm{GD}}\norm{\nabla_{\mathbb S}L(\xhat_t)})\in[0,\pi/2)\).
Combining the two orthonormal components gives, for nonzero gradient,
\[
 \xhat_{t+1}
 =\cos(\dphi_t)\xhat_t
 -\sin(\dphi_t)\frac{\nabla_{\mathbb S}L(\xhat_t)}
                           {\norm{\nabla_{\mathbb S}L(\xhat_t)}}.
\]
Substituting the displayed sine and cosine coordinates recovers
\eqref{eq:raw-gradient-directional-step-proof}, including the negative gradient sign.
This proves \eqref{eq:raw-gradient-angle} and \eqref{eq:raw-gradient-angular-step}.
\end{proof}

With the rotational update established, we now develop the geometric tools
needed to measure progress: distance on the sphere, the exponential and
logarithm maps, and the supporting inequality for geodesically convex losses.

\subsection{Sphere Geometry and Geodesic Convexity}
\label[appendix]{app:sphere-geometry}
For unit directions, the spherical distance is
\begin{equation}
 \dist(\xhat,\bm y):=\arccos\inner{\xhat}{\bm y}.
 \label{eq:spherical-distance}
\end{equation}
The exponential map follows the great circle selected by a tangent vector. For
\(\bm v\in T_{\xhat}\Sph\) with \(\bm v\ne0\), set
\begin{equation}
 \bm e:=\frac{\bm v}{\norm{\bm v}}.
\end{equation}
Tangency gives \(\inner{\bm e}{\xhat}=0\), while both vectors have unit norm. They
therefore form an orthonormal basis of the two-dimensional plane containing the motion.
Moving through angle \(\norm{\bm v}\) in that plane gives
\begin{equation}
 \begin{aligned}
 \Exp_{\xhat}(\bm v)
 &=\cos(\norm{\bm v})\xhat+\sin(\norm{\bm v})\bm e\\
 &=\cos(\norm{\bm v})\xhat+
   \sin(\norm{\bm v})\frac{\bm v}{\norm{\bm v}},\\
 \Exp_{\xhat}(0)&=\xhat.
 \end{aligned}
 \label{eq:sphere-exp}
\end{equation}
Varying the angle parametrizes a unit circle with unit speed, so the traveled arc has length
\(\norm{\bm v}\).

The logarithm map reverses this construction. Let \(\bm y\ne\xhat\) be non-antipodal to
\(\xhat\), and write \(\theta=\dist(\xhat,\bm y)\in(0,\pi)\). Since
\(\inner{\xhat}{\bm y}=\cos\theta\), the vector \(\cos\theta\,\xhat\) is the radial
component of \(\bm y\) along \(\xhat\). Subtracting it leaves the tangent component:
\begin{equation}
 \begin{aligned}
 \inner{\bm y-\cos\theta\,\xhat}{\xhat}
 &=\cos\theta-\cos\theta=0,\\
 \norm{\bm y-\cos\theta\,\xhat}^2
 &=1-2\cos^2\theta+\cos^2\theta=\sin^2\theta.
 \end{aligned}
\end{equation}
Therefore the vector
\begin{equation}
 \bm e:=\frac{\bm y-\cos\theta\,\xhat}{\sin\theta}
\end{equation}
is the unit tangent direction from \(\xhat\) toward \(\bm y\), and
\(\bm y=\cos\theta\,\xhat+\sin\theta\,\bm e\). The logarithm vector points in this
direction and has length \(\theta\), because Exp travels a unit-sphere arc whose length
equals its angle. Hence
\begin{equation}
 \Log_{\xhat}(\bm y)=\theta\bm e
 =\frac{\theta}{\sin\theta}(\bm y-\cos\theta\,\xhat),
 \qquad\Log_{\xhat}(\xhat)=0.
 \label{eq:sphere-log}
\end{equation}
Substitution into \eqref{eq:sphere-exp} returns \(\bm y\). These are the standard sphere
maps \citep[Secs.~7.2 and 10.2]{boumal2023}.

For \(0\leq a\leq1\), the point
\(\gamma(a)=\Exp_{\xhat}(a\Log_{\xhat}(\bm y))\) travels the fraction \(a\) of the
shortest arc from \(\xhat\) to \(\bm y\). The tangent vector records both direction and
arc length: scaling it scales the distance traveled along that same geodesic.
The tangent space is a vector space through the origin; the drawing attaches its translated
copy to \(\xhat\) so that the vector and its spherical endpoint can be compared.

\begin{figure}[!htbp]
\centering
\begin{subfigure}[t]{.48\linewidth}
\centering
\includegraphics[width=0.8\linewidth]{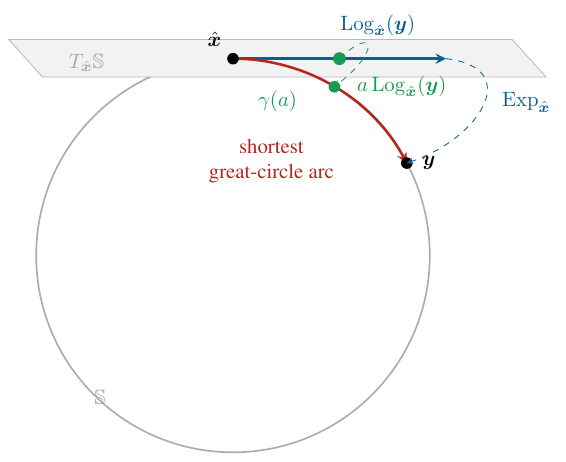}
\caption{}\label{fig:sphere-exp-log}
\end{subfigure}\hfill
\begin{subfigure}[t]{.48\linewidth}
\centering
\includegraphics[width=0.8\linewidth]{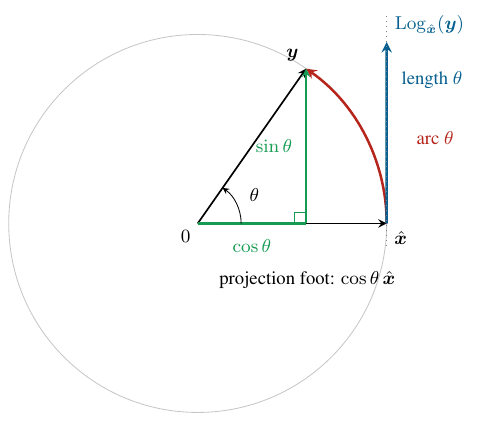}
\caption{}\label{fig:sphere-planar-construction}
\end{subfigure}
\caption{Sphere geometry used in the proof. \textbf{(a)} The tangent plane is drawn as
\(\xhat+T_{\xhat}\Sph\). The logarithm map assigns to \(\bm y\) a tangent vector
whose length is the shortest arc distance; the exponential map returns this vector, or any
fraction of it, to the corresponding point on that arc.
\textbf{(b)} In the plane spanned by \(\xhat\) and \(\bm y\), the radial projection is
\(\cos\theta\,\xhat\), the perpendicular residual has length \(\sin\theta\), and the
Log vector and arc have length \(\theta\).}
\label{fig:sphere-geometry}
\end{figure}

\FloatBarrier
Recall the open spherical cap in \eqref{eq:working-cap}:
{\color{black}%
\begin{equation}
 \mathcal C:=
 \left\{\xhat\in\Sph:\dist(\xhat,\bm c_{\mathrm{cap}})<\thetacap\right\},
 \qquad 0<\thetacap\leq\frac{\pi}{2},
\end{equation}
}%
\textcolor{black}{with fixed center \(\bm c_{\mathrm{cap}}\).} For any \(\xhat,\bm y\in\mathcal C\), the triangle
inequality gives
{\color{black}%
\begin{equation}
 \begin{aligned}
 \dist(\xhat,\bm y)
 &\leq\dist(\xhat,\bm c_{\mathrm{cap}})+\dist(\bm c_{\mathrm{cap}},\bm y)\\
 &<2\thetacap\leq\pi.
 \end{aligned}
\end{equation}
}%

\begin{corollary}{Shortest-arc containment of the cap}{cap-shortest-arc}
For any two points in \(\mathcal C\), the unique shortest spherical arc joining them
lies entirely in \(\mathcal C\).
\end{corollary}
\begin{proof}
{\color{black}%
Fix two unit directions \(\xhat,\bm y\in\mathcal C\), the endpoints of the arc.
If they coincide, the shortest arc is constant and lies in the cap.
For distinct endpoints, write \(\theta=\dist(\xhat,\bm y)\in(0,\pi)\), using
the distance bound above. Here \(\theta\) measures the separation of the
endpoints, while \(\thetacap\) is the cap radius.
Consider the sphere curve introduced earlier,
\[
 \gamma(a)=\Exp_{\xhat}\!\left(a\Log_{\xhat}(\bm y)\right),
 \qquad 0\leq a\leq1.
\]
The point \(\gamma(a)\) is a fraction \(a\) of the way along the shortest
spherical arc from \(\xhat\) to \(\bm y\). Containment of this arc means
\(\dist(\bm c_{\mathrm{cap}},\gamma(a))<\thetacap\) for every \(a\).
Both this distance and \(\thetacap\) lie in \([0,\pi]\), where cosine is
strictly decreasing. Applying cosine therefore reverses the inequality,
so our target is equivalently
\[
 \cos\!\bigl(\dist(\bm c_{\mathrm{cap}},\gamma(a))\bigr)>\cos\thetacap,
 \qquad 0\leq a\leq1.
\]
Since \(\bm c_{\mathrm{cap}}\) and \(\gamma(a)\) are unit directions,
the spherical-distance definition \eqref{eq:spherical-distance} identifies
the first cosine with their inner product. Thus the same target reads
\[
 \inner{\bm c_{\mathrm{cap}}}{\gamma(a)}
 =\cos\!\bigl(\dist(\bm c_{\mathrm{cap}},\gamma(a))\bigr)
 >\cos\thetacap.
\]

Each endpoint lies strictly within the cap radius of the center:
\[
 \dist(\bm c_{\mathrm{cap}},\xhat)<\thetacap,\qquad
 \dist(\bm c_{\mathrm{cap}},\bm y)<\thetacap.
\]
Using the cosine of each endpoint's distance to the center gives
\[
 \begin{aligned}
 \inner{\bm c_{\mathrm{cap}}}{\xhat}
 &=\cos\!\bigl(\dist(\bm c_{\mathrm{cap}},\xhat)\bigr)>\cos\thetacap,\\
 \inner{\bm c_{\mathrm{cap}}}{\bm y}
 &=\cos\!\bigl(\dist(\bm c_{\mathrm{cap}},\bm y)\bigr)>\cos\thetacap.
 \end{aligned}
\]

To use these endpoint bounds at an intermediate point, we express
\(\gamma(a)\) as a combination of \(\xhat\) and \(\bm y\). For an interior
arc point, these bounds will prove the target if we can find coefficients
satisfying
\[
 \gamma(a)=A(a)\xhat+B(a)\bm y,\qquad
 A(a),B(a)>0,\qquad A(a)+B(a)\geq1.
\]
Indeed, under these conditions, linearity of the inner product gives
\[
 \begin{aligned}
 \inner{\bm c_{\mathrm{cap}}}{\gamma(a)}
 &=A(a)\inner{\bm c_{\mathrm{cap}}}{\xhat}
   +B(a)\inner{\bm c_{\mathrm{cap}}}{\bm y}\\
 &>\bigl(A(a)+B(a)\bigr)\cos\thetacap\\
 &\geq\cos\thetacap.
 \end{aligned}
\]
The strict inequality uses positivity of both coefficients and the strict
endpoint bounds. For the last inequality, \(A(a)+B(a)\geq1\) can be
multiplied by \(\cos\thetacap\geq0\), since \(\thetacap\leq\pi/2\).

It remains to find the coefficients and verify their positivity and the bound
on their sum. To obtain the endpoint representation, we substitute
\eqref{eq:sphere-log} into \eqref{eq:sphere-exp}:
\[
 \begin{aligned}
 \gamma(a)
 &=\cos(a\theta)\xhat+
   \frac{\sin(a\theta)}{\sin\theta}
   \bigl(\bm y-\cos\theta\,\xhat\bigr)\\
 &=\left(\cos(a\theta)-
   \frac{\sin(a\theta)\cos\theta}{\sin\theta}\right)\xhat
   +\frac{\sin(a\theta)}{\sin\theta}\bm y\\
 &=\frac{\sin((1-a)\theta)}{\sin\theta}\xhat+
   \frac{\sin(a\theta)}{\sin\theta}\bm y.
 \end{aligned}
\]
The second line collects the coefficient of \(\xhat\).
The last line uses the sine subtraction identity
\(\sin((1-a)\theta)=\sin\theta\cos(a\theta)-\cos\theta\sin(a\theta)\).
Thus the desired representation is obtained by setting
\[
 A(a):=\frac{\sin((1-a)\theta)}{\sin\theta},\qquad
 B(a):=\frac{\sin(a\theta)}{\sin\theta}.
\]

At \(a=0\) and \(a=1\), this formula gives \(\xhat\) and \(\bm y\), which
already lie in the cap. For \(0<a<1\), both numerator angles lie in
\((0,\theta)\subset(0,\pi)\). Their sines and the denominator
\(\sin\theta\) are therefore positive, so \(A(a),B(a)>0\).

We finally verify the remaining condition \(A(a)+B(a)\geq1\).
The arc point \(\gamma(a)\) has unit norm, and the coefficients are positive
for \(0<a<1\), so the triangle inequality gives
\[
 \begin{aligned}
 1=\norm{\gamma(a)}
 &=\left\|\frac{\sin((1-a)\theta)}{\sin\theta}\xhat+
          \frac{\sin(a\theta)}{\sin\theta}\bm y\right\|\\
 &\leq\frac{\sin((1-a)\theta)}{\sin\theta}\norm{\xhat}
      +\frac{\sin(a\theta)}{\sin\theta}\norm{\bm y}\\
 &=\frac{\sin((1-a)\theta)+\sin(a\theta)}{\sin\theta}
 =A(a)+B(a).
 \end{aligned}
\]
The last line uses the unit norms of the endpoints and the definitions of
\(A(a)\) and \(B(a)\).

The constructed coefficients satisfy all three conditions above, so the
general inner-product argument gives
\(\inner{\bm c_{\mathrm{cap}}}{\gamma(a)}>\cos\thetacap\) for every
\(0<a<1\). Equivalently,
\(\dist(\bm c_{\mathrm{cap}},\gamma(a))<\thetacap\).
Every interior point of the arc therefore lies in \(\mathcal C\).
Together with the two endpoints, this proves containment of the entire arc.
}%
\end{proof}

\paragraph{Geodesic convexity and the supporting inequality.}
In Euclidean optimization, convexity compares the loss along a straight line segment
with the linear interpolation of the losses at its endpoints. On our spherical cap,
we make the same comparison while moving along the shortest spherical arc between
the endpoints. At a fraction \(a\) of the distance along the arc, the loss must be
at most \((1-a)L(\xhat)+aL(\bm y)\). This is \emph{geodesic convexity}.
The arc stays in the cap by \Cref{cor:cap-shortest-arc}.
\begin{definition}{Geodesic convexity on the cap}{geodesic-convexity}
For \(\xhat,\bm y\in\mathcal C\), let \(\gamma:[0,1]\to\mathcal C\) be the
constant-speed shortest spherical arc with \(\gamma(0)=\xhat\) and \(\gamma(1)=\bm y\).
Here \(a\in[0,1]\) denotes the fraction of the arc traversed.
The loss \(L\) is geodesically convex on \(\mathcal C\) if, for every such pair,
\begin{equation}
 L(\gamma(a))\leq(1-a)L(\xhat)+aL(\bm y),\qquad 0\leq a\leq1.
 \label{eq:geodesic-convexity}
\end{equation}
\end{definition}
The arc in \Cref{def:geodesic-convexity} has the representation
\begin{equation}
 \gamma(a):=\Exp_{\xhat}\!\left(a\Log_{\xhat}(\bm y)\right),
 \qquad 0\leq a\leq1.
 \label{eq:geodesic-restriction}
\end{equation}
For differentiable \(L\), Boumal's first-order characterization gives the following
supporting inequality.
\begin{lemma}{Supporting inequality on the cap}{convex-support}
If \(L\) is differentiable and geodesically convex on \(\mathcal C\), then
\begin{equation}
 L(\bm y)\geq L(\xhat)+
 \inner{\nabla_{\mathbb S}L(\xhat)}{\Log_{\xhat}(\bm y)}
 \qquad(\xhat,\bm y\in\mathcal C).
 \label{eq:convex-support}
\end{equation}
\end{lemma}
\begin{proof}[Proof of \Cref{lem:convex-support}]
Use the curve \(\gamma\) from \eqref{eq:geodesic-restriction}, which stays in
\(\mathcal C\) and satisfies \(\gamma(0)=\xhat\) and \(\gamma(1)=\bm y\).
By \citet[Theorem~11.21, Eq.~(11.6)]{boumal2023}, a function that is differentiable
and geodesically convex on the open cap lies above its first-order expansion along
every geodesic in the cap.
On the sphere, the Riemannian gradient is \(\nabla_{\mathbb S}L\) and inner products
are the usual Euclidean ones, so that the theorem reads
\[
 L(\gamma(a))\geq L(\xhat)
 +a\inner{\nabla_{\mathbb S}L(\xhat)}{\gamma'(0)},
 \qquad 0\leq a\leq1.
\]
Here the prime denotes differentiation with respect to \(a\).
Differentiating the spherical exponential formula \eqref{eq:sphere-exp} along
this curve at \(a=0\) identifies its initial velocity:
\begin{equation}
 \gamma'(0)=\Log_{\xhat}(\bm y).
 \label{eq:geodesic-initial-velocity}
\end{equation}
Substituting this velocity and setting \(a=1\) gives \eqref{eq:convex-support}.
For coincident endpoints, the
curve is constant, the Log vector is zero, and the inequality is an equality.
\end{proof}

Compared with the Euclidean first-order condition
\(L(\bm y)\geq L(\bm x)+\inner{\nabla L(\bm x)}{\bm y-\bm x}\),
\eqref{eq:convex-support} replaces the displacement \(\bm y-\bm x\) by
\(\Log_{\xhat}(\bm y)\): the tangent vector at \(\xhat\) that points along the
shortest arc toward \(\bm y\) and has length \(\dist(\xhat,\bm y)\).

\subsection{One-Step Progress for the Normalized Update}
\label[appendix]{app:chord-progress}
We now combine the geometric tools of the preceding subsection with the stochastic
oracle of \Cref{ass:oracle} to derive the expected one-step progress inequality for
the normalized update \eqref{eq:exact-finite-step}.
The standard Euclidean SGD calculation starts from
\begin{equation}
 \bm x^+=\bm x-s\gtilde
\end{equation}
and expands squared distance to a comparator \(\bm y\):
\begin{equation}
 \norm{\bm x^+-\bm y}^2
 =\norm{\bm x-\bm y}^2
 -2s\inner{\gtilde}{\bm x-\bm y}
 +s^2\norm{\gtilde}^2.
\end{equation}
On the sphere, the analogous distance comparison lifts both endpoints into a common
tangent space via the logarithm and exponential maps.
For an exponential-map step, the resulting inequality is the nonnegative-curvature case of
\citet[Lemma~6 and Corollary~8]{zhangSra2016gconvex}, in which their curvature factor
equals one; \Cref{lem:exp-contraction} gives a self-contained derivation on the sphere.
\begin{lemma}{Non-expansion of the spherical exponential map}{exp-contraction}
At \(\xhat\in\Sph\), let \(\bm v,\bm w\in T_{\xhat}\Sph\) satisfy
\(\norm{\bm v},\norm{\bm w}<\pi\). Then
\begin{equation}
 \dist(\Exp_{\xhat}(\bm v),\Exp_{\xhat}(\bm w))
 \leq\norm{\bm v-\bm w}.
 \label{eq:exp-contraction}
\end{equation}
Mapping two tangent vectors to the sphere by the exponential map does not increase
their separation: the spherical distance between the endpoints is at most the
Euclidean distance between the tangent vectors.
\end{lemma}
\begin{proof}[Proof of \Cref{lem:exp-contraction}]
We compare the length of the straight tangent segment with that of its exponential image. Set
\[
 \bm c(a)=(1-a)\bm v+a\bm w,\qquad
 r(a)=\norm{\bm c(a)},\qquad
 \bm e(a)=\frac{\bm c(a)}{r(a)}\quad\text{when }r(a)>0.
\]
For \(0\leq a\leq1\), the triangle inequality gives
\(r(a)\leq(1-a)\norm{\bm v}+a\norm{\bm w}<\pi\).
Primes denote derivatives with respect to \(a\). Differentiating
\(\norm{\bm e}^2=1\) gives \(2\inner{\bm e'}{\bm e}=0\), so
\(\bm e'\perp\bm e\) and the product rule on \(\bm c=r\bm e\) gives
\begin{equation}
 \begin{aligned}
 \bm c'&=r'\bm e+r\bm e',\\
 \norm{\bm c'}^2
 &=\norm{r'\bm e+r\bm e'}^2\\
 &=(r')^2\norm{\bm e}^2+2r'r\inner{\bm e}{\bm e'}
   +r^2\norm{\bm e'}^2\\
 &=(r')^2+r^2\norm{\bm e'}^2.
 \end{aligned}
 \label{eq:tangent-line-element}
\end{equation}
Since \(\xhat\) is fixed, the chain and product rules give
\[
 \begin{aligned}
 \frac{\mathrm d}{\mathrm da}\Exp_{\xhat}(\bm c(a))
 &=\frac{\mathrm d}{\mathrm da}
   \bigl[\cos(r(a))\,\xhat+\sin(r(a))\,\bm e(a)\bigr]\\
 &=\left(\frac{\mathrm d}{\mathrm da}\cos r\right)\xhat
   +\left(\frac{\mathrm d}{\mathrm da}\sin r\right)\bm e
   +\sin r\,\bm e'\\
 &=-r'\sin r\,\xhat+r'\cos r\,\bm e+\sin r\,\bm e'\\
 &=r'(-\sin r\,\xhat+\cos r\,\bm e)+\sin r\,\bm e'.
 \end{aligned}
\]
Tangency gives \(\inner{\xhat}{\bm e}=0\); differentiating this identity at fixed
\(\xhat\) gives \(\inner{\xhat}{\bm e'}=0\). Together with
\(\inner{\bm e}{\bm e'}=0\) and \(\norm{\xhat}=\norm{\bm e}=1\), these identities
eliminate the cross terms below. Expanding the squared speed and then using
\(0\leq\sin r\leq r\), we obtain
\begin{equation}
 \begin{aligned}
 \left\|\frac{\mathrm d}{\mathrm da}\Exp_{\xhat}(\bm c(a))\right\|^2
 &=\left\|r'(-\sin r\,\xhat+\cos r\,\bm e)+\sin r\,\bm e'\right\|^2\\
 &=(r')^2\norm{-\sin r\,\xhat+\cos r\,\bm e}^2
   +\sin^2r\,\norm{\bm e'}^2\\
 &\quad+2r'\sin r\bigl(-\sin r\,\inner{\xhat}{\bm e'}
                          +\cos r\,\inner{\bm e}{\bm e'}\bigr)\\
 &=(r')^2(\sin^2r+\cos^2r)+\sin^2r\,\norm{\bm e'}^2\\
 &=(r')^2+\sin^2r\,\norm{\bm e'}^2\\
 &\leq(r')^2+r^2\norm{\bm e'}^2
 =\norm{\bm c'}^2.
 \end{aligned}
 \label{eq:spherical-line-element}
\end{equation}

As \(a\) runs from zero to one, \(\Exp_{\xhat}(\bm c(a))\) traces a curve on
the sphere joining \(\Exp_{\xhat}(\bm v)\) to \(\Exp_{\xhat}(\bm w)\).
Its length is the integral of its speed, where speed is the norm of the derivative
with respect to \(a\). Spherical distance is the length of the shortest joining curve, so it is at
most this integral. Taking square roots of the preceding squared-speed inequality
then bounds the image speed by \(\norm{\bm c'(a)}\). Finally, the straight segment
has constant velocity \(\bm c'(a)=-\bm v+\bm w\), giving
\begin{equation}
\begin{aligned}
 \dist(\Exp_{\xhat}(\bm v),\Exp_{\xhat}(\bm w))
 &\leq\int_0^1\left\|\frac{\mathrm d}{\mathrm da}
          \Exp_{\xhat}(\bm c(a))\right\|\,\mathrm da\\
 &\leq\int_0^1\norm{\bm c'(a)}\,\mathrm da\\
 &=\int_0^1\norm{\bm w-\bm v}\,\mathrm da\\
 &=\norm{\bm w-\bm v}\int_0^1 1\,\mathrm da
 =\norm{\bm w-\bm v}.
\end{aligned}
\end{equation}
\end{proof}

\begin{figure}[!htbp]
\centering
\begin{subfigure}{.48\linewidth}
\centering
\includegraphics[width=0.8\linewidth]{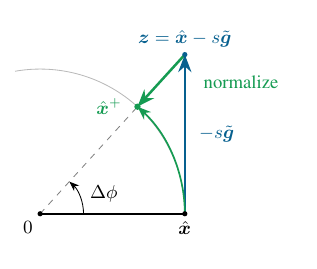}
\caption{}
\end{subfigure}\hfill
\begin{subfigure}{.48\linewidth}
\centering
\includegraphics[width=0.8\linewidth]{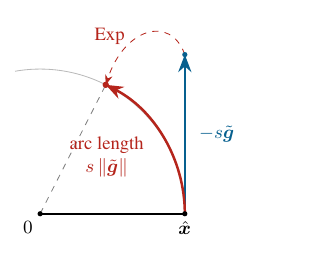}
\caption{}
\end{subfigure}
\caption{Two operations on the same tangent vector, shown at
\(s\norm{\gtilde}=1.1\) to make their difference visible. (a) Normalizing the proposal
\(\xhat-s\gtilde\) projects back to the sphere, turning by
\(\arctan(1.1)\approx0.833\) radians. (b) Applying Exp follows the great circle for
\(1.1\) radians. Both endpoints lie in the same plane.}
\label{fig:normalized-versus-exp}
\end{figure}

The non-expansion lemma applies to exponential-map endpoints, whereas the algorithm
normalizes a tangent proposal; \Cref{fig:normalized-versus-exp} illustrates the difference,
and the following lemma gives the exact correspondence.

\FloatBarrier

\begin{lemma}{Exponential representation of the normalized tangent update}{normalized-update-exp}
Let \(\xhat\in\Sph\), \(\gtilde\in T_{\xhat}\Sph\), and \(s>0\). Define
\[
 \xhat^+:=\frac{\xhat-s\gtilde}{\sqrt{1+s^2\norm{\gtilde}^2}},
 \qquad \chi:=s\norm{\gtilde},
\]
and
\begin{equation}
 h:=\begin{cases}\arctan\chi/\norm{\gtilde},&\gtilde\ne0,\\s,&\gtilde=0.\end{cases}
 \label{eq:chord-exponential-coefficient}
\end{equation}
Then
\begin{equation}
 \xhat^+=\Exp_{\xhat}(-h\gtilde),\qquad 0<h\leq s,
 \label{eq:chord-exponential-endpoint}
\end{equation}
and
\begin{equation}
 0\leq(s-h)\norm{\gtilde}=\chi-\arctan\chi\leq\frac{\chi^2}{4}.
 \label{eq:chord-quadratic-error}
\end{equation}
\end{lemma}
\begin{proof}
For \(\gtilde\ne0\), the definition gives \(h\norm{\gtilde}=\arctan\chi\).
Using \(\cos(\arctan\chi)=1/\sqrt{1+\chi^2}\) and
\(\sin(\arctan\chi)=\chi/\sqrt{1+\chi^2}\) in the exponential formula yields
\[
\begin{aligned}
 \Exp_{\xhat}(-h\gtilde)
 &=\cos(\arctan\chi)\xhat
   -\sin(\arctan\chi)\frac{\gtilde}{\norm{\gtilde}}\\
 &=\frac{\xhat-(\chi/\norm{\gtilde})\gtilde}{\sqrt{1+\chi^2}}
 =\frac{\xhat-s\gtilde}{\sqrt{1+s^2\norm{\gtilde}^2}}
 =\xhat^+.
\end{aligned}
\]
Since \(0<\arctan\chi\leq\chi\) for \(\chi>0\), division by
\(\norm{\gtilde}>0\) gives \(0<h\leq s\). If \(\gtilde=0\), then
\(h=s\), \(\chi=0\), and both endpoints equal \(\xhat\).

In either case, \((s-h)\norm{\gtilde}=\chi-\arctan\chi\). For every
\(\chi\geq0\), integration gives
\[
 0\leq\chi-\arctan\chi
 =\int_0^\chi\frac{a^2}{1+a^2}\,\mathrm da
 \leq\int_0^\chi\frac a2\,\mathrm da
 =\frac{\chi^2}{4}.
\]
The integrand inequality follows from
\(a/2-a^2/(1+a^2)=a(a-1)^2/[2(1+a^2)]\geq0\) for \(a\geq0\).
\end{proof}

The next lemma combines \Cref{lem:exp-contraction} with \Cref{lem:normalized-update-exp} to give a
distance inequality in the prescribed coefficient \(s\).
\begin{lemma}{Distance comparison for the normalized tangent update}{normalized-update-distance}
Let \(\xhat,\bm y\in\mathcal C\), \(\gtilde\in T_{\xhat}\Sph\), and \(s>0\),
and set \(\xhat^+=(\xhat-s\gtilde)/\sqrt{1+s^2\norm{\gtilde}^2}\). Then
\begin{equation}
\begin{aligned}
 \dist(\xhat^+,\bm y)^2
 \leq{}&\dist(\xhat,\bm y)^2
       +2s\inner{\gtilde}{\Log_{\xhat}(\bm y)}\\
 &+(1+\thetacap)s^2\norm{\gtilde}^2.
\end{aligned}
 \label{eq:normalized-update-distance}
\end{equation}
The left side is the squared distance to the comparator after one normalized step.
The right side has the same three-term structure as the Euclidean expansion above:
the squared distance before the step, a signed inner-product term that will become
the descent direction after taking conditional expectations, and a quadratic cost
proportional to \(s^2\norm{\gtilde}^2\).
\end{lemma}
\begin{proof}
Set \(\chi=s\norm{\gtilde}\) and use \(h\) from
\Cref{lem:normalized-update-exp}. Both \(-h\gtilde\) and
\(\Log_{\xhat}(\bm y)\) are tangent at \(\xhat\), with lengths
\[
 h\norm{\gtilde}=\arctan\chi<\pi/2,\qquad
{\color{black}%
 \norm{\Log_{\xhat}(\bm y)}=\dist(\xhat,\bm y)<2\thetacap\leq\pi.
}%
\]
Their exponential-map endpoints are \(\xhat^+\) and \(\bm y\), respectively.
Thus \Cref{lem:exp-contraction} applies. Squaring its inequality and expanding gives
\[
\begin{aligned}
 \dist(\xhat^+,\bm y)^2
 &\leq\norm{-h\gtilde-\Log_{\xhat}(\bm y)}^2\\
 &=\norm{\Log_{\xhat}(\bm y)}^2
   +2h\inner{\gtilde}{\Log_{\xhat}(\bm y)}+h^2\norm{\gtilde}^2\\
 &=\dist(\xhat,\bm y)^2
   +2h\inner{\gtilde}{\Log_{\xhat}(\bm y)}+h^2\norm{\gtilde}^2.
\end{aligned}
\]
The inner product can have either sign, so \(h\leq s\) alone does not justify
replacing its coefficient by \(s\). Add and subtract the term with coefficient \(s\):
\begin{equation}
\begin{aligned}
 2h\inner{\gtilde}{\Log_{\xhat}(\bm y)}
 ={}&\underbrace{2s\inner{\gtilde}{\Log_{\xhat}(\bm y)}}_{
       \text{term with coefficient }s}\\
 &+\underbrace{2(h-s)\inner{\gtilde}{\Log_{\xhat}(\bm y)}}_{
       \text{error from replacing }h\text{ by }s}.
\end{aligned}
 \label{eq:restore-deterministic-step}
\end{equation}
Since \(s-h\geq0\), the magnitude of the error is bounded by Cauchy--Schwarz:
\[
\begin{aligned}
 2(h-s)\inner{\gtilde}{\Log_{\xhat}(\bm y)}
 &\leq2(s-h)\left|\inner{\gtilde}{\Log_{\xhat}(\bm y)}\right|\\
 &\leq2(s-h)\norm{\gtilde}\norm{\Log_{\xhat}(\bm y)}\\
 &=2\dist(\xhat,\bm y)(s-h)\norm{\gtilde}.
\end{aligned}
\]
The estimate in \Cref{lem:normalized-update-exp} therefore gives
\begin{equation}
\begin{aligned}
 2\dist(\xhat,\bm y)(s-h)\norm{\gtilde}
 &=2\dist(\xhat,\bm y)(\chi-\arctan\chi)\\
 &\leq\frac{\dist(\xhat,\bm y)}2\chi^2.
\end{aligned}
 \label{eq:angular-shortfall-charge}
\end{equation}
Also, \(0<h\leq s\) implies
\[
 h^2\norm{\gtilde}^2\leq s^2\norm{\gtilde}^2=\chi^2.
\]
After the decomposition \eqref{eq:restore-deterministic-step}, the expanded
distance inequality has two terms besides the desired
\(2s\inner{\gtilde}{\Log_{\xhat}(\bm y)}\): the error and \(h^2\norm{\gtilde}^2\).
Adding their bounds and using \(\dist(\xhat,\bm y)<2\thetacap\) gives
\[
 \frac{\dist(\xhat,\bm y)}2\chi^2+\chi^2
 =\left(1+\frac{\dist(\xhat,\bm y)}2\right)\chi^2
 \leq(1+\thetacap)s^2\norm{\gtilde}^2.
\]
Substitution into the squared-distance bound at the start of the proof gives
\eqref{eq:normalized-update-distance}.
\end{proof}

For the stochastic update, \(h\) depends on the sampled gradient norm, so conditional
unbiasedness does not directly simplify \(\E[h\gtilde_t\mid\mathcal F_t]\).
\Cref{lem:normalized-update-distance} absorbs this difference into the error term, producing a
distance inequality whose inner-product term has the deterministic coefficient \(s_t\).
\begin{proposition}{Expected one-step progress}{chord-one-step}
Under \Cref{ass:oracle}, let \(\bm y\in\mathcal C\) be \(\mathcal F_t\)-measurable. Then
\begin{align}
 s_t\E[L(\xhat_t)-L(\bm y)]
 \leq{}&\frac12\E\bigl[\dist(\xhat_t,\bm y)^2-
                         \dist(\xhat_{t+1},\bm y)^2\bigr] \nonumber\\
 &+\frac{1+\thetacap}{2}s_t^2\E\norm{\gtilde_t}^2.
 \label{eq:chord-one-step}
\end{align}
\end{proposition}
\begin{proof}
\emph{Step 1: deterministic distance bound.}
Apply \Cref{lem:normalized-update-distance} with
\(\xhat=\xhat_t\), \(\gtilde=\gtilde_t\), and \(s=s_t\).
The normalized endpoint is exactly \(\xhat_{t+1}\), so, on every sample path,
\[
\begin{aligned}
 \dist(\xhat_{t+1},\bm y)^2
 \leq{}&\dist(\xhat_t,\bm y)^2
       +2s_t\inner{\gtilde_t}{\Log_{\xhat_t}(\bm y)}\\
 &+(1+\thetacap)s_t^2\norm{\gtilde_t}^2.
\end{aligned}
\]

\emph{Step 2: conditional expectation and unbiasedness.}
The points \(\xhat_t,\bm y\) and the Log vector are \(\mathcal F_t\)-measurable,
while \(s_t\) is deterministic. The Log vector has norm at most \(2\thetacap\),
so Cauchy--Schwarz and the second-moment bound make the inner product integrable.
Taking conditional expectation term by term gives
\[
\begin{aligned}
 \E[\dist(\xhat_{t+1},\bm y)^2\mid\mathcal F_t]
 \leq{}&\dist(\xhat_t,\bm y)^2
       +\E[2s_t\inner{\gtilde_t}{\Log_{\xhat_t}(\bm y)}\mid\mathcal F_t]\\
 &+(1+\thetacap)s_t^2\E[\norm{\gtilde_t}^2\mid\mathcal F_t].
\end{aligned}
\]
Pulling the \(\mathcal F_t\)-measurable Log vector out of the inner product
and applying conditional unbiasedness gives
\begin{equation}
\begin{aligned}
 \E[2s_t\inner{\gtilde_t}{\Log_{\xhat_t}(\bm y)}\mid\mathcal F_t]
 &=2s_t\inner{\E[\gtilde_t\mid\mathcal F_t]}{\Log_{\xhat_t}(\bm y)}\\
 &=2s_t\inner{\nabla_{\mathbb S}L(\xhat_t)}{\Log_{\xhat_t}(\bm y)}.
\end{aligned}
 \label{eq:conditional-descent-pairing}
\end{equation}
Insertion into the conditional distance bound gives
\[
\begin{aligned}
 \E[\dist(\xhat_{t+1},\bm y)^2\mid\mathcal F_t]
 \leq{}&\dist(\xhat_t,\bm y)^2
       +2s_t\inner{\nabla_{\mathbb S}L(\xhat_t)}{\Log_{\xhat_t}(\bm y)}\\
 &+(1+\thetacap)s_t^2\E[\norm{\gtilde_t}^2\mid\mathcal F_t].
\end{aligned}
\]

\emph{Step 3: geodesic convexity.}
By geodesic convexity, \Cref{lem:convex-support} gives
\[
 \inner{\nabla_{\mathbb S}L(\xhat_t)}{\Log_{\xhat_t}(\bm y)}
 \leq L(\bm y)-L(\xhat_t).
\]
Substituting and multiplying through by \(2s_t>0\) gives
\[
\begin{aligned}
 \E[\dist(\xhat_{t+1},\bm y)^2\mid\mathcal F_t]
 \leq{}&\dist(\xhat_t,\bm y)^2-2s_t[L(\xhat_t)-L(\bm y)]\\
 &+(1+\thetacap)s_t^2\E[\norm{\gtilde_t}^2\mid\mathcal F_t].
\end{aligned}
\]
Rearranging:
\[
\begin{aligned}
 2s_t[L(\xhat_t)-L(\bm y)]
 \leq{}&\dist(\xhat_t,\bm y)^2
       -\E[\dist(\xhat_{t+1},\bm y)^2\mid\mathcal F_t]\\
 &+(1+\thetacap)s_t^2\E[\norm{\gtilde_t}^2\mid\mathcal F_t].
\end{aligned}
\]
\emph{Step 4: total expectation.}
Taking total expectation, with \(s_t\) deterministic, gives
\[
\begin{aligned}
 2s_t\E[L(\xhat_t)-L(\bm y)]
 \leq{}&\E\dist(\xhat_t,\bm y)^2
       -\E\!\left[\E[\dist(\xhat_{t+1},\bm y)^2\mid\mathcal F_t]\right]\\
 &+(1+\thetacap)s_t^2
       \E\!\left[\E[\norm{\gtilde_t}^2\mid\mathcal F_t]\right].
\end{aligned}
\]
The tower property collapses each nested conditional expectation.
Dividing by two gives
\[
\begin{aligned}
 s_t\E[L(\xhat_t)-L(\bm y)]
 \leq{}&\frac12\E\bigl[\dist(\xhat_t,\bm y)^2
                          -\dist(\xhat_{t+1},\bm y)^2\bigr]\\
 &+\frac{1+\thetacap}{2}s_t^2\E\norm{\gtilde_t}^2,
\end{aligned}
\]
which is \eqref{eq:chord-one-step}.
\end{proof}

\subsection{From Tail Averages to the Compact Last-Iterate Bound}
\label[appendix]{app:last-iterate-proof}
We follow the last-iterate argument of
\citet[Lemma~5 and Theorem~10]{defazio2023linear},
also used by \citet[Lemma~D.1 and Theorem~E.2]{schaipp2025surprising},
with \Cref{prop:chord-one-step} supplying the spherical one-step inequality.

\paragraph{Proof idea.}
\Cref{prop:chord-one-step} bounds expected progress at each step.
Summing those bounds and telescoping the squared distances gives a bound on the
weighted-average expected loss, but \Cref{thm:last-iterate} needs the expected loss
at the last iterate alone. The average could be small because early iterates are
good, even if the final one is not. We therefore remove the earlier entries one at
a time, renormalizing the remaining weights, until the average contains only the
last iterate. Removing an entry with expected loss below the remaining average
raises that average, so we must allow the upper bound to increase accordingly.

To bound this increase when removing \(\xhat_k\), we sum
\Cref{prop:chord-one-step} over \(t=k,\ldots,T\), choosing \(\bm y=\xhat_k\)
throughout. This compares the weighted-average expected loss with the expected
loss of the entry being removed. The resulting bound limits how much removing
that entry can raise the average. The last iterate's expected loss equals the
original average plus the successive changes in the average. Its upper bound is
therefore the bound on the original average plus the sum of the upper bounds on
those changes.

Recall \(D=\dist(\xhat_1,\xhat^\star)\). Write \(s_t=M\omega_t\) as in
\eqref{eq:schedule-mass-profile}, and define the remaining normalized sum
\begin{equation}
 W_k:=\sum_{t=k}^T\omega_t.
 \label{eq:proof-allocation-tail}
\end{equation}
Thus \(W_1=1\), every \(W_k\) is positive, and
\(W_k-W_{k+1}=\omega_k\).

\begin{proof}[Proof of \Cref{thm:last-iterate}]
\emph{Step 1: sum one-step progress on an arbitrary tail.}
Let \(\bm y\) be a cap-valued comparator with integrable loss that is
\(\mathcal F_k\)-measurable. It is then known before every draw on the tail \(t\geq k\).
Summing \eqref{eq:chord-one-step} gives
\begin{equation}
\begin{aligned}
 M\sum_{t=k}^T \omega_t\E[L(\xhat_t)-L(\bm y)]
 &\leq\frac12\sum_{t=k}^T
  \left(\E\dist(\xhat_t,\bm y)^2-\E\dist(\xhat_{t+1},\bm y)^2\right)\\
 &\quad+\frac{(1+\thetacap)M^2}{2}
                  \sum_{t=k}^T \omega_t^2\E\norm{\gtilde_t}^2.
\end{aligned}
\end{equation}
Consecutive distance terms cancel exactly:
\[
 \sum_{t=k}^T
 \left(\E\dist(\xhat_t,\bm y)^2-\E\dist(\xhat_{t+1},\bm y)^2\right)
 =\E\dist(\xhat_k,\bm y)^2-\E\dist(\xhat_{T+1},\bm y)^2.
\]
Using the unconditional bound \(\E\norm{\gtilde_t}^2\leq G^2\), we obtain
\begin{equation}
\begin{aligned}
 M\sum_{t=k}^T \omega_t\E[L(\xhat_t)-L(\bm y)]
 &\leq\frac12\E\dist(\xhat_k,\bm y)^2
       -\frac12\E\dist(\xhat_{T+1},\bm y)^2\\
 &\quad+\frac{(1+\thetacap)G^2M^2}{2}\sum_{t=k}^T \omega_t^2\\
 &\leq\frac12\E\dist(\xhat_k,\bm y)^2
       +\frac{(1+\thetacap)G^2M^2}{2}\sum_{t=k}^T \omega_t^2.
\end{aligned}
\label{eq:tail-bound}
\end{equation}
Starting at \(\xhat_1\), the first \(T-1\) updates take us to \(\xhat_T\), which is
the iterate we care about. But \Cref{prop:chord-one-step} bounds the loss at the
point before an update. So, to include the loss at \(\xhat_T\), we apply it once more,
using \(s_T\), and introduce the next point \(\xhat_{T+1}\). This extra update is
only part of the proof; it does not change \(\xhat_T\). Once we sum the inequalities,
the term involving \(\xhat_{T+1}\) has a minus sign, so we can drop it when taking an
upper bound. \textcolor{black}{The coefficient \(s_T\) remains in the bound. We can choose it in
advance to improve the bound subject to the theorem's assumptions, including
containment of \(\xhat_{T+1}\).}
The left side of \eqref{eq:tail-bound} is still a weighted sum; division by its tail
coefficient sum will turn it into an average.

This step is the geodesic counterpart of Lemma E.1 of \citet{schaipp2025surprising}.

\emph{Step 2: compare the full average to the optimum.}
Let \(b_t\) denote the expected excess loss at step \(t\), and let \(A_k\) denote its
weighted average over the tail beginning at \(k\):
\begin{equation}
 b_t:=\E[L(\xhat_t)-L(\xhat^\star)],\qquad
 A_k:=\frac{\sum_{t=k}^T \omega_tb_t}{W_k}.
\end{equation}
Taking \(k=1\) and
\(\bm y=\xhat^\star\) in \eqref{eq:tail-bound}, then dividing by \(M>0\), gives
\begin{equation}
 A_1=\sum_{t=1}^T \omega_tb_t
 \leq\frac{D^2}{2M}
 +\frac{(1+\thetacap)G^2M}{2}\sum_{t=1}^T \omega_t^2.
 \label{eq:full-average-bound}
\end{equation}
This controls the full average. The remaining steps account for the change in that average
when early entries are removed.

\emph{Step 3: express the endpoint through changes of tail average.}
The final tail has \(A_T=b_T\). Removing its first entry from tail \(k\) removes
\(\omega_kb_k\) from the weighted sum and \(\omega_k\) from its coefficient sum, so
\[
 W_{k+1}A_{k+1}=W_kA_k-\omega_kb_k.
\]
For \(k<T\), dividing and subtracting \(A_k\) yields
\begin{equation}
\begin{aligned}
 A_{k+1}-A_k
 &=\frac{W_kA_k-\omega_kb_k-W_{k+1}A_k}{W_{k+1}}\\
 &=\frac{(W_k-W_{k+1})A_k-\omega_kb_k}{W_{k+1}}
 =\frac{\omega_k}{W_{k+1}}(A_k-b_k).
\end{aligned}
\end{equation}
Here \(A_{k+1}-A_k\) is the change in the weighted average when its first entry is removed,
while \(A_k-b_k\) compares the current tail average with its first entry.
The factor \(\omega_k/W_{k+1}\) is the removed fraction divided by the
remaining normalized sum.
Our target is \(b_T=\E[L(\xhat_T)-L(\xhat^\star)]\). The last tail contains only
this entry, so \(A_T=b_T\). We therefore express the target as the full average plus
the accumulated changes from removing earlier entries. Telescoping gives the weighted identity of
\citet[Lemma~5]{defazio2023linear}, reproduced by \citet[Lemma~D.1]{schaipp2025surprising}:
\begin{equation}
\begin{aligned}
 b_T=A_T
 &=A_1+\sum_{k=1}^{T-1}(A_{k+1}-A_k)\\
 &=A_1 + \sum_{k=1}^{T-1}\frac{\omega_k}{W_{k+1}}(A_k-b_k).
\end{aligned}
\label{eq:weighted-identity}
\end{equation}

\emph{Step 4: compare each tail to its own starting iterate.}
The identity \eqref{eq:weighted-identity} requires an upper bound on \(A_k-b_k\).
To obtain it, fix a tail starting at \(k\) and choose \(\bm y=\xhat_k\) in
\eqref{eq:tail-bound}. This comparator lies in \(\mathcal C\) and is
\(\mathcal F_k\)-measurable, as Step~1 requires; it remains to check that its loss
is integrable. Since \(\xhat^\star\) minimizes \(L\) on \(\mathcal C\),
\(L(\xhat_k)-L(\xhat^\star)\geq0\). For an upper bound, apply
\Cref{lem:convex-support} with \(\xhat=\xhat_k\) and \(\bm y=\xhat^\star\):
\[
 L(\xhat^\star)\geq L(\xhat_k)
 +\inner{\nabla_{\mathbb S}L(\xhat_k)}{\Log_{\xhat_k}(\xhat^\star)}.
\]
{\color{black}%
Both points lie in the cap, so the logarithm vector has norm
\(\dist(\xhat_k,\xhat^\star)<2\thetacap\).
Rearranging and using Cauchy--Schwarz gives
}%
\[
\begin{aligned}
 0\leq L(\xhat_k)-L(\xhat^\star)
 &\leq-\inner{\nabla_{\mathbb S}L(\xhat_k)}{\Log_{\xhat_k}(\xhat^\star)}\\
 &\leq\norm{\nabla_{\mathbb S}L(\xhat_k)}\,\dist(\xhat_k,\xhat^\star)
 \mathrel{\color{black}\leq}{\color{black}2\thetacap\norm{\nabla_{\mathbb S}L(\xhat_k)}.}
\end{aligned}
\]
It remains to bound the expected gradient norm. By conditional unbiasedness,
\(\nabla_{\mathbb S}L(\xhat_k)=\E[\gtilde_k\mid\mathcal F_k]\). The norm is convex,
so the conditional Jensen inequality gives
\[
 \norm{\nabla_{\mathbb S}L(\xhat_k)}
 =\norm{\E[\gtilde_k\mid\mathcal F_k]}
 \leq\E[\norm{\gtilde_k}\mid\mathcal F_k].
\]
Taking expectations and applying Jensen's inequality to the concave square root,
\[
 \E\norm{\nabla_{\mathbb S}L(\xhat_k)}
 \leq\E\norm{\gtilde_k}
 \leq\sqrt{\E\norm{\gtilde_k}^2}
 \leq G.
\]
Together with the previous display,
\textcolor{black}{\(0\leq\E[L(\xhat_k)-L(\xhat^\star)]\leq2\thetacap G<\infty\)}, so \(L(\xhat_k)\)
is integrable.
The same comparator is used for every step \(t=k,\ldots,T\).
Substitution gives
\[
\begin{aligned}
 M\sum_{t=k}^T \omega_t\E[L(\xhat_t)-L(\xhat_k)]
 &\leq\frac12\E\dist(\xhat_k,\xhat_k)^2
       +\frac{(1+\thetacap)G^2M^2}{2}\sum_{t=k}^T \omega_t^2\\
 &=\frac{(1+\thetacap)G^2M^2}{2}\sum_{t=k}^T \omega_t^2.
\end{aligned}
\]
The initial distance is zero because the comparator is the tail's starting iterate.
Dividing by the positive quantity \(MW_k\) yields
\[
 \frac{\sum_{t=k}^T \omega_t\E[L(\xhat_t)-L(\xhat_k)]}{W_k}
 \leq\frac{(1+\thetacap)G^2M}{2W_k}\sum_{t=k}^T \omega_t^2.
\]
To identify the left side, cancel the common optimum loss:
\[
\begin{aligned}
  \E[L(\xhat_t)-L(\xhat_k)] 
  &=\E[L(\xhat_t)-L(\xhat^\star)]-\E[L(\xhat_k)-L(\xhat^\star)]\\
  &=b_t-b_k.
\end{aligned}
\]
Since \(\sum_{t=k}^T \omega_t=W_k\), subtracting \(b_k\) from the tail average gives
\begin{equation}
\begin{aligned}
 A_k-b_k
 &=\frac{\sum_{t=k}^T \omega_tb_t}{W_k}-\frac{W_kb_k}{W_k}\\
 &=\frac{\sum_{t=k}^T \omega_t(b_t-b_k)}{W_k}\\
 &=\frac{\sum_{t=k}^T \omega_t\E[L(\xhat_t)-L(\xhat_k)]}{W_k}
 \leq\frac{(1+\thetacap)G^2M}{2W_k}\sum_{t=k}^T \omega_t^2.
\end{aligned}
\label{eq:tail-start-bound}
\end{equation}
This is the comparison with the starting iterate used by
\citet[App.~F, derivation preceding Corollary~11]{defazio2023linear},
here obtained from \Cref{prop:chord-one-step}.

\emph{Step 5: insert the full-average and tail-start bounds.}
Substitute \eqref{eq:full-average-bound} and \eqref{eq:tail-start-bound} into
\eqref{eq:weighted-identity}, retaining each positive multiplier:
\begin{equation}
\begin{aligned}
 \E[L(\xhat_T)-L(\xhat^\star)]
 &=A_1+\sum_{k=1}^{T-1}\frac{\omega_k}{W_{k+1}}(A_k-b_k)\\
 &\leq\frac{D^2}{2M}
   +\frac{(1+\thetacap)G^2M}{2}\sum_{t=1}^T \omega_t^2\\
 &\quad+\sum_{k=1}^{T-1}\frac{\omega_k}{W_{k+1}}
       \left(\frac{(1+\thetacap)G^2M}{2W_k}
             \sum_{t=k}^T \omega_t^2\right)\\
 &=\frac{D^2}{2M}+\frac{(1+\thetacap)G^2M}{2}
   \left[\sum_{t=1}^T \omega_t^2
   +\sum_{k=1}^{T-1}\frac{\omega_k}{W_{k+1}}
                         \frac{\sum_{t=k}^T \omega_t^2}{W_k}\right].
\end{aligned}
\end{equation}
This step is the geodesic counterpart of Theorem E.2 of \citet{schaipp2025surprising}. 

\emph{Step 6: express the bound using \(C_T(\omega_{1:T})\).}
We now show that the bracket in Step~5 equals \(C_T(\omega_{1:T})\). Each squared
coefficient appears in several sums. Writing these sums out makes their
occurrences explicit (the displayed expansion is for \(T\geq3\)):
\[
\begin{aligned}
 &\sum_{t=1}^T\omega_t^2
 +\sum_{k=1}^{T-1}\frac{\omega_k}{W_kW_{k+1}}
                   \sum_{t=k}^T\omega_t^2\\
 ={}&(\omega_1^2+\omega_2^2+\cdots+\omega_T^2)\\
 &+\frac{\omega_1}{W_1W_2}
        (\omega_1^2+\omega_2^2+\cdots+\omega_T^2)\\
 &+\frac{\omega_2}{W_2W_3}(\omega_2^2+\cdots+\omega_T^2)\\
 &+\cdots+
   \frac{\omega_{T-1}}{W_{T-1}W_T}(\omega_{T-1}^2+\omega_T^2).
\end{aligned}
\]
For example, \(\omega_1^2\) occurs in the first two lines of the expanded
expression, whereas \(\omega_2^2\) also occurs in the next line. Collecting
their multipliers and continuing in the same way gives
\[
\begin{aligned}
 ={}&\omega_1^2\left(1+\frac{\omega_1}{W_1W_2}\right)\\
 &+\omega_2^2\left(1+\frac{\omega_1}{W_1W_2}
                    +\frac{\omega_2}{W_2W_3}\right)\\
 &+\cdots+
 \omega_T^2\left(1+\frac{\omega_1}{W_1W_2}
 +\cdots+\frac{\omega_{T-1}}{W_{T-1}W_T}\right).
\end{aligned}
\]
For every \(T\geq2\) and \(t<T\), the coefficient \(\omega_t^2\) occurs in the
initial sum and in the sums starting at \(k=1,\ldots,t\). Its multiplier is therefore
\[
 1+\sum_{k=1}^{t}\frac{\omega_k}{W_kW_{k+1}}.
\]
The final coefficient \(\omega_T^2\) occurs in every sum, but the outer sum
stops at \(T-1\). Its multiplier is
\[
 1+\sum_{k=1}^{T-1}\frac{\omega_k}{W_kW_{k+1}}.
\]

To simplify these multipliers, recall that
\[
 W_k=\sum_{j=k}^T\omega_j
 =\omega_k+\sum_{j=k+1}^T\omega_j
 =\omega_k+W_{k+1}.
\]
Hence
\[
\begin{aligned}
 \frac{\omega_k}{W_kW_{k+1}}
 &=\frac{W_k-W_{k+1}}{W_kW_{k+1}}\\
 &=\frac{W_k}{W_kW_{k+1}}-\frac{W_{k+1}}{W_kW_{k+1}}\\
 &=\frac1{W_{k+1}}-\frac1{W_k}.
\end{aligned}
\]
For \(t<T\), substitution gives
\[
\begin{aligned}
 1+\sum_{k=1}^{t}\frac{\omega_k}{W_kW_{k+1}}
 &=1+\sum_{k=1}^{t}\left(\frac1{W_{k+1}}-\frac1{W_k}\right)\\
 &=1+\frac1{W_{t+1}}-\frac1{W_1}\\
 &=\frac1{W_{t+1}}.
\end{aligned}
\]
The intermediate reciprocal terms cancel, and the last equality uses \(W_1=1\).
For the final coefficient, the same substitution with upper limit \(T-1\) gives
\[
\begin{aligned}
 1+\sum_{k=1}^{T-1}\frac{\omega_k}{W_kW_{k+1}}
 &=1+\sum_{k=1}^{T-1}\left(\frac1{W_{k+1}}-\frac1{W_k}\right)\\
 &=1+\frac1{W_T}-\frac1{W_1}\\
 &=\frac1{W_T}.
\end{aligned}
\]
Substituting the multipliers back into the collected expression shows that the
bracket in Step~5 equals
\[
\begin{aligned}
 \sum_{t=1}^{T-1}\frac{\omega_t^2}{W_{t+1}}+\frac{\omega_T^2}{W_T}
 &=\sum_{t=1}^{T-1}\frac{\omega_t^2}{\sum_{j=t+1}^T\omega_j}+\omega_T\\
 &=C_T(\omega_{1:T}),
\end{aligned}
\]
where \(W_T=\omega_T\). For \(T=1\), the bracket is simply
\(\omega_1^2=1=C_1(\omega_{1:1})\). Thus Step~5 becomes
\[
 \E[L(\xhat_T)-L(\xhat^\star)]
 \leq\frac{D^2}{2M}
 +\frac{(1+\thetacap)G^2M}{2}C_T(\omega_{1:T}),
\]
which is the claimed bound \eqref{eq:chord-last-iterate}.

Finally, using \(s_t=M\omega_t\), the same cost can be written in the original
coefficients:
\begin{equation}
\begin{aligned}
 M C_T(\omega_{1:T})
 &=M\omega_T+M\sum_{k=1}^{T-1}
        \frac{\omega_k^2}{\sum_{t=k+1}^T\omega_t}\\
 &=s_T+\sum_{k=1}^{T-1}
        \frac{M^2\omega_k^2}{M\sum_{t=k+1}^T\omega_t}\\
 &=s_T+\sum_{k=1}^{T-1}
        \frac{s_k^2}{\sum_{t=k+1}^Ts_t}.
\end{aligned}
 \label{eq:raw-normalized-cost-bridge}
\end{equation}

\end{proof}

The scalar last-iterate conversion above is the general-schedule argument of
\citet[Theorem~10]{defazio2023linear} and
\citet[Theorem~3.1 and App.~E]{schaipp2025surprising}.
The bound \eqref{eq:chord-last-iterate} depends on coefficient sum \(M\) and
normalized schedule \(\omega_{1:T}\) separately and the next subsection optimizes
\(M\) at fixed \(\omega_{1:T}\).

\subsection{Separating the Coefficient Sum from the Normalized Schedule}
\label[appendix]{app:optimal-mass}
Changing the schedule changes the visited points, so moments measured on one trajectory
do not by themselves bound those on another one; as remarked by \citet[Remark~3.2]{schaipp2025surprising}, we therefore need to use a common oracle envelope \(G\).

Every positive coefficient sequence has the unique decomposition
\[
 M=\sum_{t=1}^Ts_t,\qquad \omega_t=\frac{s_t}{M},\qquad
 s_t=M\omega_t,\qquad \sum_{t=1}^T\omega_t=1.
\]

\begin{proof}[Proof of \Cref{prop:optimal-mass}]
Fix a positive normalized schedule \(\omega_{1:T}\) and \(D,G>0\).
The preceding identities transform the bound into
\[
 \Omega_T(M,\omega_{1:T})
 =\frac{D^2}{2M}+\frac{(1+\thetacap)G^2}{2}M C_T(\omega_{1:T}).
\]
Here only \(M>0\) varies; the schedule and its cost are fixed. Differentiating gives
\begin{equation}
\begin{aligned}
 \frac{\mathrm d}{\mathrm dM}\Omega_T(M,\omega_{1:T})
 &=-\frac{D^2}{2M^2}+\frac{(1+\thetacap)G^2}{2}C_T(\omega_{1:T}),\\
 \frac{\mathrm d^2}{\mathrm dM^2}\Omega_T(M,\omega_{1:T})
 &=\frac{D^2}{M^3}>0.
\end{aligned}
\end{equation}
Setting the first derivative to zero yields
\[
 M^2=\frac{D^2}{(1+\thetacap)G^2C_T(\omega_{1:T})},\qquad
 M^\star(\omega_{1:T})=\frac{D}{G\sqrt{(1+\thetacap)C_T(\omega_{1:T})}}.
\]
Strict convexity makes this the
unique global minimum. The first objective term diverges as \(M\downarrow0\),
and the second diverges as \(M\to\infty\), so neither boundary supplies a competing
minimum. Substitution into the two terms separately gives
\begin{equation}
\begin{aligned}
 \frac{D^2}{2M^\star}
 &=\frac{D^2G\sqrt{(1+\thetacap)C_T(\omega_{1:T})}}{2D}
 =\frac{DG}{2}\sqrt{(1+\thetacap)C_T(\omega_{1:T})},\\
 \frac{(1+\thetacap)G^2}{2}M^\star C_T(\omega_{1:T})
 &=\frac{(1+\thetacap)G^2D C_T(\omega_{1:T})}
        {2G\sqrt{(1+\thetacap)C_T(\omega_{1:T})}}
 =\frac{DG}{2}\sqrt{(1+\thetacap)C_T(\omega_{1:T})}.
\end{aligned}
\end{equation}
Adding these equal contributions proves the minimized bound is
\(DG\sqrt{(1+\thetacap)C_T(\omega_{1:T})}\).
\end{proof}

\input{appendix-schedules.tex}
\input{appendix-witness.tex}
\input{appendix-hyperball.tex}

%% file: related-work.tex
\section{Related Work}
\label[appendix]{sec:related-work}

\paragraph{Finite-horizon learning-rate scheduling.}
\citet{jain2019makingiteratesgdinformation} design schedules that give the final SGD iterate the optimal convergence rate without averaging iterates.
\citet{defazio2023linear} develop a last-iterate theory of learning-rate scheduling for stochastic convex optimization: a regret-based formulation yields linear decay with the optimal worst-case convergence rate, while a complementary general-schedule bound allows for arbitrary positive learning-rate sequences.
\citet{zamani2025exact} show that linear decay gives the smallest possible
worst-case error for subgradient methods with Euclidean-convex losses.
Their results also cover normalized subgradient directions followed by
projection onto a convex set.
\citet{schaipp2025surprising} build on the analysis of \citet{defazio2023linear} and show that its predictions closely reproduce several empirical features of large-model training, including cosine--WSD behavior, cooldown dynamics, and learning-rate transfer.
All these last-iterate guarantees assume Euclidean-convex losses.
However, a finite loss on the full parameter space that is both Euclidean convex and scale invariant must be constant (\Cref{app:si-witness}). Instead, we keep parameters on the sphere by taking tangent steps and normalizing the updated parameters. Assuming the loss is convex along spherical arcs (geodesic convexity), we derive a last-iterate bound for arbitrary rotational schedules, in a setting that admits nontrivial scale-invariant losses. On Riemannian manifolds, \citet{bonnabel2013sgd} proves asymptotic convergence of stochastic gradient methods, including retraction-based updates. \citet{zhangSra2016gconvex} give non-asymptotic rates for geodesically convex optimization, and their stochastic bounds hold for averaged iterates. \citet{schaipp2025surprising} solve the schedule-optimization problem only numerically, finding a linear-decay profile, but they leave global optimality open. We instead solve the functional analytically, proving uniqueness, an exact recursion, and convergence to linear decay.

\paragraph{Scale invariance and rotational dynamics.}
For scale-invariant groups, parameter norms regulate rotational speed even though they do not directly affect the loss \citep{vanLaarhoven2017,arora2018theoretical,liArora2020,heo2021adamp}.
\citet{wan2021smd} formalize these coupled dynamics for SGD and momentum-SGD, identifying angular updates as the intrinsic measure of optimization progress; \citet{roburin2022spherical} characterize the spherical geometry of generic first-order optimizers.
More recently, \citet{kosson2024rotational} analyze stationary rotational equilibria, showing that direct rotational control can reproduce key effects of weight decay.
These results describe how effective rotation arises. We study instead how a finite budget of rotational coefficients should be distributed \textit{over time} under the stated spherical bound.

\paragraph{On-sphere optimization and angular control.}
Constraining optimization to the sphere has served both as an analytical tool to isolate directional dynamics under fixed effective learning rates \citep{NEURIPS2022_5aea56ee} and, more recently, as a way to place rotational motion under explicit control. In particular, Hyperball \citep{wen2026hyperball} and Magnitude--Direction Decoupling \citep{haegele2026md} fix the sphere radius and expose directional step control; related directional controls appear in Muon-style optimizers \citep{lion2026muown,huebler2026angularmuown}.
Empirical pretraining studies report effects of matching or transferring effective rotational trajectories on the loss \citep{liu2026elr,xiao2026hyperballfreelunch,yuan2026hypertransfer}.
\citet[Sec.~4.1.5]{haegele2026md} find linear decay better than the tested WSD variants under magnitude--direction decoupling and attribute the difference to the absence of norm-induced annealing on the sphere. Our framework connects such norm and directional dynamics with finite-horizon schedule design. We test its predictive power through modeled rotational profiles, schedule comparisons with and without norm control, and schedules derived from the theory. The predictions recover the observed WSD--cosine reversal, and the derived schedules improve on standard profiles.

%% file: appendix-schedules.tex
\section{Proofs for Schedule Optimization}
\label[appendix]{app:schedule-optimization}
\Cref{app:optimal-mass} reduced the schedule-design problem to minimizing the
schedule-shape cost \(C_T(\omega_{1:T})\) over positive normalized schedules.
\Cref{sec:schedule-design} stated the resulting optimality and convergence results;
we now supply their full proofs, together with the exact cost of the prescribed
linear schedule used in the asymptotic comparison.

\subsection{Exact Finite-Horizon Optimization}
\label[appendix]{app:exact-schedule-optimization}
{\color{black}%
Once \(\omega_1\) is chosen, how should the remaining sum \(1-\omega_1\) be
distributed among the later coefficients? Dividing those coefficients by their
sum turns them into a shorter normalized schedule. We show how its cost enters
\(C_n\), use the shorter optimizer to determine \(\omega_1\), and repeat this
choice to construct the full schedule.
}%

\begin{proof}[Proof of \Cref{lem:first-step-decomposition}]
{\color{black}%
Let us first consider an example to see how the remaining entries' contribution
relates to the cost of a shorter normalized schedule. For a positive normalized
schedule with two entries, \eqref{eq:shape-score} gives
\[
 \begin{gathered}
 C_2(\omega_1,\omega_2)=\frac{\omega_1^2}{\omega_2}+\omega_2,\\
 \omega_1,\omega_2>0,\qquad \omega_1+\omega_2=1.
 \end{gathered}
\]
Now consider a positive normalized schedule with three entries, so that
\(\omega_1+\omega_2+\omega_3=1\). Its cost is
\[
 C_3(\omega_{1:3})
 =\frac{\omega_1^2}{1-\omega_1}
  +\frac{\omega_2^2}{\omega_3}+\omega_3.
\]
After separating the first contribution, the remaining two terms have the
same form as the displayed two-entry cost. Their coefficients sum to
\[
 \omega_2+\omega_3=1-\omega_1<1,
\]
whereas the two-entry optimization problem requires a sum of one. To apply
that shorter problem, we must first normalize the remaining entries.

Dividing both entries by their sum preserves their relative proportions
and gives a normalized schedule. Define
\[
 \widetilde\omega_t:=\frac{\omega_t}{\omega_2+\omega_3}
 =\frac{\omega_t}{1-\omega_1},\qquad t=2,3.
\]
The tilde denotes the original remaining coefficients rescaled by their own
sum. This sum is positive because \(\omega_2,\omega_3>0\), so both
\(\widetilde\omega_2\) and \(\widetilde\omega_3\) are positive. Moreover,
\[
 \widetilde\omega_2+\widetilde\omega_3
 =\frac{\omega_2+\omega_3}{\omega_2+\omega_3}=1.
\]
Rearranging the definition gives
\(\omega_t=(1-\omega_1)\widetilde\omega_t\) for \(t=2,3\).

To compare the costs, we now determine how this rescaling changes the
remaining contribution. Substituting these expressions gives
\[
\begin{aligned}
 \frac{\omega_2^2}{\omega_3}+\omega_3
 &=\frac{(1-\omega_1)^2\widetilde\omega_2^2}{(1-\omega_1)\widetilde\omega_3}
   +(1-\omega_1)\widetilde\omega_3\\
 &=(1-\omega_1)\left(\frac{\widetilde\omega_2^2}{\widetilde\omega_3}+\widetilde\omega_3\right)\\
 &=(1-\omega_1)C_2(\widetilde\omega_{2:3}).
\end{aligned}
\]
Substituting this result into the expression for \(C_3\) yields
\[
 C_3(\omega_{1:3})
 =\frac{\omega_1^2}{1-\omega_1}
  +(1-\omega_1)C_2(\widetilde\omega_{2:3}).
\]

For a general length \(n\geq2\), separate the first contribution in
\eqref{eq:shape-score}:
\[
 C_n(\omega_{1:n})
 =\frac{\omega_1^2}{1-\omega_1}
 +\sum_{k=2}^{n-1}\frac{\omega_k^2}{\sum_{t=k+1}^n\omega_t}
 +\omega_n.
\]
The first denominator is \(1-\omega_1\) because
\(\sum_{t=2}^n\omega_t=1-\omega_1\). Positivity of the entries gives
\(0<\omega_1<1\), so we can again divide the remaining coefficients by their
sum. The normalized coefficients \(\widetilde\omega_t=\omega_t/(1-\omega_1)\)
are positive and satisfy
\[
 \sum_{t=2}^n \widetilde\omega_t
 =\frac{\sum_{t=2}^n\omega_t}{1-\omega_1}
 =\frac{1-\omega_1}{1-\omega_1}=1.
\]
Thus \(\widetilde\omega_{2:n}\) is a normalized schedule with \(n-1\) entries, retaining
the original indices \(2,\ldots,n\).

For each \(2\leq k<n\), substitute \(\omega_t=(1-\omega_1)\widetilde\omega_t\) into the
corresponding ratio:
\[
\begin{aligned}
 \frac{\omega_k^2}{\sum_{t=k+1}^n\omega_t}
 &=\frac{(1-\omega_1)^2\widetilde\omega_k^2}{(1-\omega_1)\sum_{t=k+1}^n \widetilde\omega_t}\\
 &=(1-\omega_1)\frac{\widetilde\omega_k^2}{\sum_{t=k+1}^n \widetilde\omega_t}.
\end{aligned}
\]
The squared numerator contributes two factors of \(1-\omega_1\), while the
sum in the denominator contributes one. One factor remains after cancellation.
The terminal term is \(\omega_n=(1-\omega_1)\widetilde\omega_n\), so it carries the same
factor. Collecting the terms gives
\[
\begin{aligned}
 C_n(\omega_{1:n})
 &=\frac{\omega_1^2}{1-\omega_1}
 +(1-\omega_1)\left[
 \widetilde\omega_n+\sum_{k=2}^{n-1}\frac{\widetilde\omega_k^2}{\sum_{t=k+1}^n \widetilde\omega_t}\right]\\
 &=\frac{\omega_1^2}{1-\omega_1}
 +(1-\omega_1)C_{n-1}(\widetilde\omega_{2:n}).
\end{aligned}
\]
The bracket is the cost of the normalized remainder, including its final
entry \(\widetilde\omega_n\). This proves \eqref{eq:normalized-tail-cost-scaling}.
For \(n=2\), the sum is empty and the remainder has its single entry
\(\widetilde\omega_2=1\), so its cost is \(C_1(\widetilde\omega_2)=1\).

Conversely, choose any \(\omega_1\in(0,1)\) and any positive normalized
schedule \(\widetilde\omega_{2:n}\). Setting \(\omega_t=(1-\omega_1)\widetilde\omega_t\) for \(2\leq t\leq n\)
produces positive entries with total sum
\(\omega_1+(1-\omega_1)\sum_{t=2}^n \widetilde\omega_t=1\).
Thus, for each fixed \(\omega_1\), the normalized remainder can be chosen
freely among all schedules of length \(n-1\).
}%
\end{proof}

\begin{proof}[Proof of \Cref{thm:optimal-schedule}]
{\color{black}%
We prove the result by induction on \(n\). For \(n=1\), the only normalized
schedule is \(\omega_1=1\), whose cost is \(C_1^\star=1\). Suppose now that
the length-\((n-1)\) problem has a unique positive normalized optimizer
and that \(C_{n-1}^\star>0\).

For a fixed \(\omega_1\in(0,1)\), \eqref{eq:normalized-tail-cost-scaling} gives
\[
\begin{aligned}
 C_n(\omega_{1:n})
 &=\frac{\omega_1^2}{1-\omega_1}+(1-\omega_1)C_{n-1}(\widetilde\omega_{2:n})\\
 &\geq\frac{\omega_1^2}{1-\omega_1}+(1-\omega_1)C_{n-1}^\star.
\end{aligned}
\]
The factor \(1-\omega_1\) is positive, so equality holds exactly when
\(\widetilde\omega_{2:n}\) is the unique shorter optimizer. As shown above, this remainder
gives an admissible schedule for every \(\omega_1\in(0,1)\).
We therefore only need to choose \(\omega_1\)
to minimize the expression in \eqref{eq:optimal-scalar-problem}.
}%

{\color{black}%
To differentiate that expression, rewrite it as
}%
\[
 \frac{\omega_1^2}{1-\omega_1}+(1-\omega_1)C_{n-1}^\star
 =\frac1{1-\omega_1}-2+(1+C_{n-1}^\star)(1-\omega_1).
\]
{\color{black}%
Holding \(C_{n-1}^\star\) fixed, its first and second derivatives are
}%
\[
\begin{aligned}
 \frac{\mathrm d}{\mathrm d\omega_1}
 \left[\frac1{1-\omega_1}-2+(1+C_{n-1}^\star)(1-\omega_1)\right]
 &=\frac1{(1-\omega_1)^2}-(1+C_{n-1}^\star),\\
 \frac{\mathrm d^2}{\mathrm d\omega_1^2}
 \left[\frac1{1-\omega_1}-2+(1+C_{n-1}^\star)(1-\omega_1)\right]
 &=\frac2{(1-\omega_1)^3}>0.
\end{aligned}
\]
{\color{black}%
Setting the first derivative to zero gives
\[
 \frac1{(1-\omega_1)^2}=1+C_{n-1}^\star
 \quad\Longrightarrow\quad
 1-\omega_1=\frac1{\sqrt{1+C_{n-1}^\star}},
\]
where we take the positive square root because \(1-\omega_1>0\).
The minimizing coefficient is therefore
\begin{equation}
 \omega_1=1-\frac1{\sqrt{1+C_{n-1}^\star}}.
 \label{eq:optimal-first-coefficient}
\end{equation}
}%
Since \(C_{n-1}^\star>0\), we have
\(0<1/\sqrt{1+C_{n-1}^\star}<1\). Thus the stationary point lies in
{\color{black}%
the interval \(0<\omega_1<1\). The second derivative is positive throughout
this interval, so this is the unique global minimum.
}%

{\color{black}%
Substituting \eqref{eq:optimal-first-coefficient} into the expression in
\eqref{eq:optimal-scalar-problem} gives
\[
\begin{aligned}
 \sqrt{1+C_{n-1}^\star}-2
       +\frac{1+C_{n-1}^\star}{\sqrt{1+C_{n-1}^\star}}
 &=2\sqrt{1+C_{n-1}^\star}-2\\
 &=2\bigl(\sqrt{1+C_{n-1}^\star}-1\bigr)>0.
\end{aligned}
\]

Every admissible schedule has cost at least this value. To attain it, choose
\(\omega_1\) from \eqref{eq:optimal-first-coefficient}, take \(\widetilde\omega_{2:n}\) to be
the shorter optimizer, and set \(\omega_t=(1-\omega_1)\widetilde\omega_t\) for \(t\geq2\).
The converse construction above gives a positive normalized schedule, and
both minimizations are equalities for this choice. Hence the infimum
\(C_n^\star\) is attained and equals the value in
\eqref{eq:optimal-cost-recursion}.

Any schedule attaining this cost must have equality both when replacing
\(C_{n-1}(\widetilde\omega_{2:n})\) by \(C_{n-1}^\star\) and when minimizing over
\(\omega_1\). The first equality forces \(\widetilde\omega_{2:n}\) to be the unique shorter
optimizer; the second forces \eqref{eq:optimal-first-coefficient}.
These choices determine every entry \(\omega_t=(1-\omega_1)\widetilde\omega_t\) uniquely.
The calculated cost is positive, completing the induction.
}%
\end{proof}

\begin{proof}[Proof of \Cref{cor:optimal-allocation}]
{\color{black}%
For \(T=1\), the only schedule is \(\omega_1^\star=1\).
For \(T\geq2\), let \(\omega_{1:T}^\star\) be the unique optimizer from
\Cref{thm:optimal-schedule} and define
\[
 W_k^\star:=\sum_{t=k}^T\omega_t^\star
 \qquad(1\leq k\leq T).
\]
The remaining sum \(W_k^\star\) is the amount available before choosing
entry \(k\); in particular, \(W_1^\star=1\).

The equality conditions in the proof of \Cref{thm:optimal-schedule} show
that removing the first entry of an optimal schedule and normalizing the
remainder gives the unique optimizer of the shorter problem. Applying this
conclusion successively shows that
\[
 \left(\frac{\omega_k^\star}{W_k^\star},\ldots,
       \frac{\omega_T^\star}{W_k^\star}\right)
\]
is the optimal normalized schedule of length \(T-k+1\).
Its first entry is \(\omega_k^\star/W_k^\star\), with \(T-k\) entries
after it. To determine this first entry for \(k<T\), apply
\eqref{eq:optimal-first-coefficient} with \(n=T-k+1\). The cost index in
that formula is \(n-1=T-k\), the number of entries after the first, so
\[
 \frac{\omega_k^\star}{W_k^\star}
 =1-\frac1{\sqrt{1+C_{T-k}^\star}}.
\]
This gives the fraction of the remaining sum assigned to entry \(k\).
Multiplying by \(W_k^\star\) recovers the coefficient in the original schedule:
\[
 \omega_k^\star
 =W_k^\star\left(1-\frac1{\sqrt{1+C_{T-k}^\star}}\right).
\]

To repeat the construction at the next index, we need the sum left after
assigning \(\omega_k^\star\). Separating this entry from the remaining sum gives
\[
 W_k^\star=\omega_k^\star+\sum_{t=k+1}^T\omega_t^\star
 =\omega_k^\star+W_{k+1}^\star.
\]
Subtracting the current coefficient and substituting its formula therefore yields
\[
\begin{aligned}
 W_{k+1}^\star
 &=W_k^\star-\omega_k^\star\\
 &=W_k^\star-W_k^\star
   \left(1-\frac1{\sqrt{1+C_{T-k}^\star}}\right)\\
 &=\frac{W_k^\star}{\sqrt{1+C_{T-k}^\star}}.
\end{aligned}
\]
Dividing this update by \(W_k^\star>0\) expresses the next remaining sum
as a fraction of the current one. Together with the fraction assigned to
entry \(k\), this gives
\begin{equation}
 \frac{\omega_k^\star}{W_k^\star}
 =1-\frac1{\sqrt{1+C_{T-k}^\star}},\qquad
 \frac{W_{k+1}^\star}{W_k^\star}
 =1-\frac{\omega_k^\star}{W_k^\star}
 =\frac1{\sqrt{1+C_{T-k}^\star}}.
\label{eq:optimal-local-allocation-fraction}
\end{equation}

Starting with \(W_1^\star=1\), calculate \(\omega_k^\star\) from the current
remaining sum, then update that sum to \(W_{k+1}^\star\).
Repeat for \(k=1,\ldots,T-1\), then assign the remaining sum to the final
entry, \(\omega_T^\star=W_T^\star\).
}%
Since \(C_{T-k}^\star>0\), each step satisfies
\(0<W_{k+1}^\star<W_k^\star\), so every coefficient is positive.
Their sum is
\[
 \sum_{k=1}^{T}\omega_k^\star
 =\sum_{k=1}^{T-1}(W_k^\star-W_{k+1}^\star)+W_T^\star
 =W_1^\star=1.
\]
{\color{black}%
Every entry is determined by the unique optimal remainder, so these formulas
construct the optimizer already established in \Cref{thm:optimal-schedule}.
Finally, \(W_k^\star=1-\sum_{j=1}^{k-1}\omega_j^\star\); substituting this
into the coefficient formula gives \eqref{eq:optimal-allocation-recursion},
including its terminal entry.
}%
\end{proof}

For \(D,G>0\), combining this normalized schedule with the unique optimal coefficient sum in
\Cref{prop:optimal-mass} gives the full schedule
\[
 s_t^\star=M^\star(\omega_{1:T}^\star)\omega_t^\star.
\]
Indeed, the amplitude-optimized bound is
\(DG\sqrt{(1+\thetacap)C_T(\omega_{1:T})}\), which strictly increases with
\(C_T(\omega_{1:T})\). It therefore has the same unique minimizing schedule.

\subsection{From Optimal Cost to the Linear Profile}
\label[appendix]{app:optimal-profile-proof}
\label[appendix]{app:optimal-asymptotics}
We first prove the optimal-cost asymptotics, then use the schedule
recursion to establish convergence to linear decay.

\begin{proof}[Proof of \Cref{prop:optimal-profile-limit}]

\emph{Part 1: asymptotic optimal cost.}
We begin with \eqref{eq:optimal-cost-recursion} from
\Cref{thm:optimal-schedule}.
Rearranging and squaring gives, for \(n\geq2\),
\[
 \sqrt{1+C_{n-1}^\star}=1+\frac{C_n^\star}{2},
 \qquad
 1+C_{n-1}^\star=1+C_n^\star+\frac{(C_n^\star)^2}{4}.
\]
Hence the decrease between successive costs satisfies the exact identity
\begin{equation}
 C_{n-1}^\star=C_n^\star+\frac{(C_n^\star)^2}{4}.
 \label{eq:cost-reciprocal-increment}
\end{equation}
The costs are positive by \Cref{thm:optimal-schedule}, so this identity shows
that they strictly decrease. They therefore converge to a limit
\(\ell\geq0\). Both \(C_{n-1}^\star\) and \(C_n^\star\) converge to
\(\ell\), so taking limits in \eqref{eq:cost-reciprocal-increment} gives
\[
 \ell=\ell+\frac{\ell^2}{4}
 \quad\Longrightarrow\quad \ell^2=0
 \quad\Longrightarrow\quad \ell=0.
\]
Thus \(C_n^\star\to0\).

We now prove the sharper claim \(C_n^\star\sim4/n\). Because the costs
are positive, this claim is equivalent to
\[
 C_n^\star\sim\frac4n
 \quad\Longleftrightarrow\quad
 \frac{1/C_n^\star}{n}\longrightarrow\frac14.
\]
The expression on the right is a ratio of two sequences. We evaluate it
with the Stolz--Cesàro theorem, which states that if \(b_n\) is strictly
increasing and unbounded and
\[
 \frac{a_n-a_{n-1}}{b_n-b_{n-1}}\longrightarrow L,
\]
then \(a_n/b_n\to L\).

Apply this theorem with
\[
 a_n=\frac1{C_n^\star},\qquad b_n=n.
\]
The denominator \(b_n=n\) is strictly increasing and unbounded, and
\(b_n-b_{n-1}=1\). It remains to calculate the numerator increment.
Using \eqref{eq:cost-reciprocal-increment},
\[
\begin{aligned}
 a_n-a_{n-1}
 &=\frac1{C_n^\star}-\frac1{C_{n-1}^\star}\\
 &=\frac{C_{n-1}^\star-C_n^\star}
         {C_n^\star C_{n-1}^\star}
 =\frac{C_n^\star}{4C_{n-1}^\star}
 =\frac1{4+C_n^\star}.
\end{aligned}
\]
The last equality uses
\(C_{n-1}^\star=C_n^\star(1+C_n^\star/4)\).
Since \(C_n^\star\to0\), continuity gives
\[
 \frac{a_n-a_{n-1}}{b_n-b_{n-1}}
 =\frac1{4+C_n^\star}\longrightarrow\frac14.
\]
All hypotheses of Stolz--Cesàro are therefore satisfied, and it yields
\[
 \frac{a_n}{b_n}
 =\frac{1/C_n^\star}{n}
 =\frac1{nC_n^\star}
 \longrightarrow\frac14.
\]
This limit is positive, so taking reciprocals gives
\begin{equation}
 nC_n^\star\longrightarrow4.
 \label{eq:optimal-cost-asymptotic}
\end{equation}
In particular, setting \(n=T\) proves the optimal-cost claim of the proposition.

\emph{Part 2: convergence to linear decay.}
{\color{black}%
Fix \(0\leq\tau<1\). For each \(T\geq2\), let
\(k=1+\lfloor\tau(T-1)\rfloor\) be the selected index.
Every limit in Part~2 is taken as \(T\to\infty\), with \(\tau\) fixed.
The index \(k\) is determined separately for each \(T\):
\(k\to\infty\) when \(\tau>0\), while \(k=1\) when \(\tau=0\).
Thus \(T-k+1\) also depends on \(T\). We must prove
\[
 T\omega_k^\star\longrightarrow2(1-\tau).
\]
Recall that \(W_k^\star=\sum_{t=k}^T\omega_t^\star>0\) is the sum remaining
before entry \(k\). The coefficient \(\omega_k^\star\) is \(W_k^\star\)
multiplied by the fraction \(\omega_k^\star/W_k^\star\):
\[
 T\omega_k^\star
 =W_k^\star\left(T\frac{\omega_k^\star}{W_k^\star}\right).
\]
We first find the limit of \(T\omega_k^\star/W_k^\star\) using Part~1,
then determine the limit of \(W_k^\star\).

\emph{Step 1: evaluate \(T\omega_k^\star/W_k^\star\).}
We first express \(\omega_k^\star/W_k^\star\) using the optimal cost
\(C_{T-k+1}^\star\) for the \(T-k+1\) entries from \(k\) through \(T\).
The formula in \eqref{eq:optimal-local-allocation-fraction} applies for
\(k<T\). Since \(\tau(T-1)<T-1\) and \(T-1\) is an integer,
\(\lfloor\tau(T-1)\rfloor\leq T-2\). Hence
\(k=1+\lfloor\tau(T-1)\rfloor\leq T-1\), and therefore \(T-k+1\geq2\).

To replace \(C_{T-k}^\star\) in
\eqref{eq:optimal-local-allocation-fraction} by \(C_{T-k+1}^\star\),
write \eqref{eq:optimal-cost-recursion} as
\[
 \begin{aligned}
 C_{T-k+1}^\star
 &=2\bigl(\sqrt{1+C_{T-k}^\star}-1\bigr),\\
 \sqrt{1+C_{T-k}^\star}
 &=1+\frac{C_{T-k+1}^\star}{2}.
 \end{aligned}
\]
Replacing \(\sqrt{1+C_{T-k}^\star}\) by \(1+C_{T-k+1}^\star/2\) in
\eqref{eq:optimal-local-allocation-fraction} gives
\[
 \begin{aligned}
 \frac{\omega_k^\star}{W_k^\star}
 &=1-\frac1{\sqrt{1+C_{T-k}^\star}}
 =1-\frac1{1+C_{T-k+1}^\star/2}\\
 &=\frac{C_{T-k+1}^\star/2}{1+C_{T-k+1}^\star/2}
 =\frac{C_{T-k+1}^\star}{2+C_{T-k+1}^\star}.
 \end{aligned}
\]

Part~1 establishes \eqref{eq:optimal-cost-asymptotic}, which at horizon \(T\) reads
\[
 TC_T^\star\longrightarrow4\qquad(T\to\infty).
\]
To deduce \((T-k+1)C_{T-k+1}^\star\to4\) from
\eqref{eq:optimal-cost-asymptotic}, we must first prove
\(T-k+1\to\infty\) as \(T\to\infty\).
The growth of \(T-k+1\) needs checking because \(k\) also depends on \(T\).

To prove \(T-k+1\to\infty\), use \(k=1+\lfloor\tau(T-1)\rfloor\).
The floor inequalities and subtraction from \(T\) give
\[
 \begin{gathered}
 \tau(T-1)-1<\lfloor\tau(T-1)\rfloor\leq\tau(T-1),\\
 T-\tau(T-1)
 \leq T-k+1=T-\lfloor\tau(T-1)\rfloor<T-\tau(T-1)+1.
 \end{gathered}
\]
Expanding \(T-\tau(T-1)\) yields
\begin{equation}
 (1-\tau)T+\tau
 \leq T-k+1<(1-\tau)T+\tau+1.
 \label{eq:optimal-profile-index-bounds}
\end{equation}
Since \(1-\tau>0\), the lower bound gives
\[
 T-k+1\geq(1-\tau)T+\tau\longrightarrow\infty.
\]
We can therefore apply \eqref{eq:optimal-cost-asymptotic} to
\(C_{T-k+1}^\star\):
\begin{equation}
 (T-k+1)C_{T-k+1}^\star\longrightarrow4,\qquad
 C_{T-k+1}^\star
 =\frac{(T-k+1)C_{T-k+1}^\star}{T-k+1}\longrightarrow0.
 \label{eq:optimal-profile-tail-cost-limits}
\end{equation}

We now return to \(T\omega_k^\star/W_k^\star\).
Equation~\eqref{eq:optimal-profile-tail-cost-limits} gives the limit of
\((T-k+1)C_{T-k+1}^\star\), while \(T\omega_k^\star/W_k^\star\)
contains \(TC_{T-k+1}^\star\). Multiplying and dividing by \(T-k+1\)
expresses the target using the known cost limits:
\begin{equation}
 \begin{aligned}
 T\frac{\omega_k^\star}{W_k^\star}
 &=T\frac{C_{T-k+1}^\star}{2+C_{T-k+1}^\star}\\
 &=\frac{T}{T-k+1}
   \frac{(T-k+1)C_{T-k+1}^\star}{2+C_{T-k+1}^\star}.
 \end{aligned}
 \label{eq:optimal-profile-fraction-scaling}
\end{equation}
Equation~\eqref{eq:optimal-profile-tail-cost-limits} gives
\((T-k+1)C_{T-k+1}^\star\to4\) and \(C_{T-k+1}^\star\to0\).
Since \(2+C_{T-k+1}^\star\to2>0\), taking the quotient gives
\[
 \frac{(T-k+1)C_{T-k+1}^\star}{2+C_{T-k+1}^\star}
 \longrightarrow\frac4{2+0}=2.
\]
It remains to find \(\lim_{T\to\infty}T/(T-k+1)\). Dividing
\eqref{eq:optimal-profile-index-bounds} by \(T\) gives
\[
 1-\tau+\frac{\tau}{T}
 \leq\frac{T-k+1}{T}<1-\tau+\frac{\tau+1}{T}.
\]
For fixed \(\tau\), the lower and upper bounds satisfy
\[
 \begin{aligned}
 \lim_{T\to\infty}\left(1-\tau+\frac{\tau}{T}\right)
 &=1-\tau+0=1-\tau,\\
 \lim_{T\to\infty}\left(1-\tau+\frac{\tau+1}{T}\right)
 &=1-\tau+0=1-\tau.
 \end{aligned}
\]
Because \((T-k+1)/T\) lies between these two bounds,
\[
 \lim_{T\to\infty}\frac{T-k+1}{T}=1-\tau.
\]
The limit \(1-\tau\) is positive, so the reciprocal function is continuous
at \(1-\tau\). Writing \(T/(T-k+1)\) as the reciprocal of
\((T-k+1)/T\), we obtain
\begin{equation}
 \begin{aligned}
 \lim_{T\to\infty}\frac{T}{T-k+1}
 &=\lim_{T\to\infty}\frac{1}{(T-k+1)/T}\\
 &=\frac{1}{\displaystyle\lim_{T\to\infty}(T-k+1)/T}\\
 &=\frac1{1-\tau}.
 \end{aligned}
 \label{eq:optimal-profile-index-limits}
\end{equation}
Using \eqref{eq:optimal-profile-index-limits} and
\eqref{eq:optimal-profile-tail-cost-limits} in
\eqref{eq:optimal-profile-fraction-scaling} gives
\begin{equation}
 T\frac{\omega_k^\star}{W_k^\star}
 =\frac{T}{T-k+1}
   \frac{(T-k+1)C_{T-k+1}^\star}{2+C_{T-k+1}^\star}
 \longrightarrow\frac1{1-\tau}\frac4{2+0}
 =\frac2{1-\tau}.
 \label{eq:optimal-profile-fraction-limit}
\end{equation}
The desired limit of \(T\omega_k^\star\) will therefore follow once we
prove \(W_k^\star\to(1-\tau)^2\).

\emph{Step 2: choose a comparison for \(W_k^\star\).}
To prove \(W_k^\star\to(1-\tau)^2\), we seek a comparison with that limit.
The limits of \(TC_T^\star\), \((T-k+1)C_{T-k+1}^\star\), and
\((T-k+1)/T\) give
\begin{equation}
 \begin{aligned}
 \frac{C_T^\star}{C_{T-k+1}^\star}
 &=\frac{TC_T^\star}{(T-k+1)C_{T-k+1}^\star}\frac{T-k+1}{T}\\
 &\longrightarrow\frac44(1-\tau)=1-\tau.
 \end{aligned}
 \label{eq:optimal-profile-cost-ratio-limit}
\end{equation}
Thus \((C_T^\star/C_{T-k+1}^\star)^2\to(1-\tau)^2\).
To conclude \(W_k^\star\to(1-\tau)^2\), it suffices to prove
\[
 \frac{W_k^\star}{(C_T^\star/C_{T-k+1}^\star)^2}\longrightarrow1.
\]
To show \(W_k^\star/(C_T^\star/C_{T-k+1}^\star)^2\to1\), we will express
\(W_k^\star\) and
\((C_T^\star/C_{T-k+1}^\star)^2\) as products and compare their factors.

\emph{Step 3: write \(W_k^\star\) and \((C_T^\star/C_{T-k+1}^\star)^2\) as products.}
We obtain \(W_k^\star\) by multiplying \(W_{i+1}^\star/W_i^\star\) over
the earlier updates \(i=1,\ldots,k-1\). To compare
\(W_{i+1}^\star/W_i^\star\) with \((C_{T-i+1}^\star/C_{T-i}^\star)^2\),
first express \(W_{i+1}^\star/W_i^\star\) using \(C_{T-i+1}^\star\).
Equation~\eqref{eq:optimal-cost-recursion} gives
\[
 C_{T-i+1}^\star=2\bigl(\sqrt{1+C_{T-i}^\star}-1\bigr),
 \qquad
 \sqrt{1+C_{T-i}^\star}=1+\frac{C_{T-i+1}^\star}{2}.
\]
Replacing \(\sqrt{1+C_{T-i}^\star}\) by \(1+C_{T-i+1}^\star/2\)
in the second identity of
\eqref{eq:optimal-local-allocation-fraction} yields
\begin{equation}
 \frac{W_{i+1}^\star}{W_i^\star}
 =\frac1{\sqrt{1+C_{T-i}^\star}}
 =\frac1{1+C_{T-i+1}^\star/2}.
 \label{eq:optimal-profile-retained-fraction}
\end{equation}
For \(k\geq2\), multiplying over \(i=1,\ldots,k-1\) cancels the
intermediate factors \(W_i^\star\):
\[
 \prod_{i=1}^{k-1}\frac{W_{i+1}^\star}{W_i^\star}
 =\frac{W_2^\star\cdots W_k^\star}
        {W_1^\star\cdots W_{k-1}^\star}
 =\frac{W_k^\star}{W_1^\star}=W_k^\star.
\]
The last equality uses \(W_1^\star=1\).
Using \eqref{eq:optimal-profile-retained-fraction} for each
\(W_{i+1}^\star/W_i^\star\) gives
\begin{equation}
 W_k^\star
 =\prod_{i=1}^{k-1}\frac{W_{i+1}^\star}{W_i^\star}
 =\prod_{i=1}^{k-1}\frac1{1+C_{T-i+1}^\star/2}.
 \label{eq:optimal-profile-product}
\end{equation}
For \(k=1\), both products are empty and equal one, agreeing with
\(W_1^\star=1\).

To express \((C_T^\star/C_{T-k+1}^\star)^2\) over \(i=1,\ldots,k-1\),
multiply \((C_{T-i+1}^\star/C_{T-i}^\star)^2\) for \(i=1,\ldots,k-1\).
For \(k\geq2\), cancellation gives
\[
 \begin{aligned}
 \prod_{i=1}^{k-1}
 \left(\frac{C_{T-i+1}^\star}{C_{T-i}^\star}\right)^2
 &=\left(
 \frac{C_T^\star\cdots C_{T-k+2}^\star}
      {C_{T-1}^\star\cdots C_{T-k+1}^\star}
 \right)^2\\
 &=\left(\frac{C_T^\star}{C_{T-k+1}^\star}\right)^2.
 \end{aligned}
\]
When \(k=1\), the product is empty and \(C_T^\star/C_{T-k+1}^\star=1\).
Thus both product representations use the same index \(i=1,\ldots,k-1\).

\emph{Step 4: bound each factor of \(W_k^\star/(C_T^\star/C_{T-k+1}^\star)^2\).}
We want to prove \(W_k^\star/(C_T^\star/C_{T-k+1}^\star)^2\to1\).
First divide the two product representations established in Step~3:
\begin{equation}
 \begin{aligned}
 \frac{W_k^\star}{(C_T^\star/C_{T-k+1}^\star)^2}
 &=\frac{\displaystyle\prod_{i=1}^{k-1}(1+C_{T-i+1}^\star/2)^{-1}}
        {\displaystyle\prod_{i=1}^{k-1}(C_{T-i+1}^\star/C_{T-i}^\star)^2}\\
 &=\prod_{i=1}^{k-1}
   \frac{1/(1+C_{T-i+1}^\star/2)}
        {(C_{T-i+1}^\star/C_{T-i}^\star)^2}.
 \end{aligned}
 \label{eq:optimal-profile-correction-product}
\end{equation}
Equation~\eqref{eq:optimal-profile-correction-product} identifies the
factors we need to bound. We first calculate how far each factor is
above one, then use that calculation to obtain lower and upper bounds.

For \(k\geq2\), fix \(i\in\{1,\ldots,k-1\}\). To rewrite the denominator
\((C_{T-i+1}^\star/C_{T-i}^\star)^2\), use
\eqref{eq:cost-reciprocal-increment}:
\[
 \begin{aligned}
 C_{T-i}^\star
 &=C_{T-i+1}^\star+\frac{(C_{T-i+1}^\star)^2}{4}
 =C_{T-i+1}^\star\left(1+\frac{C_{T-i+1}^\star}{4}\right),\\
 \frac{C_{T-i+1}^\star}{C_{T-i}^\star}
 &=\frac1{1+C_{T-i+1}^\star/4}.
 \end{aligned}
\]
Substituting \(C_{T-i+1}^\star/C_{T-i}^\star=1/(1+C_{T-i+1}^\star/4)\)
and expanding \((1+C_{T-i+1}^\star/4)^2\) gives
\begin{equation}
 \begin{aligned}
 \frac{1/(1+C_{T-i+1}^\star/2)}
      {(C_{T-i+1}^\star/C_{T-i}^\star)^2}
 &=\frac{(1+C_{T-i+1}^\star/4)^2}{1+C_{T-i+1}^\star/2}\\
 &=\frac{1+C_{T-i+1}^\star/2+(C_{T-i+1}^\star)^2/16}
         {1+C_{T-i+1}^\star/2}\\
 &=1+\frac{(C_{T-i+1}^\star)^2}{16(1+C_{T-i+1}^\star/2)}.
 \end{aligned}
 \label{eq:optimal-profile-factor-comparison}
\end{equation}
Thus each factor in \eqref{eq:optimal-profile-correction-product} is
one plus the nonnegative term displayed in
\eqref{eq:optimal-profile-factor-comparison}.

We need an upper bound on the product in
\eqref{eq:optimal-profile-correction-product}.
The inequality \(1+x\leq e^x\) for \(x\geq0\) gives such a bound
for each factor, since the term added to one is nonnegative.
Equation~\eqref{eq:optimal-profile-factor-comparison} gives
\begin{equation}
 \begin{aligned}
 1\leq
 \frac{1/(1+C_{T-i+1}^\star/2)}
      {(C_{T-i+1}^\star/C_{T-i}^\star)^2}
 &=1+\frac{(C_{T-i+1}^\star)^2}{16(1+C_{T-i+1}^\star/2)}\\
 &\leq\exp\left(
       \frac{(C_{T-i+1}^\star)^2}{16(1+C_{T-i+1}^\star/2)}
       \right).
 \end{aligned}
 \label{eq:optimal-profile-factor-bounds}
\end{equation}
All bounds in \eqref{eq:optimal-profile-factor-bounds} are positive.
Multiplying over \(i=1,\ldots,k-1\) and adding the exponents gives
\begin{equation}
 \begin{aligned}
 1=\prod_{i=1}^{k-1}1
 &\leq\frac{W_k^\star}{(C_T^\star/C_{T-k+1}^\star)^2}\\
 &\leq\prod_{i=1}^{k-1}
   \exp\left(
       \frac{(C_{T-i+1}^\star)^2}{16(1+C_{T-i+1}^\star/2)}
       \right)\\
 &=\exp\left(
   \sum_{i=1}^{k-1}
   \frac{(C_{T-i+1}^\star)^2}{16(1+C_{T-i+1}^\star/2)}
   \right).
 \end{aligned}
 \label{eq:optimal-profile-product-sum-bound}
\end{equation}
Equation~\eqref{eq:optimal-profile-product-sum-bound} is why a sum
appears: multiplying the exponential bounds adds their exponents.
Since the exponential function is continuous and \(\exp(0)=1\),
the lower and upper bounds in \eqref{eq:optimal-profile-product-sum-bound}
will both tend to one if we prove
\[
 \sum_{i=1}^{k-1}
 \frac{(C_{T-i+1}^\star)^2}{16(1+C_{T-i+1}^\star/2)}
 \longrightarrow0.
\]
We now establish this limit.
Since \(1+C_{T-i+1}^\star/2\geq1\), we have
\begin{equation}
 0\leq
 \frac{(C_{T-i+1}^\star)^2}{16(1+C_{T-i+1}^\star/2)}
 \leq\frac{(C_{T-i+1}^\star)^2}{16}.
 \label{eq:optimal-profile-correction-bounds}
\end{equation}
Thus we need to control \(\sum_{i=1}^{k-1}(C_{T-i+1}^\star)^2\).
The cost recursion lets us evaluate this sum exactly:
\[
 C_{T-i}^\star-C_{T-i+1}^\star
 =\frac{(C_{T-i+1}^\star)^2}{4},
 \qquad
 (C_{T-i+1}^\star)^2=4(C_{T-i}^\star-C_{T-i+1}^\star).
\]
The differences \(C_{T-i}^\star-C_{T-i+1}^\star\) are useful because
a cost appearing positively in one term appears negatively in the next.
For example, when \(k\geq3\), the first two terms give
\[
 \begin{aligned}
 &(C_{T-1}^\star-C_T^\star)
  +(C_{T-2}^\star-C_{T-1}^\star)\\
 &=C_{T-1}^\star-C_T^\star+C_{T-2}^\star-C_{T-1}^\star\\
 &=C_{T-2}^\star-C_T^\star.
 \end{aligned}
\]
Here \(+C_{T-1}^\star\) cancels with \(-C_{T-1}^\star\).
For the full sum, write the positive and negative terms separately.
For \(k\geq2\), this gives
\[
 \begin{aligned}
 \sum_{i=1}^{k-1}(C_{T-i}^\star-C_{T-i+1}^\star)
 &=(C_{T-1}^\star+\cdots+C_{T-k+1}^\star)\\
 &\quad -(C_T^\star+\cdots+C_{T-k+2}^\star)\\
 &=C_{T-k+1}^\star-C_T^\star.
 \end{aligned}
\]
For \(k=1\), the sum is empty and \(C_{T-k+1}^\star-C_T^\star=0\),
so the identity still holds. Applying the squared-cost identity term
by term therefore gives
\begin{equation}
 \begin{aligned}
 \sum_{i=1}^{k-1}(C_{T-i+1}^\star)^2
 &=\sum_{i=1}^{k-1}4(C_{T-i}^\star-C_{T-i+1}^\star)\\
 &=4\sum_{i=1}^{k-1}(C_{T-i}^\star-C_{T-i+1}^\star)\\
 &=4(C_{T-k+1}^\star-C_T^\star).
 \end{aligned}
 \label{eq:optimal-profile-squared-cost-sum}
\end{equation}

We can now bound the full sum of the terms added to one.
Summing \eqref{eq:optimal-profile-correction-bounds} and using
\eqref{eq:optimal-profile-squared-cost-sum} yields
\begin{equation}
 \begin{aligned}
 0\leq
 \sum_{i=1}^{k-1}
 \frac{(C_{T-i+1}^\star)^2}{16(1+C_{T-i+1}^\star/2)}
 &\leq\frac1{16}\sum_{i=1}^{k-1}(C_{T-i+1}^\star)^2\\
 &=\frac{4(C_{T-k+1}^\star-C_T^\star)}{16}\\
 &=\frac{C_{T-k+1}^\star-C_T^\star}{4}\\
 &\leq\frac{C_{T-k+1}^\star}{4}\longrightarrow0.
 \end{aligned}
 \label{eq:optimal-profile-correction-sum-bound}
\end{equation}
The last inequality uses \(C_T^\star>0\), and the limit uses
\(C_{T-k+1}^\star\to0\) from \eqref{eq:optimal-profile-tail-cost-limits}.
Thus \eqref{eq:optimal-profile-correction-sum-bound} controls all
\(k-1\) terms together, including the effect of the changing number of terms.

\emph{Step 5: bound the product \(W_k^\star/(C_T^\star/C_{T-k+1}^\star)^2\).}
We have proved \eqref{eq:optimal-profile-correction-sum-bound}.
Using this bound on the exponent in
\eqref{eq:optimal-profile-product-sum-bound}, and the fact that
the exponential function is increasing, gives
\begin{equation}
 \begin{aligned}
 1\leq\frac{W_k^\star}{(C_T^\star/C_{T-k+1}^\star)^2}
 &\leq\exp\left(
   \sum_{i=1}^{k-1}
   \frac{(C_{T-i+1}^\star)^2}{16(1+C_{T-i+1}^\star/2)}
   \right)\\
 &\leq\exp\left(\frac{C_{T-k+1}^\star-C_T^\star}{4}\right).
 \end{aligned}
 \label{eq:optimal-profile-comparison-bound}
\end{equation}

It remains to evaluate the upper bound in
\eqref{eq:optimal-profile-comparison-bound}.
Equation~\eqref{eq:optimal-profile-tail-cost-limits} gives
\(C_{T-k+1}^\star\to0\), and Part~1 gives
\[
 C_T^\star=\frac{TC_T^\star}{T}\longrightarrow0,
 \qquad\text{since }TC_T^\star\to4\text{ and }T\to\infty.
\]
Consequently,
\[
 \begin{aligned}
 \lim_{T\to\infty}\frac{C_{T-k+1}^\star-C_T^\star}{4}
 &=\frac{\displaystyle\lim_{T\to\infty}C_{T-k+1}^\star
          -\displaystyle\lim_{T\to\infty}C_T^\star}{4}\\
 &=\frac{0-0}{4}=0.
 \end{aligned}
\]
Continuity of the exponential function now gives
\[
 \begin{aligned}
 \lim_{T\to\infty}
 \exp\left(\frac{C_{T-k+1}^\star-C_T^\star}{4}\right)
 &=\exp\left(\lim_{T\to\infty}
             \frac{C_{T-k+1}^\star-C_T^\star}{4}\right)\\
 &=\exp(0)=1.
 \end{aligned}
\]
To prove \(W_k^\star/(C_T^\star/C_{T-k+1}^\star)^2\to1\), subtract one from
\eqref{eq:optimal-profile-comparison-bound}:
\[
 0\leq
 \frac{W_k^\star}{(C_T^\star/C_{T-k+1}^\star)^2}-1
 \leq\exp\left(\frac{C_{T-k+1}^\star-C_T^\star}{4}\right)-1
 \longrightarrow1-1=0.
\]
By the squeeze theorem,
\(W_k^\star/(C_T^\star/C_{T-k+1}^\star)^2-1\to0\).
Adding one back gives
\begin{equation}
 \frac{W_k^\star}{(C_T^\star/C_{T-k+1}^\star)^2}
 =1+\left(\frac{W_k^\star}{(C_T^\star/C_{T-k+1}^\star)^2}-1\right)
 \longrightarrow1+0=1.
 \label{eq:optimal-profile-comparison-limit}
\end{equation}

\emph{Step 6: obtain the limits of \(W_k^\star\) and \(T\omega_k^\star\).}
The limits \eqref{eq:optimal-profile-comparison-limit} and
\eqref{eq:optimal-profile-cost-ratio-limit} give the limit of \(W_k^\star\):
\begin{equation}
 \begin{aligned}
 W_k^\star
 &=\frac{W_k^\star}{(C_T^\star/C_{T-k+1}^\star)^2}
   \left(\frac{C_T^\star}{C_{T-k+1}^\star}\right)^2\\
 &\longrightarrow1\cdot(1-\tau)^2=(1-\tau)^2.
 \end{aligned}
 \label{eq:optimal-remaining-mass-limit}
\end{equation}
Multiplying the limits in
\eqref{eq:optimal-remaining-mass-limit} and \eqref{eq:optimal-profile-fraction-limit} gives
\begin{equation}
 T\omega_k^\star
 =W_k^\star\left(T\frac{\omega_k^\star}{W_k^\star}\right)
 \longrightarrow(1-\tau)^2\frac2{1-\tau}
 =2(1-\tau).
 \label{eq:optimal-coefficient-scale}
\end{equation}

At \(\tau=0\), we have \(k=1\) and \(T-k+1=T\).
The products over \(i=1,\ldots,k-1\) are empty and equal one.
The lower and upper bounds in \eqref{eq:optimal-profile-comparison-bound}
then both equal one:
\[
 W_1^\star=1,\qquad
 \frac{W_1^\star}{(C_T^\star/C_T^\star)^2}=1.
\]
Thus the argument includes \(\tau=0\), where \(T\omega_1^\star\to2\),
and proves \eqref{eq:optimal-profile-limit} for every fixed
\(0\leq\tau<1\).

To find the limit of \(\omega_k^\star/\omega_1^\star\), multiply numerator
and denominator by \(T\) and use the positive limit
\(T\omega_1^\star\to2\):
\[
 \frac{\omega_{1+\lfloor\tau(T-1)\rfloor}^\star}{\omega_1^\star}
 =\frac{T\omega_{1+\lfloor\tau(T-1)\rfloor}^\star}{T\omega_1^\star}
 \longrightarrow\frac{2(1-\tau)}2=1-\tau.
\]
}%
\end{proof}

\subsection{Performance of Linear Decay}
\label[appendix]{app:linear-benchmark}
The exact optimizer approaches linear decay; we now verify that the
closed-form linear schedule has the same leading cost~\(4/T\).
Start from the linear shape \(1-\tau\), where \(\tau\) is a fraction of
the horizon. At step \(t=1,\ldots,T\), set \(\tau=(t-1)/T\).
The linear weights are then
\[
 q_t=1-\frac{t-1}{T}=\frac{T+1-t}{T}.
\]
These weights decrease from one to \(1/T\), reaching zero at the virtual
index \(T+1\). Their coefficient sum is
\[
 \sum_{t=1}^T q_t
 =\frac1T\sum_{t=1}^T(T+1-t)
 =\frac1T\frac{T(T+1)}2=\frac{T+1}2.
\]
Dividing each weight by this sum preserves its linear shape and gives the
normalized schedule
\begin{equation}
 \omega_t^{\mathrm{lin}}
 =\frac{q_t}{\sum_{j=1}^Tq_j}
 =\frac{(T+1-t)/T}{(T+1)/2}
 =\frac{2(T+1-t)}{T(T+1)},
 \qquad 1\leq t\leq T.
 \label{eq:positive-linear-allocation}
\end{equation}
Every entry is positive (the smallest is
\(\omega_T^{\mathrm{lin}}=2/[T(T+1)]\)), so
every following-tail denominator in the schedule-shape cost is nonzero.

Since \((T+1-k)/(T-k)=1+1/(T-k)\), the cost sum will reduce to a
harmonic series. The tail coefficient sum after step~\(k\) is
\[
 \sum_{t=k+1}^T\omega_t^{\mathrm{lin}}
 =\frac{2}{T(T+1)}\sum_{t=k+1}^T(T+1-t)
 =\frac{(T-k)(T-k+1)}{T(T+1)}.
\]
Therefore, the contribution of step \(k\) to the cost is
\[
 \frac{(\omega_k^{\mathrm{lin}})^2}
      {\sum_{t=k+1}^T\omega_t^{\mathrm{lin}}}
 =\frac{4(T+1-k)}{T(T+1)(T-k)}.
\]
Adding these contributions to the terminal term
\(\omega_T^{\mathrm{lin}}=2/[T(T+1)]\) gives
\begin{equation}
\begin{aligned}
 C_T(\omega_{1:T}^{\mathrm{lin}})
 &=\frac{2}{T(T+1)}
   +\frac{4}{T(T+1)}
      \sum_{k=1}^{T-1}\frac{T+1-k}{T-k}\\
 &=\frac{2}{T(T+1)}
   +\frac{4}{T(T+1)}
      \left(T-1+\sum_{k=1}^{T-1}\frac1k\right)\\
 &=\frac4T+\frac{4\sum_{k=1}^{T-1}k^{-1}-6}{T(T+1)}.
\end{aligned}
\label{eq:positive-linear-cost}
\end{equation}
For \(T=1\), the sums are empty, and the cost is one, as required.

For \(T\geq2\), integral comparison gives
\(\sum_{k=1}^{T-1}k^{-1}\leq1+\log(T-1)\). The correction in
\eqref{eq:positive-linear-cost} is therefore \(O(\log T/T^2)\).
Multiplying the cost by \(T\) makes that correction vanish, so
\(TC_T(\omega_{1:T}^{\mathrm{lin}})\to4\). Comparing with the exact optimal cost,
\[
 \frac{C_T(\omega_{1:T}^{\mathrm{lin}})}{C_T^\star}
 =\frac{TC_T(\omega_{1:T}^{\mathrm{lin}})}{TC_T^\star}
 \longrightarrow\frac44=1.
\]
Optimizing each schedule's amplitude separately via
\Cref{prop:optimal-mass},
\begin{equation}
\begin{aligned}
 \frac{\min_{M>0}\Omega_T(M,\omega_{1:T}^{\mathrm{lin}})}
      {\min_{M>0,\,\omega_{1:T}}\Omega_T(M,\omega_{1:T})}
 &=\frac{DG\sqrt{(1+\thetacap)C_T(\omega_{1:T}^{\mathrm{lin}})}}
         {DG\sqrt{(1+\thetacap)C_T^\star}}\\
 &=\sqrt{\frac{C_T(\omega_{1:T}^{\mathrm{lin}})}{C_T^\star}}
 \longrightarrow1.
\end{aligned}
\label{eq:linear-bound-asymptotic}
\end{equation}
The remaining cost ratio tends to one, so prescribed linear decay, with its own
optimal amplitude, attains the same leading bound as the exact finite-horizon optimizer.

After optimizing their amplitudes, both the exact optimal schedule and linear decay
yield the same leading value for our spherical last-iterate bound:
\[
 \frac{2DG\sqrt{1+\thetacap}}{\sqrt T}.
\]

\subsection{Schedule and Bound Trajectories}
\label[appendix]{app:schedule-comparison-protocol}
\Cref{fig:schedule-bound-theory} compares the spherical last-iterate bound across
schedule families, each with its amplitude optimized for \(T=20{,}000\), alongside
the optimal-bound envelope. Here we detail the calculation, then examine the effects
of training horizon and schedule amplitude.

\subsubsection{Bound Evaluations}
For a horizon \(T\), normalize each of the following positive sequences to have
unit sum:
\begin{equation}
 \omega_t^{\mathrm{const}}\propto1,\qquad
 \omega_t^{\mathrm{lin}}\propto1-\frac{t-1}{T},\qquad
 \omega_t^{\mathrm{cos}}\propto\frac{1+\cos\!\left(\pi(t-1)/T\right)}2,
 \qquad 1\leq t\leq T.
 \label{eq:comparison-schedule-shapes}
\end{equation}
For the horizons below, set \(T_0=4T/5\). Following
\citet[Eq.~(3)]{schaipp2025surprising}, the WSD shape has no warmup, stays
constant through most of the horizon, and then uses a linear cooldown:
\begin{equation}
 \omega_t^{\mathrm{WSD}}\propto
 \begin{cases}
 1,&1\leq t<T_0,\\
 1-\dfrac{t-T_0}{T+1-T_0},&T_0\leq t\leq T.
 \end{cases}
 \label{eq:comparison-wsd-shape}
\end{equation}
The proportionality constant in each line is fixed by
\(\sum_{t=1}^T\omega_t=1\). The linear, cosine, and WSD sequences reach zero at
the virtual index \(T+1\), whereas the constant sequence remains constant. The
exact schedule is \(\omega_{1:T}^\star\) from
\Cref{cor:optimal-allocation}.

\begin{figure}[!htbp]
 \centering
 \includegraphics[width=.5\linewidth]{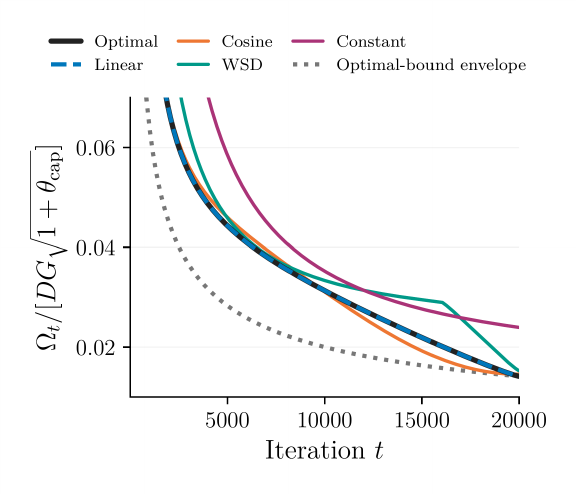}
 \caption{Normalized last-iterate bounds at \(T=20{,}000\). Each schedule uses its final-horizon optimal amplitude; the dotted envelope reoptimizes normalized schedule and amplitude at every iteration.}
 \label{fig:schedule-bound-theory}
\end{figure}

Each schedule receives the optimal coefficient sum from \Cref{prop:optimal-mass}.
We divide the bounds by the common factor \(DG\sqrt{1+\thetacap}\) to remove
the problem-dependent scale while preserving schedule rankings and bound ratios.
By \Cref{prop:optimal-mass}, the normalized final value is then
\(\sqrt{C_T(\omega_{1:T})}\).
To compute the bound trajectory in \Cref{fig:schedule-bound-theory}, retain this
final-horizon amplitude and apply \Cref{thm:last-iterate} to the first \(t\)
coefficients. Direct substitution gives the normalized bound value:
\begin{equation}
 \frac{\Omega_t}{DG\sqrt{1+\thetacap}}
 =\frac{\sqrt{C_T(\omega_{1:T})}}{2\sum_{j=1}^t\omega_j}
 +\frac{1}{2\sqrt{C_T(\omega_{1:T})}}
 \left[\omega_t+\sum_{k=1}^{t-1}
 \frac{\omega_k^2}{\sum_{j=k+1}^t\omega_j}\right].
 \label{eq:comparison-prefix-bound}
\end{equation}
Each curve therefore uses one normalized schedule and amplitude chosen for its final
horizon \(T\).

For comparison, suppose instead that each displayed index \(t\) were chosen as
the final horizon and both normalized schedule and amplitude were optimized again.
\Cref{thm:optimal-schedule} and \Cref{prop:optimal-mass} give the gray optimal-bound
envelope directly:
\begin{equation}
 \frac{1}{DG\sqrt{1+\thetacap}}
   \min_{\substack{M>0,\;\omega_j>0\\\sum_{j=1}^t\omega_j=1}}
   \Omega_t(M,\omega_{1:t})
 =\sqrt{C_t^\star}.
 \label{eq:comparison-optimal-bound-envelope}
\end{equation}
Each point on this envelope uses a different normalized schedule and amplitude;
the envelope therefore does not represent one fixed schedule.

\begin{figure}[!htbp]
 \centering
 \begin{subfigure}{.42\linewidth}
  \includegraphics[width=\linewidth]{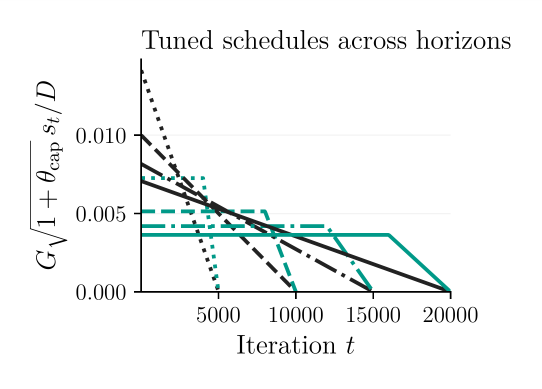}\caption{}
 \end{subfigure}\hspace{.04\linewidth}
 \begin{subfigure}{.42\linewidth}
  \includegraphics[width=\linewidth]{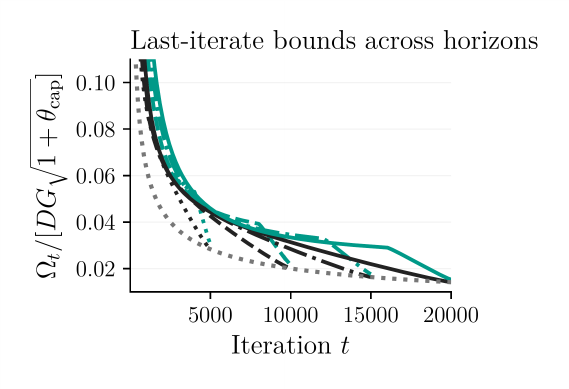}\caption{}
 \end{subfigure}
 \caption{Exact optimum (black) and WSD (teal) trajectories at horizons
 \(T\in\{5{,}000,10{,}000,15{,}000,20{,}000\}\), shown against iteration. WSD uses no warmup and a final 20\% linear
 cooldown. Dotted, dashed, dash-dotted, and solid lines indicate \(T=5{,}000,10{,}000,15{,}000,20{,}000\), respectively. \textbf{(a)} Rescaled rotational coefficients.
 \textbf{(b)} Prefix bounds from \eqref{eq:comparison-prefix-bound},
 with the gray dotted optimal-bound envelope
 from \eqref{eq:comparison-optimal-bound-envelope}. Each exact optimum meets the envelope at its final horizon.}
 \label{fig:schedule-bound-horizons}
\end{figure}

\Cref{fig:schedule-bound-horizons} compares the exact optimum and WSD across
training horizons, showing rotational coefficients in panel (a) and bound
trajectories from \eqref{eq:comparison-prefix-bound} in panel (b).
Multiplying the optimal coefficients by \(G\sqrt{1+\thetacap}/D\) removes their
common problem-dependent scale while retaining differences in schedule shape
and optimal amplitude. The rescaled coefficient in \Cref{fig:schedule-bound-horizons}(a) is
\[
 \frac{G\sqrt{1+\thetacap}}{D}s_t
 =\frac{\omega_t}{\sqrt{C_T(\omega_{1:T})}}.
\]

\begin{figure}[!htbp]
 \centering
 \includegraphics[width=.5\linewidth]{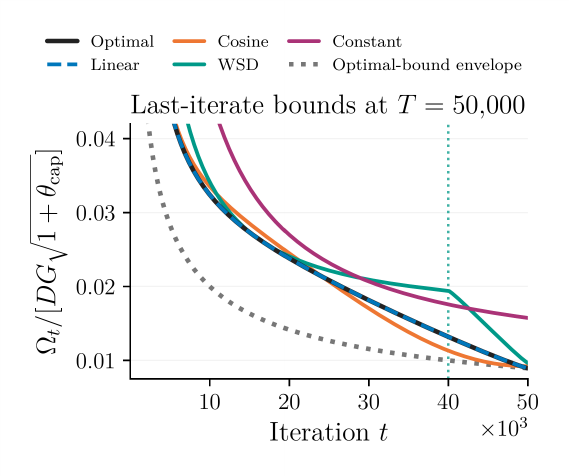}
 \caption{Last-iterate bounds at the longer horizon \(T=50{,}000\). The dotted gray curve is the
 optimal-bound envelope, and the vertical dotted line marks the start of WSD's
 final 20\% linear cooldown at \(t=40{,}000\).}
 \label{fig:schedule-bound-large-horizon}
\end{figure}

\Cref{fig:schedule-bound-large-horizon} extends the bound comparison to
\(T=50{,}000\), with WSD's final 20\% cooldown starting at \(t=40{,}000\).

\FloatBarrier
\subsubsection{Sensitivity to Schedule Amplitude}
\begin{figure}[!htbp]
 \centering
 \includegraphics[width=.50\linewidth]{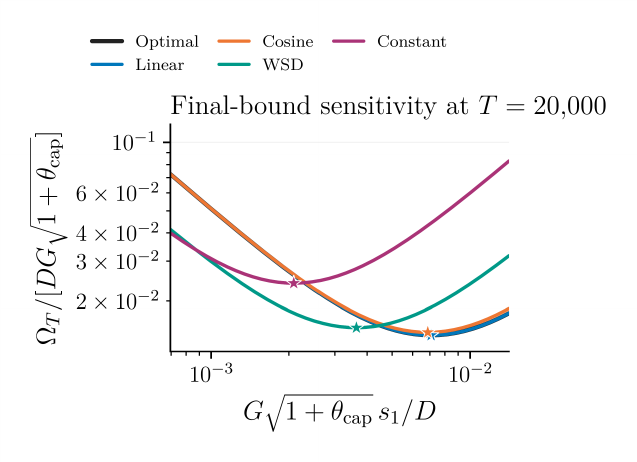}
 \caption{Final bound versus the rescaled initial coefficient
 \(G\sqrt{1+\thetacap}\,s_1/D\) at \(T=20{,}000\). Stars mark the minima from
 \eqref{eq:comparison-amplitude-sweep}. \textcolor{black}{The minima occur at different rescaled
 initial coefficients, while relative sensitivity around each minimum is}
 \(\bigl(r+r^{-1}\bigr)/2\).}
 \label{fig:schedule-bound-amplitude}
\end{figure}

The preceding comparisons optimize each schedule's amplitude; we now examine
how the bound changes when that amplitude varies.
For the amplitude-sensitivity comparison in \Cref{fig:schedule-bound-amplitude},
hold \(\omega_{1:T}\) fixed and vary its initial coefficient \(s_1\), which scales
the entire schedule through \(s_t=s_1\omega_t/\omega_1\).
The horizontal coordinate \(G\sqrt{1+\thetacap}\,s_1/D\) uses the same rescaling
as \Cref{fig:schedule-bound-horizons}(a), evaluated at the first step.
This common axis shows both the different optimal initial coefficients and
the bounds at the same initial coefficient. The final bound becomes
\begin{equation}
 \begin{aligned}
 \frac{\Omega_T}{DG\sqrt{1+\thetacap}}
 &=\frac{D\omega_1}{2G\sqrt{1+\thetacap}\,s_1}
 +\frac{G\sqrt{1+\thetacap}\,s_1}{2D\omega_1}
 C_T(\omega_{1:T}),\\
 s_1^\star&=\frac{D\omega_1}
 {G\sqrt{(1+\thetacap)C_T(\omega_{1:T})}}.
 \end{aligned}
 \label{eq:comparison-amplitude-sweep}
\end{equation}
In \Cref{fig:schedule-bound-amplitude}, linear decay, the exact optimum, and cosine
remain close as the amplitude varies, extending the similarity of their optimized
bounds in \Cref{fig:schedule-bound-theory}. WSD reaches its minimum at a smaller
amplitude and a higher bound value; the constant schedule requires an even smaller
amplitude and yields a still higher minimum.

Multiplying any schedule's
own optimum \(s_1^\star\) by a factor \(r>0\) multiplies its minimum bound by
\(\bigl(r+r^{-1}\bigr)/2\), so all schedules have the same relative sensitivity
around their own optimum.

\FloatBarrier

%% file: appendix-witness.tex
\section{A Simple Admissible Scale-Invariant Example}
\label[appendix]{app:si-witness}

\paragraph{Full-domain Euclidean convexity.}
Suppose \(L:\R^C\to\R\) is finite and convex, and
\(L(\kappa\bx)=L(\bx)\) for every \(\bx\in\R^C\) and \(\kappa>0\).
{\color{black}%
Convexity at the midpoint \(\bx=(0+2\bx)/2\) and scale invariance give
\[
 L(\bx)\leq\frac{L(0)+L(2\bx)}2=\frac{L(0)+L(\bx)}2.
\]
Thus \(L(\bx)\leq L(0)\) for every \(\bx\). Applying convexity at the midpoint
of \(\bx\) and \(-\bx\) gives
\[
 L(0)\leq\frac{L(\bx)+L(-\bx)}2\leq L(0).
\]
Both losses in the average are at most \(L(0)\), so equality forces them
both to equal \(L(0)\). Hence \(L\) is constant.
}%

{\color{black}%
We now construct a nonconstant scale-invariant loss satisfying the
assumptions of \Cref{thm:last-iterate}. We verify geodesic convexity on
\(\mathcal C\), the oracle conditions, and containment of the iterates
when \(s_t=M^\star(\omega_{1:T})\omega_t\).
\Cref{prop:optimal-mass} minimizes the bound over \(M\), but applying
\Cref{thm:last-iterate} at \(M=M^\star\) still requires that the resulting
iterates stay in \(\mathcal C\). The example verifies this requirement
for every positive normalized schedule \(\omega_{1:T}\).
}%

\begin{proposition}{A simple admissible scale-invariant model}{si-witness}
{\color{black}%
Fix an integer \(T\geq1\) and \(\Lambda>0\). Choose \(\xhat^\star\in\Sph\)
and set \(\bm c_{\mathrm{cap}}=\xhat^\star\) in \eqref{eq:working-cap}, with
\(0<\thetacap\leq\pi/2\). For \(\bx\ne0\), define
}%
\begin{equation}
 L(\bx):=\Lambda\left(1-\inner{\frac{\bx}{\norm{\bx}}}{\xhat^\star}\right).
 \label{eq:witness-loss}
\end{equation}
Choose a deterministic initial point satisfying
\begin{equation}
 \xhat_1\in\mathcal C\setminus\{\xhat^\star\},\qquad
 \textcolor{black}{D=\dist(\xhat_1,\xhat^\star)\in(0,\thetacap).}
 \label{eq:witness-initialization}
\end{equation}
{\color{black}%
For the oracle and its norm bound, set
}%
\begin{equation}
 \gtilde_t:=\nabla_{\mathbb S}L(\xhat_t),\qquad G:=\Lambda\thetacap.
 \label{eq:witness-oracle}
\end{equation}
The loss is smooth and positively scale-invariant on \(\R^C\setminus\{0\}\).
{\color{black}%
On \(\mathcal C\), it is strictly geodesically convex with unique minimizer
\(\xhat^\star\), and its spherical gradient has norm at most \(G\).
}%

{\color{black}%
For every fixed positive normalized schedule \(\omega_{1:T}\), the coefficients
}%
\begin{equation}
 s_t=M^\star(\omega_{1:T})\omega_t,\qquad 1\leq t\leq T,
 \label{eq:witness-schedule}
\end{equation}
{\color{black}%
with \(M^\star\) from \Cref{prop:optimal-mass}, keep the iterates of
\eqref{eq:exact-finite-step} through \(\xhat_{T+1}\) inside \(\mathcal C\).
Thus \Cref{thm:last-iterate} applies, including for the profile
}%
\(\omega_{1:T}^\star\) from \Cref{cor:optimal-allocation}.
\end{proposition}

\begin{proof}
{\color{black}%
We verify the claims and the assumptions of \Cref{thm:last-iterate}
in the following order:
\begin{enumerate}[label=\arabic*.,leftmargin=*,itemsep=0pt]
 \item \(L\) is smooth, scale-invariant, and nonconstant, with unique
 minimizer \(\xhat^\star\) on \(\mathcal C\).
 \item \(L\) is strictly geodesically convex on \(\mathcal C\).
 \item The exact-gradient oracle is tangent and unbiased, and
 \(\norm{\nabla_{\mathbb S}L(\xhat)}\leq G\) for \(\xhat\in\mathcal C\).
 \item The prescribed \(s_t=M^\star(\omega_{1:T})\omega_t\) are
 deterministic and satisfy \(0<\Lambda s_t<1\).
 \item Every iterate through \(\xhat_{T+1}\) stays in \(\mathcal C\),
 so the oracle norm bound holds throughout the trajectory.
\end{enumerate}

\emph{Step 1: loss properties and minimizer.}
We first verify that \eqref{eq:witness-loss} defines a smooth,
scale-invariant loss with minimizer \(\xhat^\star\).
The map \(\bx\mapsto\bx/\norm{\bx}\) is smooth for \(\bx\ne0\), so
\(L\) is smooth on \(\R^C\setminus\{0\}\). For \(\kappa>0\),
\(\norm{\kappa\bx}=\kappa\norm{\bx}\), and substitution gives
\[
 \begin{aligned}
 L(\kappa\bx)
 &=\Lambda\left(1-\inner{\frac{\kappa\bx}{\norm{\kappa\bx}}}{\xhat^\star}\right)\\
 &=\Lambda\left(1-\inner{\frac{\kappa\bx}{\kappa\norm{\bx}}}{\xhat^\star}\right)
 =\Lambda\left(1-\inner{\frac{\bx}{\norm{\bx}}}{\xhat^\star}\right)
 =L(\bx).
 \end{aligned}
\]
On the sphere, \(\norm{\xhat}=1\), so
\[
 L(\xhat)=\Lambda(1-\inner{\xhat}{\xhat^\star})\geq0.
\]
The inner product of two unit vectors is at most one, with equality
only when they coincide. Thus \(L(\xhat)=0\) exactly when
\(\xhat=\xhat^\star\), proving that \(\xhat^\star\) is the unique minimizer.
Since \(\xhat_1\ne\xhat^\star\) by \eqref{eq:witness-initialization},
\(L(\xhat_1)>L(\xhat^\star)=0\), so \(L\) is nonconstant on \(\mathcal C\).

\emph{Step 2: geodesic convexity on \(\mathcal C\).}
We next verify the geodesic-convexity assumption of
\Cref{thm:last-iterate}. By \Cref{def:geodesic-convexity}, this requires
the loss restricted to every shortest arc in \(\mathcal C\) to be
convex as a function of the arc parameter. We will show the stronger
property \((L\circ\gamma)''(a)>0\) along every nonconstant shortest arc.

Take the shortest arc \(\gamma\) between any two distinct points of
\(\mathcal C\). By \Cref{cor:cap-shortest-arc}, the entire arc lies in
\(\mathcal C\). Parametrize \(\gamma\) by arc length \(a\), so that
\(\norm{\gamma'(0)}=1\); primes below denote derivatives with respect
to \(a\). To calculate the derivatives of \(L\circ\gamma\), first
differentiate the spherical arc formula \eqref{eq:sphere-exp}:
\[
 \begin{aligned}
 \gamma(a)&=\cos(a)\gamma(0)+\sin(a)\gamma'(0),\\
 \gamma'(a)&=-\sin(a)\gamma(0)+\cos(a)\gamma'(0),\\
 \gamma''(a)&=-\cos(a)\gamma(0)-\sin(a)\gamma'(0)=-\gamma(a).
 \end{aligned}
\]
Substituting \(\gamma(a)\) into the loss and differentiating gives
\[
 \begin{aligned}
 L(\gamma(a))
 &=\Lambda(1-\inner{\gamma(a)}{\xhat^\star}),\\
 (L\circ\gamma)'(a)
 &=-\Lambda\inner{\gamma'(a)}{\xhat^\star},\\
 (L\circ\gamma)''(a)
 &=-\Lambda\inner{\gamma''(a)}{\xhat^\star}\\
 &=\Lambda\inner{\gamma(a)}{\xhat^\star}\\
 &=\Lambda\cos\!\bigl(\dist(\gamma(a),\xhat^\star)\bigr).
 \end{aligned}
\]
To determine the sign, use the fact that \(\mathcal C\) is centered
at \(\xhat^\star\):
\[
 \dist(\gamma(a),\xhat^\star)<\thetacap\leq\frac{\pi}{2}
 \quad\Longrightarrow\quad
 (L\circ\gamma)''(a)
 =\Lambda\cos\!\bigl(\dist(\gamma(a),\xhat^\star)\bigr)>0.
\]
Thus \(L\circ\gamma\) is strictly convex along every nonconstant
shortest arc in \(\mathcal C\), proving strict geodesic convexity.
For coincident endpoints the arc is constant and the convexity
inequality is an equality. The strict sign above also holds when
\(\thetacap=\pi/2\), since the cap is open.

\emph{Step 3: identify the oracle and bound its norm on \(\mathcal C\).}
We chose \(\gtilde_t=\nabla_{\mathbb S}L(\xhat_t)\).
Tangency follows from the spherical-gradient definition.
It also follows from scale invariance: Step~1 allows us to use the
gradient relation proved in \Cref{app:si-directional-recursion},
\[
 \inner{\nabla_{\bx}L(\bx)}{\bx}=0,\qquad
 \nabla_{\mathbb S}L(\xhat)=\norm{\bx}\nabla_{\bx}L(\bx).
\]
We use \Cref{lem:si-gradient-bridge} to obtain the formula for
\(\nabla_{\mathbb S}L\) by differentiating the explicit loss
\eqref{eq:witness-loss} in Euclidean coordinates.

First differentiate the reciprocal norm:
\[
 \begin{aligned}
 \nabla_{\bx}\norm{\bx}^{-1}
 &=\nabla_{\bx}\bigl(\inner{\bx}{\bx}^{-1/2}\bigr)\\
 &=-\tfrac12\inner{\bx}{\bx}^{-3/2}\,2\bx
 =-\frac{\bx}{\norm{\bx}^3}.
 \end{aligned}
\]
Using \(\nabla_{\bx}\inner{\bx}{\xhat^\star}=\xhat^\star\), the product
rule then gives
\[
 \begin{aligned}
 \nabla_{\bx}L(\bx)
 &=-\Lambda\nabla_{\bx}
       \bigl(\inner{\bx}{\xhat^\star}\norm{\bx}^{-1}\bigr)\\
 &=-\Lambda\left(
       \frac{\xhat^\star}{\norm{\bx}}
       +\inner{\bx}{\xhat^\star}\nabla_{\bx}\norm{\bx}^{-1}\right)\\
 &=-\Lambda\left(
       \frac{\xhat^\star}{\norm{\bx}}
       -\frac{\inner{\bx}{\xhat^\star}\bx}{\norm{\bx}^3}\right).
 \end{aligned}
\]
At \(\bx=\xhat\), the radius is one. Substitution into
\eqref{eq:si-gradient-bridge-proof} therefore yields
\begin{equation}
 \begin{aligned}
 \nabla_{\mathbb S}L(\xhat)
 &=\nabla_{\bx}L(\xhat)\\
 &=-\Lambda\bigl(\xhat^\star-\inner{\xhat}{\xhat^\star}\xhat\bigr).
 \end{aligned}
 \label{eq:witness-gradient}
\end{equation}
To verify tangency of \(\gtilde_t\), recall that scale invariance gives
\(\inner{\nabla_{\bx}L(\bx)}{\bx}=0\).
At the unit vector \(\xhat_t\), \Cref{lem:si-gradient-bridge} gives
\(\nabla_{\mathbb S}L(\xhat_t)=\nabla_{\bx}L(\xhat_t)\). Hence
\[
 \begin{aligned}
 \inner{\gtilde_t}{\xhat_t}
 &=\inner{\nabla_{\mathbb S}L(\xhat_t)}{\xhat_t}\\
 &=\inner{\nabla_{\bx}L(\xhat_t)}{\xhat_t}=0.
 \end{aligned}
\]
Thus \(\gtilde_t\in T_{\xhat_t}\Sph\), since the tangent space consists
of vectors orthogonal to \(\xhat_t\).
Because \(\xhat_t\) is known before draw \(t\), the exact gradient
\(\nabla_{\mathbb S}L(\xhat_t)\) is \(\mathcal F_t\)-measurable. Hence
\[
 \E[\gtilde_t\mid\mathcal F_t]
 =\gtilde_t=\nabla_{\mathbb S}L(\xhat_t),
\]
which verifies conditional unbiasedness.

To bound the exact-gradient oracle by \(G\), expand the squared norm
of the vector in \eqref{eq:witness-gradient}:
\[
 \begin{aligned}
 \norm{\xhat^\star-\inner{\xhat}{\xhat^\star}\xhat}^2
 &=\norm{\xhat^\star}^2-2\inner{\xhat}{\xhat^\star}^2
   +\inner{\xhat}{\xhat^\star}^2\norm{\xhat}^2\\
 &=1-\inner{\xhat}{\xhat^\star}^2.
 \end{aligned}
\]
For \(\xhat\in\mathcal C\), this gives
\begin{equation}
 \begin{aligned}
 \norm{\nabla_{\mathbb S}L(\xhat)}
 &=\Lambda\sqrt{1-\inner{\xhat}{\xhat^\star}^2}\\
 &=\Lambda\sin\!\bigl(\dist(\xhat,\xhat^\star)\bigr)\\
 &\leq\Lambda\dist(\xhat,\xhat^\star)
 <\Lambda\thetacap=G.
 \end{aligned}
 \label{eq:witness-gradient-bound}
\end{equation}
Here \(\sin(\dist(\xhat,\xhat^\star))\geq0\) because
\(0\leq\dist(\xhat,\xhat^\star)<\thetacap\leq\pi/2\), and
\(\sin(\dist(\xhat,\xhat^\star))\leq\dist(\xhat,\xhat^\star)\).
The bound holds at every point of \(\mathcal C\). To apply it throughout
the optimization, we still need to show that every iterate stays in
\(\mathcal C\).

\emph{Step 4: deterministic coefficients with \(0<\Lambda s_t<1\).}
The initialization, cap, and profile \(\omega_{1:T}\) are fixed.
Thus \eqref{eq:optimal-mass} and \eqref{eq:witness-schedule} determine
positive deterministic coefficients \(s_t\) before the updates.
To identify a sufficient bound on \(s_t\) for containment, substitute
\eqref{eq:witness-gradient} into the trial vector in \eqref{eq:exact-finite-step}:
\begin{equation}
 \begin{aligned}
 \xhat_t-s_t\gtilde_t
 &=\xhat_t+\Lambda s_t\bigl(\xhat^\star
             -\inner{\xhat_t}{\xhat^\star}\xhat_t\bigr)\\
 &=\bigl(1-\Lambda s_t\inner{\xhat_t}{\xhat^\star}\bigr)\xhat_t
   +\Lambda s_t\xhat^\star.
 \end{aligned}
 \label{eq:witness-trial-vector}
\end{equation}
In Step~5, we will test cap containment by comparing
\(\inner{\xhat_{t+1}}{\xhat^\star}\) with \(\cos\thetacap\).
Positive coefficients in \eqref{eq:witness-trial-vector} will let us
use the cap bound \(\inner{\xhat_t}{\xhat^\star}>\cos\thetacap\)
without reversing its inequality when taking the inner product with
\(\xhat^\star\). The condition \(0<\Lambda s_t<1\) ensures that both
coefficients are positive:
\[
 \Lambda s_t>0,\qquad
 1-\Lambda s_t\inner{\xhat_t}{\xhat^\star}
 \geq1-\Lambda s_t>0,
\]
where \(\inner{\xhat_t}{\xhat^\star}\leq1\).

To prove \(0<\Lambda s_t<1\) for \(s_t=M^\star(\omega_{1:T})\omega_t\),
first substitute \(M^\star\) from \eqref{eq:optimal-mass} and
\(G=\Lambda\thetacap\):
\begin{equation}
 \begin{aligned}
 \Lambda s_t
 &=\Lambda M^\star(\omega_{1:T})\omega_t\\
 &=\frac{\Lambda D\omega_t}
         {G\sqrt{(1+\thetacap)C_T(\omega_{1:T})}}\\
 &=\frac{D}{\thetacap\sqrt{1+\thetacap}}
     \frac{\omega_t}{\sqrt{C_T(\omega_{1:T})}}.
 \end{aligned}
 \label{eq:witness-coefficient-factorization}
\end{equation}
The factor depending on \(D\) and \(\thetacap\) is already below one:
\[
 0<\frac{D}{\thetacap\sqrt{1+\thetacap}}
 <\frac1{\sqrt{1+\thetacap}}<1,
\]
using \(0<D<\thetacap\) and \(\thetacap>0\).
By \eqref{eq:witness-coefficient-factorization}, it therefore suffices
to prove \(\omega_t/\sqrt{C_T(\omega_{1:T})}\leq1\).

To obtain this bound from the cost definition, use positivity and
normalization of \(\omega_{1:T}\). For each \(j<T\),
\[
 0<\sum_{r=j+1}^T\omega_r\leq\sum_{r=1}^T\omega_r=1,
 \qquad
 \frac{\omega_j^2}{\sum_{r=j+1}^T\omega_r}\geq\omega_j^2.
\]
Also, \(0<\omega_T\leq1\) gives \(\omega_T\geq\omega_T^2\).
Adding these inequalities yields
\[
 \begin{aligned}
 C_T(\omega_{1:T})
 &=\omega_T+\sum_{j=1}^{T-1}
                  \frac{\omega_j^2}{\sum_{r=j+1}^T\omega_r}\\
 &\geq\sum_{j=1}^T\omega_j^2\geq\omega_t^2,
 \qquad 1\leq t\leq T.
 \end{aligned}
\]
For \(T=1\), the sum over \(j<T\) is empty and the same inequality
follows from \(\omega_1=1\). Taking square roots and dividing by the
positive \(\sqrt{C_T(\omega_{1:T})}\) gives
\[
 0<\frac{\omega_t}{\sqrt{C_T(\omega_{1:T})}}\leq1.
\]
Substitution into \eqref{eq:witness-coefficient-factorization} now proves
\begin{equation}
 \begin{aligned}
 0<\Lambda s_t
 &=\frac{D}{\thetacap\sqrt{1+\thetacap}}
     \frac{\omega_t}{\sqrt{C_T(\omega_{1:T})}}\\
 &\leq\frac{D}{\thetacap\sqrt{1+\thetacap}}
 <\frac1{\sqrt{1+\thetacap}}<1.
 \end{aligned}
 \label{eq:witness-small-coefficients}
\end{equation}

\emph{Step 5: containment of the iterates and the oracle moment bound.}
Assume \(\xhat_t\in\mathcal C\).
The normalized update \eqref{eq:exact-finite-step} lies on the unit
sphere, and the cap is centered at \(\xhat^\star\).
By \eqref{eq:working-cap} and \eqref{eq:spherical-distance},
\[
 \begin{aligned}
 \xhat_{t+1}\in\mathcal C
 &\quad\Longleftrightarrow\quad
 \dist(\xhat_{t+1},\xhat^\star)<\thetacap\\
 &\quad\Longleftrightarrow\quad
 \inner{\xhat_{t+1}}{\xhat^\star}
 =\cos\!\bigl(\dist(\xhat_{t+1},\xhat^\star)\bigr)>\cos\thetacap.
 \end{aligned}
\]
We therefore need to lower-bound the inner product after normalization.
Taking the inner product of \eqref{eq:exact-finite-step} with
\(\xhat^\star\) gives
\[
 \begin{aligned}
 \inner{\xhat_{t+1}}{\xhat^\star}
 &=\frac{\inner{\xhat_t-s_t\gtilde_t}{\xhat^\star}}
         {\sqrt{1+s_t^2\norm{\gtilde_t}^2}}\\
 &=\frac{\inner{\xhat_t-s_t\gtilde_t}{\xhat^\star}}
         {\norm{\xhat_t-s_t\gtilde_t}}.
 \end{aligned}
\]
The second equality uses tangency from Step~3 and the normalized-update
identity in \Cref{lem:normalized-update-exp},
\(\norm{\xhat_t-s_t\gtilde_t}=\sqrt{1+s_t^2\norm{\gtilde_t}^2}>0\).
Thus the inequality we need is
\[
 \inner{\xhat_t-s_t\gtilde_t}{\xhat^\star}
 >\norm{\xhat_t-s_t\gtilde_t}\cos\thetacap.
\]

First bound \(\norm{\xhat_t-s_t\gtilde_t}\) using
\eqref{eq:witness-trial-vector}. Step~4 showed that both coefficients
in that expression are positive, so the triangle inequality gives
\begin{equation}
 \begin{aligned}
 \norm{\xhat_t-s_t\gtilde_t}
 &=\left\|
   \bigl(1-\Lambda s_t\inner{\xhat_t}{\xhat^\star}\bigr)\xhat_t
   +\Lambda s_t\xhat^\star\right\|\\
 &\leq
   \bigl(1-\Lambda s_t\inner{\xhat_t}{\xhat^\star}\bigr)\norm{\xhat_t}
   +\Lambda s_t\norm{\xhat^\star}\\
 &=1-\Lambda s_t\inner{\xhat_t}{\xhat^\star}+\Lambda s_t.
 \end{aligned}
 \label{eq:witness-trial-norm-bound}
\end{equation}
The last equality uses the unit norms of \(\xhat_t\) and \(\xhat^\star\).

Next bound \(\inner{\xhat_t-s_t\gtilde_t}{\xhat^\star}\).
Because \(\xhat_t\in\mathcal C\) and \(\thetacap>0\),
\[
 \inner{\xhat_t}{\xhat^\star}
 =\cos\!\bigl(\dist(\xhat_t,\xhat^\star)\bigr)>\cos\thetacap,
 \qquad
 \inner{\xhat^\star}{\xhat^\star}=1>\cos\thetacap.
\]
Multiplying these inequalities by the positive coefficients in
\eqref{eq:witness-trial-vector} yields
\[
 \begin{aligned}
 \inner{\xhat_t-s_t\gtilde_t}{\xhat^\star}
 &=\bigl(1-\Lambda s_t\inner{\xhat_t}{\xhat^\star}\bigr)
       \inner{\xhat_t}{\xhat^\star}+\Lambda s_t\\
 &>\bigl(1-\Lambda s_t\inner{\xhat_t}{\xhat^\star}\bigr)\cos\thetacap
       +\Lambda s_t\cos\thetacap\\
 &=\bigl(1-\Lambda s_t\inner{\xhat_t}{\xhat^\star}
       +\Lambda s_t\bigr)\cos\thetacap\\
 &\geq\norm{\xhat_t-s_t\gtilde_t}\cos\thetacap.
 \end{aligned}
\]
The last inequality uses \eqref{eq:witness-trial-norm-bound} and
\(\cos\thetacap\geq0\), since \(\thetacap\leq\pi/2\).
Dividing by the positive \(\norm{\xhat_t-s_t\gtilde_t}\) now gives
\[
 \begin{aligned}
 \inner{\xhat_{t+1}}{\xhat^\star}
 &=\frac{\inner{\xhat_t-s_t\gtilde_t}{\xhat^\star}}
         {\norm{\xhat_t-s_t\gtilde_t}}\\
 &>\frac{\norm{\xhat_t-s_t\gtilde_t}\cos\thetacap}
          {\norm{\xhat_t-s_t\gtilde_t}}
 =\cos\thetacap.
 \end{aligned}
\]
Hence \(\dist(\xhat_{t+1},\xhat^\star)<\thetacap\), proving
\(\xhat_{t+1}\in\mathcal C\).
The same argument includes \(\xhat_t=\xhat^\star\), since
\(\inner{\xhat^\star}{\xhat^\star}=1>\cos\thetacap\).
Starting from \(\xhat_1\in\mathcal C\) and repeating the implication
\(\xhat_t\in\mathcal C\Rightarrow\xhat_{t+1}\in\mathcal C\)
for \(t=1,\ldots,T\) proves containment through \(\xhat_{T+1}\).

We can now verify the second-moment condition in \Cref{ass:oracle}.
Step~3 established \eqref{eq:witness-gradient-bound} for points
\(\xhat\in\mathcal C\). Step~5 has shown that \(\xhat_t\in\mathcal C\)
for every \(1\leq t\leq T+1\). In particular, every point at which
the oracle is evaluated belongs to \(\mathcal C\). Substituting
\(\xhat=\xhat_t\) into \eqref{eq:witness-gradient-bound}, and using
the exact-gradient choice \eqref{eq:witness-oracle}, gives
\[
 \norm{\gtilde_t}
 =\norm{\nabla_{\mathbb S}L(\xhat_t)}
 \leq G,\qquad 1\leq t\leq T.
\]
Both sides are nonnegative, so squaring preserves the inequality:
\[
 \norm{\gtilde_t}^2
 =\norm{\nabla_{\mathbb S}L(\xhat_t)}^2
 \leq G^2.
\]
The same deterministic constant \(G=\Lambda\thetacap\) bounds every
oracle evaluation. Taking expectations therefore yields
\[
 \E\norm{\gtilde_t}^2
 \leq\E[G^2]
 =G^2,\qquad 1\leq t\leq T,
\]
which is the second-moment condition required by \Cref{ass:oracle}.

We have now checked all parts of \Cref{ass:oracle}.
Steps~1--2 established the loss properties and the minimizer
\(\xhat^\star\in\mathcal C\), and \(\xhat_1\in\mathcal C\) is
deterministic by \eqref{eq:witness-initialization}.
Step~3 verified tangency and conditional unbiasedness of \(\gtilde_t\).
Step~4 established that the coefficients \(s_t\) are positive and
deterministic. Step~5 proved containment through \(\xhat_{T+1}\),
which allowed us to verify the second-moment bound above.
Thus \Cref{thm:last-iterate} applies to
\(s_t=M^\star(\omega_{1:T})\omega_t\).
The argument used only positivity and unit sum of the fixed profile
\(\omega_{1:T}\), so it applies to every such profile, including
the optimizer \(\omega_{1:T}^\star\).
}%
\end{proof}

%% file: appendix-hyperball.tex
\section{Hyperball Geometry and Experimental Details}
\label[appendix]{app:hyperball}

\subsection{Geometric Correspondence}
\label[appendix]{app:hyperball-geometry}
Vectorization preserves both the Frobenius norm and inner product. Set
\(\xhat_t=\operatorname{vec}(\mathbf X_t)/R\) and
\(\bm v_t=\operatorname{vec}(\mathbf V_t)\). Hyperball keeps
\(\|\mathbf X_t\|_F=R\), so \(\|\xhat_t\|=1\); normalization of the base
update gives \(\|\bm v_t\|=1\). Dividing \eqref{eq:hyperball-wrapper}
by \(R\) therefore yields
\[
 \xhat_{t+1}=\frac{\xhat_t-\eta_t\bm v_t}
 {\|\xhat_t-\eta_t\bm v_t\|}.
\]
The denominator changes the length of the trial vector, but not its
direction. Thus \(\dphi_t\) is the angle between \(\xhat_t\) and
\(\xhat_t-\eta_t\bm v_t\). Write
\(c_t=\langle\xhat_t,\bm v_t\rangle\). Their inner product and the
squared length of the trial vector are
\[
 \langle\xhat_t,\xhat_t-\eta_t\bm v_t\rangle=1-\eta_tc_t,
 \qquad
 \|\xhat_t-\eta_t\bm v_t\|^2=1-2\eta_tc_t+\eta_t^2.
\]
For a nonzero trial vector, the cosine formula gives the realized angle:
\begin{equation}
 \dphi_t=\arccos\!\left(
 \frac{1-\eta_tc_t}{\sqrt{1-2\eta_tc_t+\eta_t^2}}\right).
 \label{eq:hyperball-general-angle}
\end{equation}
For a tangent direction, \(c_t=0\). The trial vector then has one unit
component along \(\xhat_t\) and an orthogonal component of length
\(\eta_t\). These are the adjacent and opposite sides of a right triangle
whose angle is \(\dphi_t\), so
\(\tan\dphi_t=\eta_t/1=\eta_t\). Since \(\eta_t\geq0\), the angle
lies in \([0,\pi/2)\), where tangent has a unique inverse. Therefore
\(\dphi_t=\arctan\eta_t\), as claimed in \eqref{eq:hyperball-rotation}.

Tangency also makes the displayed update identical to
\eqref{eq:exact-finite-step} when its coefficient is
\(s_t^{\mathrm{HB}}=\eta_t\) and its oracle is \(\gtilde_t=\bm v_t\).
Because \(\|\bm v_t\|=1\), this oracle satisfies the theorem's second-moment
bound with \(G=1\).

\subsection{Training Protocol}
\label[appendix]{app:hyperball-protocol}
We use the pretraining codebase of \citet{semenov2025benchmark}
for a 124M-parameter Llama model on FineWeb, with sequence length 512,
batch size 256, and 20,000 updates. The model has 12 layers, hidden width
768, MLP width 2048, bias-free linear projections, and tied input/output
embeddings.
We apply Hyperball to the attention and MLP projection matrices.
The embeddings and RMSNorm gains use Adam. Following \citet{semenov2025benchmark}'s
124M, large-batch recipe, Adam and AdamH use
\((\beta_1,\beta_2)=(0.8,0.999)\). Muon and MuonH use momentum 0.95,
Nesterov momentum, and five Newton--Schulz steps; all runs use global gradient-norm clipping at 0.5, FP32 parameters and
optimizer states, and BF16 computation.

Hyperball replaces the indirect norm control of weight decay with fixed parameter norms and normalized updates
\citep{wen2026hyperball}. Our default Adam and Muon comparisons therefore use
zero weight decay, allowing us to examine the rotational cooling supplied by
norm growth when this explicit control is removed.

We compare linear, cosine, warmup-stable-decay (WSD), and constant
schedules on the attention and MLP matrices. All schedules begin with
2,000 warmup updates; WSD cools over the final 20\% of the remaining
18,000 updates. The embeddings and normalization gains keep an Adam
cosine schedule with peak rate \(10^{-3}\).
For each optimizer and standard schedule in the zero-decay comparisons, we select the peak matrix learning rate
by final validation loss on one tuning seed. We then hold that rate
fixed for three independent evaluation seeds and report their mean
\(\pm\) one sample standard deviation. The learning-rate sweeps in
\Cref{fig:hyperball-lr-sensitivity} use the single-seed tuning runs.
We repeat this procedure with Adam and Muon
without Hyperball, tuning their matrix rates independently.

The zero-decay standard-profile learning-rate grids have successive values separated by a factor
\(\sqrt2\). All selected standard-profile rates
lie in the interior of their respective search grids.
\Cref{fig:hyperball-lr-sensitivity} shows the grids and selected rates;
\Cref{tab:hyperball-winners} also reports the rates for the constructed schedules of
\Cref{app:norm-aware-schedule}.

\begin{table}[!htbp]
 \centering
 \small
 \begin{tabular}{lrrrr}
  \toprule
  Schedule & AdamH & MuonH & Adam & Muon \\
  \midrule
  Linear & 0.01414 & 0.01000 & 0.00100 & 0.01000 \\
  Cosine & 0.01414 & 0.01000 & 0.00141 & 0.01000 \\
  WSD & 0.00707 & 0.00500 & 0.00100 & 0.01000 \\
  Constant & 0.00354 & 0.00177 & 0.00141 & 0.00707 \\
  Matrix-wise & -- & -- & 0.00368 & 0.01200 \\
  Shared LR (ablation) & -- & -- & 0.00130 & 0.01200 \\
  \bottomrule
 \end{tabular}
 \caption{Peak matrix learning rates used in the three-seed comparisons.
 The matrix-wise entry is the maximum across matrices and updates.
 Selection of the constructed schedules' rates is described in
 \Cref{app:norm-aware-schedule}.}
 \label{tab:hyperball-winners}
\end{table}

\subsection{Additional Pretraining Comparisons}
\begin{figure}[!htbp]
 \centering
 \includegraphics[width=.95\linewidth]{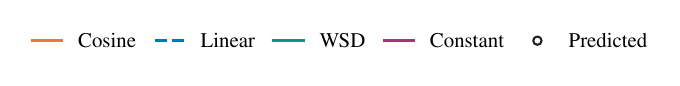}
 \begin{subfigure}{.48\linewidth}
  \includegraphics[width=\linewidth]{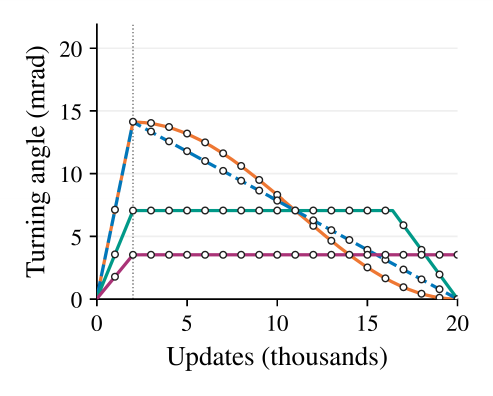}\caption{}\label{fig:hyperball-rotation-adamh}
 \end{subfigure}\hfill
 \begin{subfigure}{.48\linewidth}
  \includegraphics[width=\linewidth]{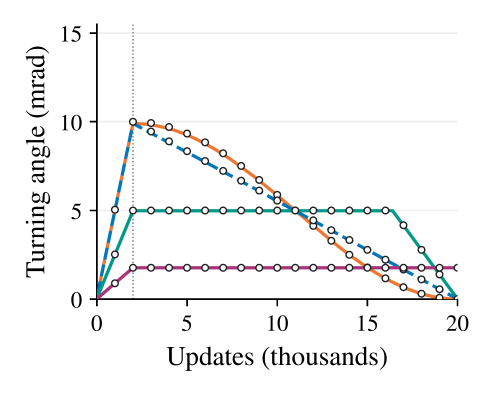}\caption{}
 \end{subfigure}
 \caption{Measured turning angles for \textbf{(a)} AdamH and \textbf{(b)}
 MuonH: mean \(\pm\) one sample standard deviation across three evaluation
 seeds. Open circles show the tangent-update prediction
 \(\arctan\eta_t\) from \eqref{eq:hyperball-rotation}.}
 \label{fig:hyperball-rotation}
\end{figure}

\label[appendix]{app:hyperball-results}
\Cref{fig:hyperball-rotation} compares the measured turning angles with
\(\arctan\eta_t\). Their agreement across all four schedules supports the
approximately tangent update geometry of \eqref{eq:hyperball-rotation}.
In particular, WSD's constant-rate phase leaves rotation constant
under Hyperball, whereas linear and cosine decay it throughout training.

\begin{figure}[!htbp]
 \centering
 \includegraphics[width=\linewidth]{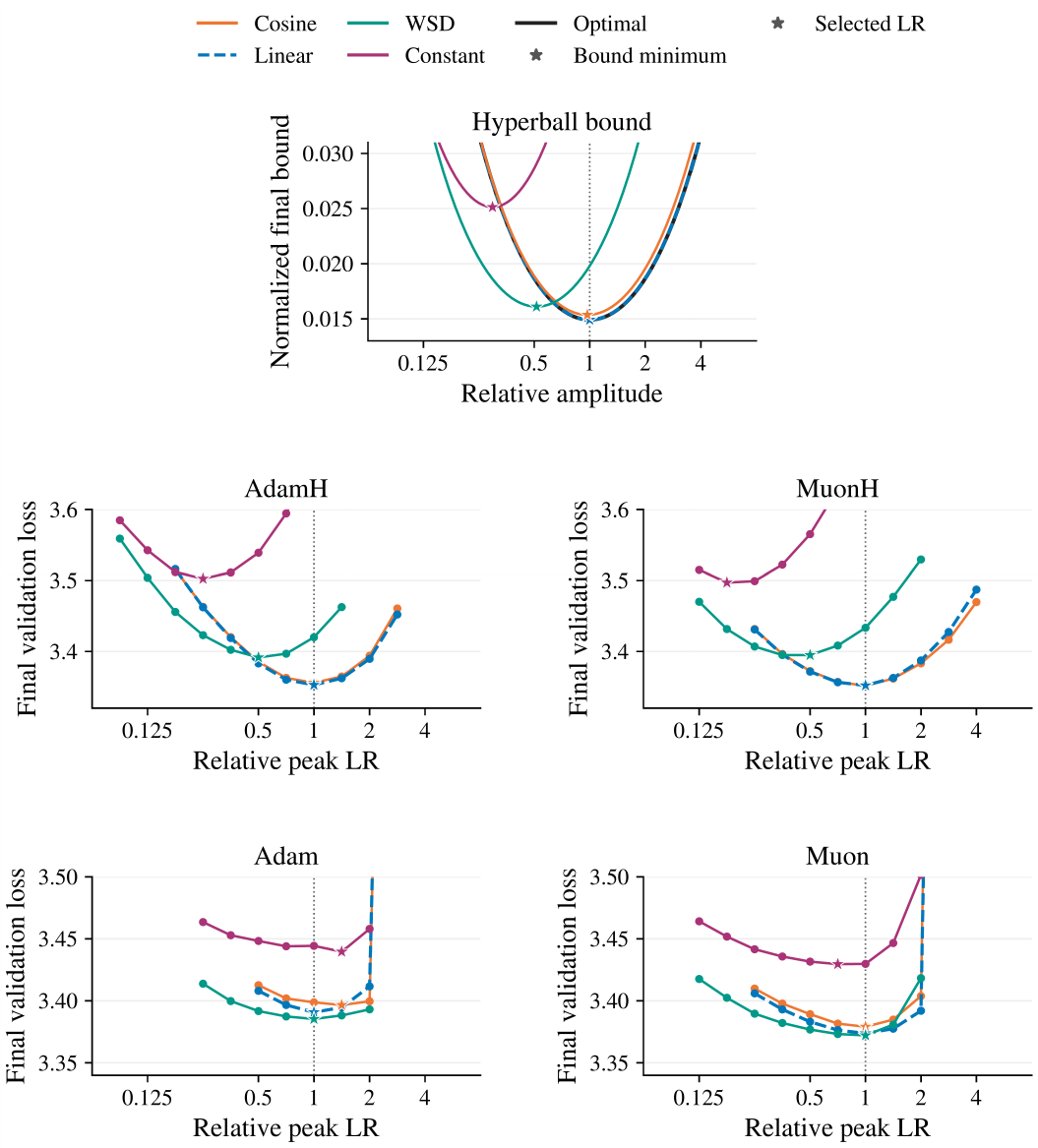}
 \caption{Learning-rate sensitivity for the four standard schedules.
 Top: the Hyperball bound over the 18,000 post-warmup updates.
 Below: final validation loss for Adam and Muon, with Hyperball in the
 middle row and without it in the bottom row. Horizontal axes are relative
 to linear's analytical optimum (top) or selected peak learning rate (below);
 dotted lines mark this reference at one. Stars mark analytical bound minima and
 selected learning rates.}
 \label{fig:hyperball-lr-sensitivity}
\end{figure}

Under Hyperball, the bound captures the dependence on learning rate across
the tuning sweeps, as well as the comparison at the selected rates
(\Cref{fig:hyperball-lr-sensitivity}). Linear and cosine have similar minima
and nearly overlapping sensitivity curves. WSD prefers a smaller peak
rate and ends at a higher loss, while constant prefers a still smaller
rate and has the highest minimum. The bound exhibits the same pattern.

\begin{figure}[!htbp]
 \centering
 \includegraphics[width=.95\linewidth]{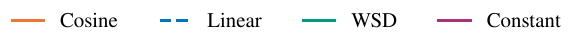}
 \begin{subfigure}{.46\linewidth}
  \includegraphics[width=\linewidth]{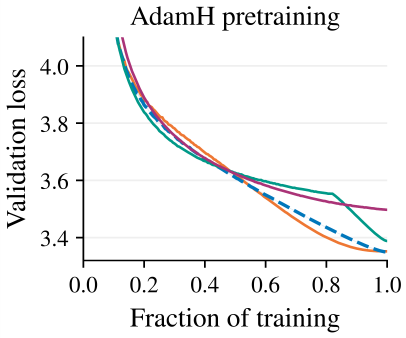}\caption{}
 \end{subfigure}\hfill
 \begin{subfigure}{.46\linewidth}
  \includegraphics[width=\linewidth]{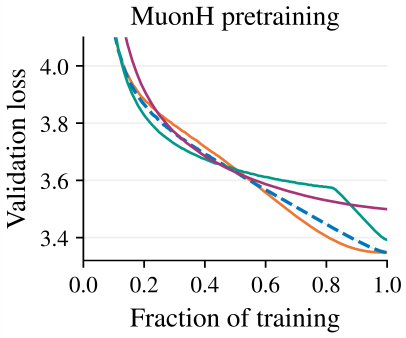}\caption{}
 \end{subfigure}
 \caption{124M Hyperball pretraining with \textbf{(a)} AdamH and \textbf{(b)} MuonH.
 Curves and bands show validation-loss means and one sample standard deviation
 across three evaluation seeds at independently tuned peak rates.
 Linear and cosine end at similar losses, both lower than those of WSD and the constant schedule.}
 \label{fig:hyperball-validation}
\end{figure}

\begin{figure}[!htbp]
 \centering
 \begin{subfigure}{.43\linewidth}
  \includegraphics[width=\linewidth]{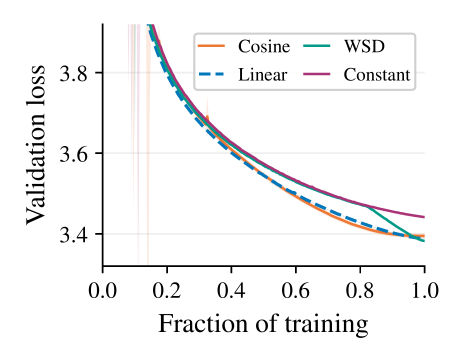}\caption{}
 \end{subfigure}\hspace{.035\linewidth}
 \begin{subfigure}{.43\linewidth}
  \includegraphics[width=\linewidth]{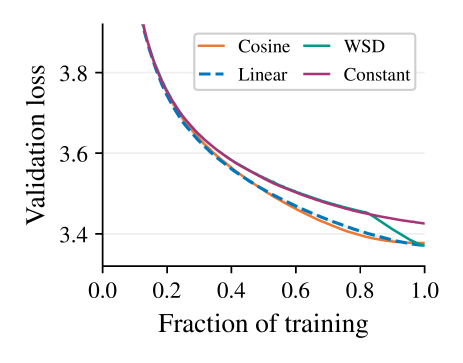}\caption{}
 \end{subfigure}
 \caption{Validation trajectories for \textbf{(a)} Adam and
 \textbf{(b)} Muon without Hyperball or weight decay: mean \(\pm\) one
 sample standard deviation across three evaluation seeds at the selected rates.
 After cooldown, WSD ends at a lower loss than cosine in both panels,
 reversing their ordering under Hyperball.
 }
 \label{fig:hyperball-unconstrained-loss}
\end{figure}

Without Hyperball, the sensitivity curves change. Linear and cosine
improve little over part of the rate range and deteriorate sharply at
larger rates, while WSD reaches a lower minimum. At the independently
selected rates, WSD ends at a lower loss than cosine after cooldown for both
optimizers (\Cref{fig:hyperball-unconstrained-loss}). Its final loss
is also lower than linear's for Adam and close to linear's for Muon, while
the constant schedule has a higher final loss. Increasing parameter norms can cool rotation
during WSD's constant-rate phase, before its nominal cooldown begins.
{\color{black}%
In \Cref{app:rotation-diagnostic}, we use a norm model to predict this
rotational schedule and evaluate our bound on it.
}%

\FloatBarrier
\subsection{\texorpdfstring{\textcolor{black}{Containment throughout Training}}{Containment throughout Training}}
\label[appendix]{app:hyperball-containment}
{\color{black}%
We check whether the observed trajectories remain in fixed spherical caps by fixing
each matrix's cap center at its initial direction and taking
\(\thetacap=\pi/2\). A suitable fixed center suffices for geometric
containment.
Using the 124M training protocol of \Cref{app:hyperball-protocol} and the
selected peak rates in \Cref{tab:hyperball-winners}, we run AdamH and MuonH
with linear, cosine, and WSD schedules.
We measure the angle to initialization for all 60 matrices at update zero
and after every update through 20,000.

All measured trajectories remain inside their initialization-centered
open hemispheres.

}%

\begin{figure}[!htbp]
 \centering
 \includegraphics[width=\linewidth]{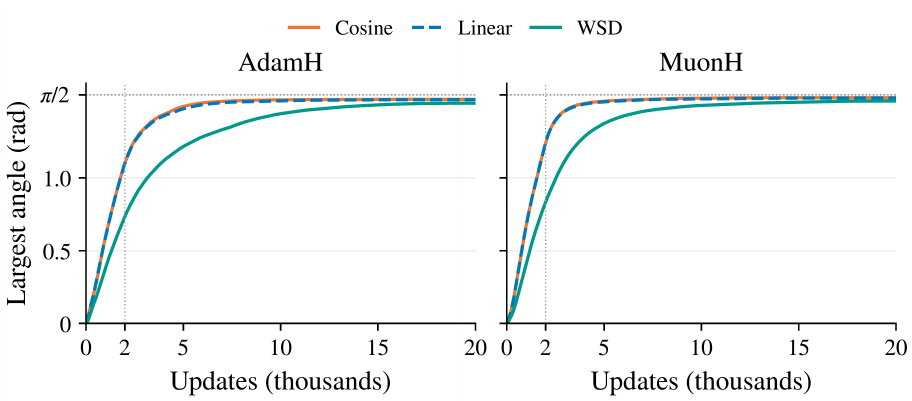}
 \caption{\textcolor{black}{Angles from initialization for AdamH (left) and MuonH
 (right). Each curve shows the largest angle across the 60 matrices at
 update zero and after every update through 20,000.
 The horizontal dotted line marks \(\pi/2\); the vertical dotted line marks
 the end of the 2,000-update warmup.}}
 \label{fig:hyperball-cap-containment}
\end{figure}

\FloatBarrier
\clearpage
\subsection{Local Scale Sensitivity along Training}
\label[appendix]{app:local-scale-sensitivity}
Exact scale invariance leaves the loss unchanged when one parameter group
is multiplied by a positive scalar. We examine how closely the trained
matrices exhibit this property locally. This complements the turning-angle
measurements in \Cref{fig:hyperball-rotation}: those describe the motion
produced by the optimizer, whereas here we measure the loss gradient itself.

Consider one matrix \(\mathbf X_t\). Write \(\ell_t(\mathbf X)\) for the
average loss on the minibatch used at update \(t\), with the other
parameters fixed at their current values. Multiplying
\(\mathbf X_t\) by \(\kappa>0\) changes its norm while preserving its
direction. Each matrix entry changes at rate
\(\frac{\mathrm d}{\mathrm d\kappa}(\kappa(\mathbf X_t)_{ij})
=(\mathbf X_t)_{ij}\). Summing the contributions of all entries, the
chain rule gives
\[
 \begin{aligned}
 \frac{\mathrm d}{\mathrm d\kappa}\ell_t(\kappa\mathbf X_t)
 &=\sum_{i,j}\left.\frac{\partial\ell_t(\mathbf X)}{\partial X_{ij}}\right|_{\mathbf X=\kappa\mathbf X_t}
     \frac{\mathrm d}{\mathrm d\kappa}(\kappa(\mathbf X_t)_{ij})\\
 &=\sum_{i,j}\left.\frac{\partial\ell_t(\mathbf X)}{\partial X_{ij}}\right|_{\mathbf X=\kappa\mathbf X_t}
     (\mathbf X_t)_{ij}\\
 &=\langle\nabla_{\mathbf X}\ell_t(\kappa\mathbf X_t),\mathbf X_t\rangle_F.
 \end{aligned}
\]
The last equality is the definition of the Frobenius inner product.
Evaluating at the current scale \(\kappa=1\) yields
\begin{equation}
 \left.\frac{\mathrm d}{\mathrm d\kappa}\ell_t(\kappa\mathbf X_t)\right|_{\kappa=1}
 =\langle\nabla_{\mathbf X}\ell_t(\mathbf X_t),\mathbf X_t\rangle_F.
 \label{eq:local-scale-derivative}
\end{equation}
To interpret this derivative, expand the loss as a function of the scalar
\(\kappa\) around one. The first-order approximation is
\[
 \ell_t(\kappa\mathbf X_t)\approx\ell_t(\mathbf X_t)
 +(\kappa-1)\left.\frac{\mathrm d}{\mathrm d\kappa}
 \ell_t(\kappa\mathbf X_t)\right|_{\kappa=1}.
\]
Substituting \eqref{eq:local-scale-derivative} and subtracting the current
loss \(\ell_t(\mathbf X_t)\) gives
\begin{equation}
 \ell_t(\kappa\mathbf X_t)-\ell_t(\mathbf X_t)
 \approx(\kappa-1)\langle\nabla_{\mathbf X}\ell_t(\mathbf X_t),\mathbf X_t\rangle_F.
 \label{eq:local-scale-loss-change}
\end{equation}
The absolute inner product therefore measures local loss sensitivity to a
relative change in matrix norm. For example, multiplying a matrix by
\(1.01\) increases its norm by 1\% without changing its direction.
Substituting \(\kappa=1.01\) gives
\[
 \ell_t(1.01\mathbf X_t)-\ell_t(\mathbf X_t)
 \approx0.01\langle\nabla_{\mathbf X}\ell_t(\mathbf X_t),\mathbf X_t\rangle_F.
\]
An absolute inner product of \(0.001\) therefore predicts a first-order
loss change of magnitude \(0.00001\) nats. The same percentage change
can be applied to both small-norm and large-norm matrices; adding the
same amount to their norms would instead give different proportional
changes.

We also measure how much of the gradient points in the radial direction.
The unit radial direction is \(\mathbf X_t/\|\mathbf X_t\|_F\), so the
magnitude of the gradient's radial component is
\( |\langle\nabla_{\mathbf X}\ell_t(\mathbf X_t),\mathbf X_t\rangle_F|/
\|\mathbf X_t\|_F\). Dividing by the gradient norm gives
\begin{equation}
 \frac{|\langle\nabla_{\mathbf X}\ell_t(\mathbf X_t),\mathbf X_t\rangle_F|}
 {\|\nabla_{\mathbf X}\ell_t(\mathbf X_t)\|_F\|\mathbf X_t\|_F}.
 \label{eq:local-radial-alignment}
\end{equation}
This is the fraction of gradient magnitude in the radial component.
Small values indicate predominantly tangent gradients.

Under exact scale invariance, the absolute loss sensitivity in
\eqref{eq:local-scale-derivative} and the radial gradient fraction in
\eqref{eq:local-radial-alignment} both vanish because their common numerator is zero
(the normalized quantity requires a nonzero gradient). For approximate
behavior, their magnitudes provide complementary information. A small
gradient can make the loss sensitivity small even when its direction is
substantially radial. Conversely, a large gradient or parameter norm can
make the loss sensitive despite a small radial fraction. We measure both
quantities to check whether the loss is locally insensitive to rescaling
and whether the gradient is predominantly tangent.

\begin{figure}[!ht]
 \centering
 \includegraphics[width=\linewidth]{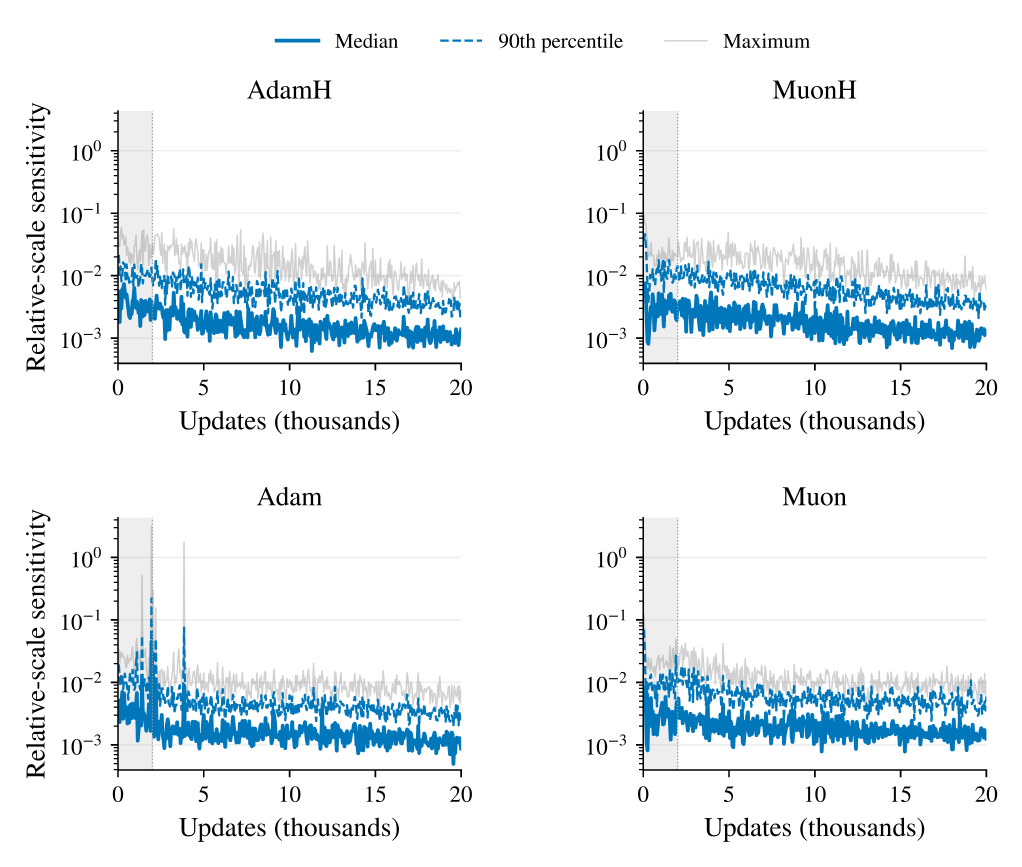}
 \caption{Local sensitivity to relative matrix rescaling:
 \( |\langle\nabla_{\mathbf X}\ell_t(\mathbf X_t),\mathbf X_t\rangle_F|\).
 AdamH and MuonH are above Adam and Muon. Curves show the median,
 90th percentile, and maximum across 60 matrices, sampled every 50 updates
 in one seed-zero linear-schedule run per optimizer. Shading marks the
 2,000-update warmup.}
 \label{fig:local-scale-sensitivity}
\end{figure}

\begin{figure}[!ht]
     \centering
     \includegraphics[width=\linewidth]{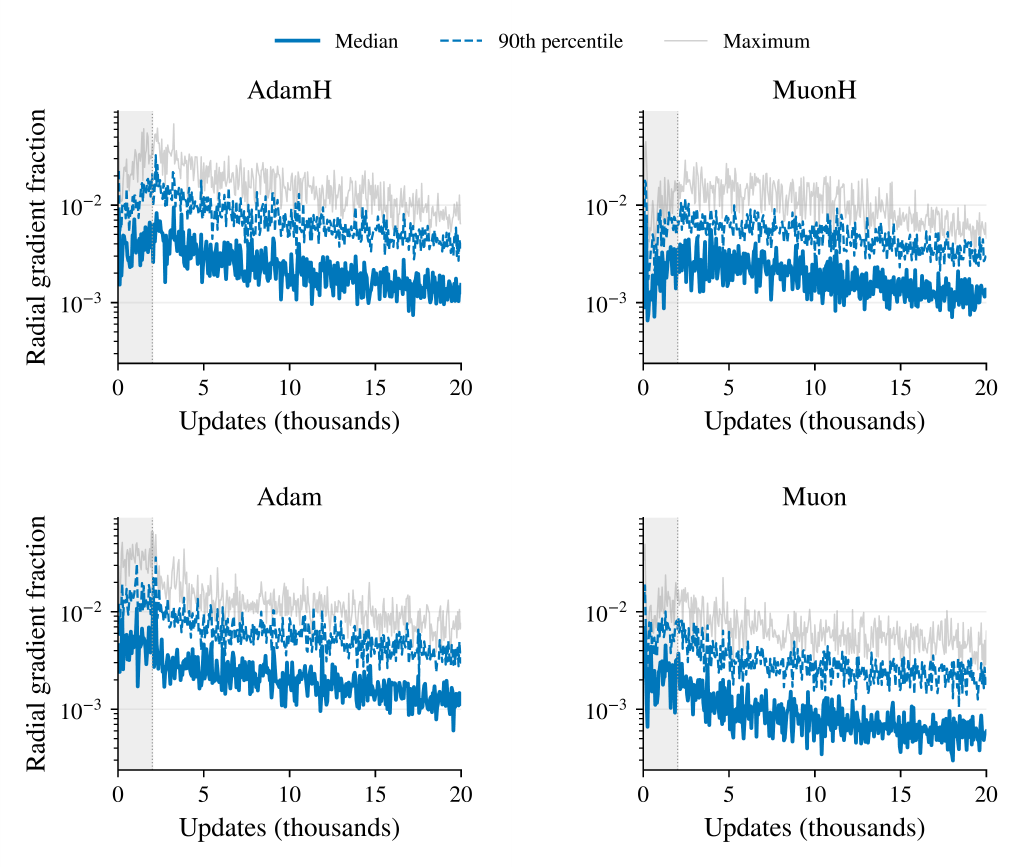}
     \caption{Radial fraction of the raw loss gradient, defined in
     \eqref{eq:local-radial-alignment}.
     Curves show the median, 90th percentile, and maximum across 60 matrices,
     sampled before clipping every 50 updates in one seed linear-schedule
     run per optimizer. Shading marks warmup.}
 \label{fig:local-radial-alignment}
\end{figure}

We repeat the 124M training protocol of \Cref{app:hyperball-protocol} with
AdamH, MuonH, Adam, and Muon and the selected linear-schedule
rates in \Cref{tab:hyperball-winners}. Every 50 updates, we measure the raw
accumulated gradients before clipping and optimizer transformations for
all 60 attention and MLP matrices. At each sampled update, we take the
absolute per-matrix quantities and report their median, 90th percentile,
and maximum across matrices. \Cref{fig:local-scale-sensitivity,fig:local-radial-alignment}
retain all 400 samples without temporal smoothing.

\clearpage

Across all four optimizers, typical matrices become less sensitive to
relative rescaling as training progresses, while their gradients become
more nearly tangent. The agreement between these two measurements shows
that the small radial derivatives are accompanied by predominantly
directional gradients. This behavior occurs both with and without
Hyperball, supporting the use of a spherical description beyond explicitly
fixed-norm training.

The turning-angle measurements establish how learning rates control
parameter rotation. These gradient measurements provide complementary
evidence that the sampled loss is locally insensitive to changes in matrix
norm along the observed trajectories.

\FloatBarrier
\subsection{Predicted and Measured Rotational Profiles}
\label[appendix]{app:rotation-diagnostic}
To predict rotation from a learning-rate schedule without Hyperball, we
first calculate the optimizer's update power and correlations between
updates. These determine the parameter norm through a finite-history
recurrence. Dividing the update displacement by this norm gives a
rotational coefficient, whose normalized profile we use to evaluate the
bound. \Cref{fig:unconstrained-rotation-shapes} compares the predicted
norms and rotation with measurements; \Cref{fig:unconstrained-bound-loss,fig:bound-contrast}
compare the resulting schedule ordering with validation loss.

\subsubsection{Norm growth}
Predicting rotation requires the parameter norm as well as the update
size. The norm depends on both the power of the current update and its
correlations with previous updates. Following the finite-history model of
\citet[App.~E.2]{anonymous2026how}, write
\[
 R_t^2:=\mathbb E\|\mathbf X_t\|_F^2,\qquad
 \Gamma_{t,r}:=\frac{\mathbb E\langle\mathbf U_t,\mathbf U_r\rangle_F}{C_a(0)},
 \qquad P_t:=\Gamma_{t,t}.
\]
Here \(C_a(0)=d_{\rm out}d_{\rm in}\) is the total power of the
standardized gradients used below. The quantity \(\Gamma_{t,r}\)
is the expected inner product of two updates divided by this input
power. Its diagonal \(P_t\) therefore gives the current update power:
\begin{equation}
 C_a(0)P_t
 =\mathbb E\langle\mathbf U_t,\mathbf U_t\rangle_F
 =\mathbb E\|\mathbf U_t\|_F^2.
 \label{eq:predicted-update-power}
\end{equation}
For deterministic learning rates and weight-decay coefficients, consider
the parameter update
\[
 \mathbf X_{t+1}=\delta_t\mathbf X_t-\eta_t\mathbf U_t,\qquad
 \delta_t:=1-\eta_t\lambda_t\in(0,1],\qquad
 W_{i,t}:=\prod_{j=i}^{t-1}\delta_j,\quad W_{t,t}=1.
\]
The factor \(W_{i,t}\) is the contraction accumulated between steps
\(i\) and \(t\). Assume that the initialization has zero expected
overlap with each modeled update,
\(\mathbb E\langle\mathbf X_1,\mathbf U_t\rangle_F=0\).
Unrolling the parameter update gives
\[
 \mathbf X_t=W_{1,t}\mathbf X_1-
       \sum_{r=1}^{t-1}W_{r+1,t}\eta_r\mathbf U_r.
\]
Taking its expected inner product with the current update then yields
\[
 \mathbb E\langle\mathbf X_t,\mathbf U_t\rangle_F
 =-C_a(0)\sum_{r=1}^{t-1}W_{r+1,t}\eta_r\Gamma_{t,r}.
\]
Earlier updates displaced the parameters in direction
\(-\mathbf U_r\), so a current update aligned with them has a negative
inner product with those displacements. Substituting this identity and
\eqref{eq:predicted-update-power} into the squared-norm expansion
\[
 R_{t+1}^2=\delta_t^2R_t^2
 -2\delta_t\eta_t\mathbb E\langle\mathbf X_t,\mathbf U_t\rangle_F
 +\eta_t^2\mathbb E\|\mathbf U_t\|_F^2
\]
gives the recurrence of
\citet[App.~E.2, Eqs.~E.11--E.14]{anonymous2026how}:
\begin{equation}
 R_{t+1}^2=\delta_t^2R_t^2+C_a(0)\left[
 \eta_t^2P_t+2\delta_t\eta_t
 \sum_{r=1}^{t-1}W_{r+1,t}\eta_r\Gamma_{t,r}\right].
 \label{eq:predicted-norm-recursion}
\end{equation}
The first term contracts the previous squared radius. Inside the bracket,
current update power and correlations with earlier updates both contribute
to norm growth; positive correlations reinforce earlier displacements.
Under the stated initialization assumption, the updates enter this
recurrence only through \(P_t\) and \(\Gamma_{t,r}\). It therefore
covers both linear momentum and Muon's nonlinear transformation. We next
calculate these two statistics for Adam and Muon. For the zero-decay
comparisons below, \(\delta_t=W_{i,t}=1\); \Cref{app:weight-decay}
returns to the general recurrence.

\subsubsection{Adam}
For Adam, we obtain \(P_t\) and \(\Gamma_{t,r}\) directly from a
linear momentum representation. We use the noise-dominated model of
\citet[Sec.~3.2]{kosson2024rotational}.
Following \citet[Eq.~18]{wen2026hyperball}, we approximate Adam's RMS
denominator by the coordinate's fixed gradient standard deviation
\(\sigma>0\), omitting bias correction and the numerical stabilizer.
Let \(a_t\) be the vectorized gradient divided by \(\sigma\), so each
coordinate has unit variance. These standardized gradients are centered
and uncorrelated across steps:
\[
 \mathbb E a_t=0,\qquad \mathbb E\|a_t\|^2=C_a(0),\qquad
 \mathbb E\langle a_t,a_r\rangle=0\quad(t\ne r).
\]
For the zero-start EMA with \(\beta=0.8\),
\[
 E_{t,i}=(1-\beta)\beta^{t-i},\qquad
 \operatorname{vec}(\mathbf U_t)=\sum_{i=1}^t E_{t,i}a_i.
\]
Two momentum updates are correlated because they contain some of the
same past standardized gradients. Expanding their expected inner product
shows which gradients contribute:
\begin{equation}
 \begin{aligned}
 \mathbb E\langle\mathbf U_t,\mathbf U_r\rangle_F
 &=\sum_{i=1}^t\sum_{j=1}^r E_{t,i}E_{r,j}
                         \mathbb E\langle a_i,a_j\rangle\\
 &=C_a(0)\sum_{i=1}^{\min(t,r)}E_{t,i}E_{r,i}.
 \end{aligned}
 \label{eq:predicted-execution-correlation}
\end{equation}
The terms with \(i\ne j\) vanish by the uncorrelated-gradient
assumption; each matching index contributes \(C_a(0)\). For \(t\geq r\),
substituting the EMA coefficients and summing the geometric series gives
\begin{equation}
 \begin{aligned}
 \Gamma_{t,r}
 &=(1-\beta)^2\beta^{t-r}\sum_{i=1}^r\beta^{2(r-i)}\\
 &=(1-\beta)^2\beta^{t-r}\frac{1-\beta^{2r}}{1-\beta^2}\\
 &=\frac{1-\beta}{1+\beta}\beta^{t-r}(1-\beta^{2r}),
 \qquad P_t=\Gamma_{t,t}.
 \end{aligned}
 \label{eq:predicted-adam-moments}
\end{equation}

\subsubsection{Muon}
For Muon, we calculate the same \(P_t\) and \(\Gamma_{t,r}\) as for
Adam: the current update power and the expected inner product of two
updates, each divided by \(C_a(0)\). The difference is that Muon applies
matrix sign to its momentum before taking a step. We must account for
this transformation when calculating the update inner product.

Let \(\mathbf G_t\) have independent standard Gaussian entries,
independently across steps. With \(\beta=0.95\), the stored buffer starts
at zero and the Nesterov momentum passed to matrix sign is
\[
 \mathbf M_t=\beta\mathbf M_{t-1}+\mathbf G_t,\qquad \mathbf M_0=0,
 \qquad \widetilde{\mathbf M}_t=\mathbf G_t+\beta\mathbf M_t.
\]
The modeled update is
\[
 \mathbf U_t=s_\mu\operatorname{msign}(\widetilde{\mathbf M}_t),
 \qquad s_\mu=\sqrt{\max(1,d_{\rm out}/d_{\rm in})}.
\]
\citet[Lemma~4.3]{wen2026hyperball} calculate update correlations for
stationary ordinary momentum. For \(t>r\), its later value is
\(\beta^{t-r}\) times its earlier value plus independent contributions
from subsequent gradients. Equal stationary variances then give
correlation \(\beta^{t-r}\). Here the buffer starts at zero and matrix
sign receives the Nesterov momentum \(\widetilde{\mathbf M}_t\).
We adapt their matrix-sign calculation by deriving the variances and
correlations of these finite-history inputs.

For a compact SVD \(\mathbf A=\mathbf L\boldsymbol\Sigma\mathbf V^\top\),
matrix sign is \(\operatorname{msign}(\mathbf A)=\mathbf L\mathbf V^\top\):
it keeps the singular vectors and replaces the nonzero singular values
by one. We use this exact map to model the five Newton--Schulz steps
used in training.

Each \(\widetilde{\mathbf M}_t\) has independent Gaussian entries with
positive variance: its current-gradient term is \((1+\beta)\mathbf G_t\).
Such Gaussian matrices have full rank almost surely
\citep[Sec.~9.3.1, p.~269]{vershynin2025highdimensional}, so each has
\(\min(d_{\rm out},d_{\rm in})\) nonzero singular values. The
squared Frobenius norm is the sum of squared singular values. After
matrix sign, this gives
\begin{equation}
 \|\operatorname{msign}(\widetilde{\mathbf M}_t)\|_F^2
 =\underbrace{1^2+\cdots+1^2}_{\min(d_{\rm out},d_{\rm in})\text{ terms}}
 =\min(d_{\rm out},d_{\rm in}).
 \label{eq:predicted-sign-norm}
\end{equation}
Including the update scale,
\[
 \|\mathbf U_t\|_F^2
 =\begin{cases}
  (d_{\rm out}/d_{\rm in})d_{\rm in}=d_{\rm out},
       &d_{\rm out}\geq d_{\rm in},\\
  1\cdot d_{\rm out}=d_{\rm out},&d_{\rm out}<d_{\rm in}.
 \end{cases}
 \qquad
 P_t=\frac{\mathbb E\|\mathbf U_t\|_F^2}{C_a(0)}
     =\frac{1}{d_{\rm in}}.
\]
Thus Muon's modeled update power is constant from the first step.

To calculate \(\Gamma_{t,r}\), we first determine how the two momentum
matrices depend on the same past gradients. Unrolling the buffer gives
\(\mathbf M_t=\sum_{i=1}^t\beta^{t-i}\mathbf G_i\), and hence
\begin{equation}
 \widetilde{\mathbf M}_t=\sum_{i=1}^tE_{t,i}\mathbf G_i,\qquad
 E_{t,i}=\begin{cases}
  1+\beta,&i=t,\\
  \beta^{t-i+1},&1\leq i<t.
 \end{cases}
 \label{eq:predicted-nesterov-coefficients}
\end{equation}
Here \(E_{t,i}\) denotes the coefficients before matrix sign. For the
entry in row \(j\), column \(k\), independence, zero mean, and unit
variance imply
\begin{equation}
 \mathbb E[(\mathbf G_i)_{jk}(\mathbf G_\ell)_{jk}]
 =\begin{cases}1,&i=\ell,\\0,&i\ne\ell.\end{cases}
 \label{eq:predicted-gaussian-gradient-moments}
\end{equation}
Expanding the product of the momentum entries at that same row and
column, then applying \eqref{eq:predicted-gaussian-gradient-moments}, gives
\begin{equation}
 \begin{aligned}
 \mathbb E[(\widetilde{\mathbf M}_t)_{jk}
           (\widetilde{\mathbf M}_r)_{jk}]
 &=\sum_{i=1}^t\sum_{\ell=1}^r
 E_{t,i}E_{r,\ell}\,
 \mathbb E[(\mathbf G_i)_{jk}(\mathbf G_\ell)_{jk}]\\
 &=\sum_{i=1}^{\min(t,r)}E_{t,i}E_{r,i}.
 \end{aligned}
 \label{eq:predicted-nesterov-entry-product}
\end{equation}
Only the same gradient entry from the same step contributes.

For example, the momenta at steps two and three are
\[
 \widetilde{\mathbf M}_2=(1+\beta)\mathbf G_2+\beta^2\mathbf G_1,
 \qquad
 \widetilde{\mathbf M}_3=(1+\beta)\mathbf G_3+
                         \beta^2\mathbf G_2+\beta^3\mathbf G_1.
\]
Their entrywise product expands as
\[
 \begin{aligned}
 (\widetilde{\mathbf M}_2)_{jk}(\widetilde{\mathbf M}_3)_{jk}
 ={}&(1+\beta)^2(\mathbf G_2)_{jk}(\mathbf G_3)_{jk}
     +(1+\beta)\beta^2(\mathbf G_2)_{jk}^2\\
 &+(1+\beta)\beta^3(\mathbf G_2)_{jk}(\mathbf G_1)_{jk}
     +\beta^2(1+\beta)(\mathbf G_1)_{jk}(\mathbf G_3)_{jk}\\
 &+\beta^4(\mathbf G_1)_{jk}(\mathbf G_2)_{jk}
     +\beta^5(\mathbf G_1)_{jk}^2.
 \end{aligned}
\]
Applying \eqref{eq:predicted-gaussian-gradient-moments} to each term
replaces the products from different steps by zero and the squared
entries by their unit second moment:
\[
 \begin{aligned}
 \mathbb E[(\widetilde{\mathbf M}_2)_{jk}
           (\widetilde{\mathbf M}_3)_{jk}]
 ={}&(1+\beta)^2\cdot0+(1+\beta)\beta^2\cdot1\\
 &+(1+\beta)\beta^3\cdot0+\beta^2(1+\beta)\cdot0\\
 &+\beta^4\cdot0+\beta^5\cdot1\\
 ={}&(1+\beta)\beta^2+\beta^5.
 \end{aligned}
\]
The two contributions come from \(\mathbf G_2\) and \(\mathbf G_1\),
which occur in both momentum matrices.

Setting \(r=t\) in \eqref{eq:predicted-nesterov-entry-product} gives
the variance of each momentum entry:
\begin{equation}
 \begin{aligned}
 \nu_t:=\mathbb E[(\widetilde{\mathbf M}_t)_{jk}^2]
 &=\sum_{i=1}^t E_{t,i}^2
 =(1+\beta)^2+\sum_{i=1}^{t-1}\beta^{2(t-i+1)}\\
 &=(1+\beta)^2+\frac{\beta^4(1-\beta^{2(t-1)})}{1-\beta^2}.
 \end{aligned}
 \label{eq:predicted-nesterov-variance}
\end{equation}
To obtain the correlation between momenta at two different updates,
we next calculate the expected product of their entries in the same
row \(j\), column \(k\). Both entries have zero mean, so this
expectation is their covariance. Fix \(t>r\). The earlier momentum
contains gradients only through \(\mathbf G_r\), and these also occur
in the later momentum. The newer gradients
\(\mathbf G_{r+1},\ldots,\mathbf G_t\) are independent of all gradients
in the earlier momentum. Their products with its entries have zero
expectation by \eqref{eq:predicted-gaussian-gradient-moments}.
Thus \eqref{eq:predicted-nesterov-entry-product} sums only over the
shared gradients, from step 1 through step \(r\).

The newest shared gradient, \(\mathbf G_r\), has different roles in
the two momenta. At update \(r\), it is the current gradient and has
coefficient \(E_{r,r}=1+\beta\). At update \(t>r\), it is a past
gradient and has coefficient \(E_{t,r}=\beta^{t-r+1}\).
For every older shared gradient \(\mathbf G_i\), with \(i<r\), both
coefficients use the past-gradient case of
\eqref{eq:predicted-nesterov-coefficients}. Separating the term \(i=r\)
and substituting these coefficients gives
\[
 \begin{aligned}
 \mathbb E[(\widetilde{\mathbf M}_t)_{jk}
           (\widetilde{\mathbf M}_r)_{jk}]
 &=\sum_{i=1}^{r}E_{t,i}E_{r,i}\\
 &=E_{t,r}E_{r,r}+\sum_{i=1}^{r-1}E_{t,i}E_{r,i}\\
 &=\beta^{t-r+1}(1+\beta)
   +\sum_{i=1}^{r-1}\beta^{t-i+1}\beta^{r-i+1}.
 \end{aligned}
\]
For an older gradient, its coefficient at update \(t\) contains
\(t-r\) additional powers of \(\beta\) compared with its coefficient
at update \(r\):
\[
 \beta^{t-i+1}
 =\beta^{t-r}\beta^{r-i+1},\qquad
 \beta^{t-i+1}\beta^{r-i+1}
 =\beta^{t-r}\beta^{2(r-i+1)}.
\]
The first term also contains \(\beta^{t-r}\), since
\(\beta^{t-r+1}=\beta^{t-r}\beta\). Factoring it out leaves
\[
 \mathbb E[(\widetilde{\mathbf M}_t)_{jk}
           (\widetilde{\mathbf M}_r)_{jk}]
 =\beta^{t-r}\left[\beta(1+\beta)
                    +\sum_{i=1}^{r-1}\beta^{2(r-i+1)}\right].
\]
The remaining sum is the contribution of the older gradients to the
variance at update \(r\). Using
\eqref{eq:predicted-nesterov-variance} at that update and subtracting
the current gradient's contribution gives
\[
 \nu_r=(1+\beta)^2+\sum_{i=1}^{r-1}\beta^{2(r-i+1)}
 \quad\Longrightarrow\quad
 \sum_{i=1}^{r-1}\beta^{2(r-i+1)}=\nu_r-(1+\beta)^2.
\]
Substituting this sum and expanding the two products yields
\[
 \begin{aligned}
 \mathbb E[(\widetilde{\mathbf M}_t)_{jk}
           (\widetilde{\mathbf M}_r)_{jk}]
 &=\beta^{t-r}\left[\beta(1+\beta)+\nu_r-(1+\beta)^2\right]\\
 &=\beta^{t-r}\left[\nu_r+\beta+\beta^2-(1+2\beta+\beta^2)\right]\\
 &=\beta^{t-r}\bigl[\nu_r-(1+\beta)\bigr].
 \end{aligned}
\]

The norm recurrence \eqref{eq:predicted-norm-recursion} requires the
expected Frobenius inner product of the updates,
\[
 \mathbb E\langle\mathbf U_t,\mathbf U_r\rangle_F
 =\mathbb E\sum_{j=1}^{d_{\rm out}}\sum_{k=1}^{d_{\rm in}}
           (\mathbf U_t)_{jk}(\mathbf U_r)_{jk}.
\]
The preceding calculation concerns the momenta before matrix sign.
Because matrix sign is nonlinear, we cannot distribute it over the
gradient sum as in the Adam calculation. We instead construct Gaussian
inputs with the same joint distribution as the standardized momenta
and calculate the expected inner product after applying matrix sign
to those inputs.

The Gaussian assumption is needed here: all entries of the gradients
\(\mathbf G_i\) are assumed independent across rows, columns, and
updates. Equation \eqref{eq:predicted-nesterov-coefficients} expresses
both momenta as linear combinations of this Gaussian collection.
Their entries, considered together, are therefore jointly Gaussian.

Positive rescaling leaves matrix sign unchanged:
\begin{equation}
 c\mathbf A=\mathbf L(c\boldsymbol\Sigma)\mathbf V^\top
 \quad\Longrightarrow\quad
 \operatorname{msign}(c\mathbf A)=\mathbf L\mathbf V^\top
 =\operatorname{msign}(\mathbf A)\qquad(c>0).
 \label{eq:predicted-sign-rescaling}
\end{equation}
We use this property to avoid a separate calculation for every pair
of momentum variances \(\nu_t,\nu_r\). These variances set the
individual scales of the Gaussian inputs, which matrix sign discards.
Dividing each momentum by its entry standard deviation preserves its
matrix-sign output and makes its entry variance one. The dependence
between the two standardized Gaussian inputs is then specified by their
correlation. We can consequently calculate one function of correlation
for each matrix shape and reuse it for every pair of updates.

The correlation between the entries in row \(j\), column \(k\) of
the two momenta is
\begin{equation}
 \rho_{t,r}:=
 \frac{\mathbb E[(\widetilde{\mathbf M}_t)_{jk}
                 (\widetilde{\mathbf M}_r)_{jk}]}
      {\sqrt{\nu_t\nu_r}}
 =\frac{\beta^{t-r}[\nu_r-(1+\beta)]}{\sqrt{\nu_t\nu_r}}
 \quad(t>r).
 \label{eq:predicted-nesterov-correlation}
\end{equation}
The calculation compares the same row and column at updates \(t\)
and \(r\). The identical gradient distributions make the resulting
\(\rho_{t,r}\) the same for every row and column. When \(t=r\),
\(\rho_{t,t}=1\); for \(r>t\), use \(\rho_{t,r}=\rho_{r,t}\).
This is the correlation before matrix sign. We still need the expected
alignment of the updates after that transformation.

To evaluate the transformed inner product at a prescribed correlation
\(\rho\), we need two matrices with centered, unit-variance Gaussian
entries whose corresponding entries have correlation \(\rho\).
Draw independent standard Gaussian matrices \(\mathbf A,\mathbf B\)
of the same dimensions as the momenta, and use
\(\mathbf A\) and \(\rho\mathbf A+\sqrt{1-\rho^2}\mathbf B\).
The contribution \(\rho\mathbf A\) shares the first matrix's random
entries, so its coefficient controls the correlation with that matrix.
The contribution \(\sqrt{1-\rho^2}\mathbf B\) is independent of
\(\mathbf A\) and supplies the remaining variance \(1-\rho^2\).
At row \(j\), column \(k\), the standard Gaussian entries have
\(\mathbb E A_{jk}=\mathbb E B_{jk}=0\) and
\(\mathbb E A_{jk}^2=\mathbb E B_{jk}^2=1\). Independence gives
\(\mathbb E[A_{jk}B_{jk}]=\mathbb E A_{jk}\mathbb E B_{jk}=0\).
Substitution verifies the required means, variances, and covariance:
\begin{equation}
 \begin{aligned}
 \mathbb E[\rho A_{jk}+\sqrt{1-\rho^2}B_{jk}]&=0,\\
 \mathbb E[A_{jk}^2]&=1,\\
 \mathbb E[(\rho A_{jk}+\sqrt{1-\rho^2}B_{jk})^2]
 &=\rho^2\cdot1+2\rho\sqrt{1-\rho^2}\cdot0+(1-\rho^2)\cdot1=1,\\
 \mathbb E[A_{jk}(\rho A_{jk}+\sqrt{1-\rho^2}B_{jk})]
 &=\rho\cdot1+\sqrt{1-\rho^2}\cdot0=\rho.
 \end{aligned}
 \label{eq:predicted-gaussian-pair-moments}
\end{equation}
Because both variances are one, the last line is also the correlation.
For example, \(\mathbf A\) and \(0.6\mathbf A+0.8\mathbf B\) have
unit entry variances because \(0.6^2+0.8^2=1\), and correlation \(0.6\).

To match this construction to updates \(t,r\), set \(\rho=\rho_{t,r}\).
The centered gradients, variance formula
\eqref{eq:predicted-nesterov-variance}, and correlation formula
\eqref{eq:predicted-nesterov-correlation} give
\[
 \begin{aligned}
 \mathbb E\!\left[\frac{(\widetilde{\mathbf M}_t)_{jk}}{\sqrt{\nu_t}}\right]
 &=0,
 &\mathbb E\!\left[\frac{(\widetilde{\mathbf M}_r)_{jk}}{\sqrt{\nu_r}}\right]
 &=0,\\
 \mathbb E\!\left[\left(\frac{(\widetilde{\mathbf M}_t)_{jk}}{\sqrt{\nu_t}}\right)^2\right]
 &=\frac{\nu_t}{\nu_t}=1,
 &\mathbb E\!\left[\left(\frac{(\widetilde{\mathbf M}_r)_{jk}}{\sqrt{\nu_r}}\right)^2\right]
 &=\frac{\nu_r}{\nu_r}=1,\\
 \mathbb E\!\left[\frac{(\widetilde{\mathbf M}_t)_{jk}
                         (\widetilde{\mathbf M}_r)_{jk}}
                        {\sqrt{\nu_t\nu_r}}\right]
 &=\rho_{t,r}.
 \end{aligned}
\]
These are the same means and covariances as in
\eqref{eq:predicted-gaussian-pair-moments} with \(\rho=\rho_{t,r}\).
For different matrix entries, the momenta use independent gradient
entries, and the constructed inputs use independent entries of
\(\mathbf A,\mathbf B\). The covariances between different entries
are therefore zero in both pairs.
Both pairs are jointly Gaussian, since each is a linear transformation
of independent Gaussian entries. They have the same means and
covariances, so the constructed pair and the standardized momenta have
the same joint distribution.

We can now calculate the expected update inner product using the
constructed inputs. Substituting the Muon update contributes a factor
\(s_\mu^2\). Equation \eqref{eq:predicted-sign-rescaling} allows us
to standardize both momenta without changing their matrix signs; their
joint distribution then allows the replacement by the constructed pair.
Dividing by \(C_a(0)\) gives
\begin{equation}
 \begin{aligned}
 \Gamma_{t,r}
 &=\frac{1}{C_a(0)}\mathbb E\langle\mathbf U_t,\mathbf U_r\rangle_F\\
 &=\frac{s_\mu^2}{C_a(0)}\mathbb E\left\langle
   \operatorname{msign}(\widetilde{\mathbf M}_t),
   \operatorname{msign}(\widetilde{\mathbf M}_r)\right\rangle_F\\
 &=\frac{s_\mu^2}{C_a(0)}\mathbb E\left\langle
   \operatorname{msign}\!\left(\frac{\widetilde{\mathbf M}_t}{\sqrt{\nu_t}}\right),
   \operatorname{msign}\!\left(\frac{\widetilde{\mathbf M}_r}{\sqrt{\nu_r}}\right)
   \right\rangle_F\\
 &=\frac{s_\mu^2}{C_a(0)}\mathbb E_{\mathbf A,\mathbf B}\left\langle
   \operatorname{msign}(\mathbf A),
   \operatorname{msign}\!\left(
   \rho_{t,r}\mathbf A+\sqrt{1-\rho_{t,r}^2}\mathbf B\right)
   \right\rangle_F.
 \end{aligned}
 \label{eq:predicted-muon-inner-product}
\end{equation}
Averaging the transformed inner products over independent Gaussian
draws approximates the expectation in this last expression. The
32-pair procedure below performs this numerical calculation.

Evaluating matrix sign by SVD for every pair of training steps would
repeat the same expensive calculation. The expectation depends only
on the input correlation and matrix dimensions, so we compute it at
selected correlations once and reuse the results.
By \eqref{eq:predicted-sign-norm}, each matrix-sign output has squared
norm \(\min(d_{\rm out},d_{\rm in})\). Dividing their inner product
by this number removes the common norm factor, giving a normalized
alignment. Following \citet[Eq.~19]{wen2026hyperball}, define its
expectation as
\begin{equation}
 \kappa_{d_{\rm out},d_{\rm in}}(\rho)
 :=\frac{1}{\min(d_{\rm out},d_{\rm in})}
 \mathbb E_{\mathbf A,\mathbf B}\left\langle
 \operatorname{msign}(\mathbf A),
 \operatorname{msign}(\rho\mathbf A+\sqrt{1-\rho^2}\mathbf B)
 \right\rangle_F.
 \label{eq:predicted-sign-correlation}
\end{equation}
The expectation averages over \(\mathbf A,\mathbf B\), leaving a
function of \(\rho\) and the dimensions alone.
When \(\rho=1\), the two inputs are identical, so
\[
 \kappa_{d_{\rm out},d_{\rm in}}(1)
 =\frac{\mathbb E\|\operatorname{msign}(\mathbf A)\|_F^2}
       {\min(d_{\rm out},d_{\rm in})}=1.
\]
When \(\rho=0\), the inputs are the independent matrices
\(\mathbf A,\mathbf B\). Gaussian symmetry and the oddness of matrix
sign give
\[
 \mathbb E\operatorname{msign}(\mathbf A)
 =\mathbb E\operatorname{msign}(-\mathbf A)
 =-\mathbb E\operatorname{msign}(\mathbf A),
 \qquad \mathbb E\operatorname{msign}(\mathbf A)=0.
\]
The same holds for \(\mathbf B\). Independence therefore gives
\[
 \kappa_{d_{\rm out},d_{\rm in}}(0)
 =\frac{\left\langle\mathbb E\operatorname{msign}(\mathbf A),
                      \mathbb E\operatorname{msign}(\mathbf B)\right\rangle_F}
       {\min(d_{\rm out},d_{\rm in})}=0.
\]
To insert this expected normalized alignment into
\eqref{eq:predicted-norm-recursion}, combine
\eqref{eq:predicted-muon-inner-product} with
\eqref{eq:predicted-sign-correlation} and use
\(s_\mu^2\min(d_{\rm out},d_{\rm in})=d_{\rm out}\):
\begin{equation}
 \begin{aligned}
 \Gamma_{t,r}
 &=\frac{s_\mu^2\min(d_{\rm out},d_{\rm in})}{C_a(0)}
      \kappa_{d_{\rm out},d_{\rm in}}(\rho_{t,r})\\
 &=\frac{d_{\rm out}}{d_{\rm out}d_{\rm in}}
      \kappa_{d_{\rm out},d_{\rm in}}(\rho_{t,r})
  =\frac{\kappa_{d_{\rm out},d_{\rm in}}(\rho_{t,r})}{d_{\rm in}}
      \quad(t>r),\\
 P_t&=\frac{1}{d_{\rm in}},\qquad \Gamma_{t,t}=P_t.
 \end{aligned}
 \label{eq:predicted-muon-moments}
\end{equation}

For each matrix shape, we draw 32 independent pairs
\((\mathbf A,\mathbf B)\). At each selected value of \(\rho\), we apply
matrix sign to \(\mathbf A\) and
\(\rho\mathbf A+\sqrt{1-\rho^2}\mathbf B\), divide their inner product
by \(\min(d_{\rm out},d_{\rm in})\), and average the 32 results.
The same 32 pairs are used at every correlation value.
We store these averages together with their input correlations.
When \(\rho_{t,r}\) lies between two tabulated inputs, we interpolate
linearly between the tabulated estimates of \(\kappa\).
The endpoint values at zero and one are set exactly.
No training measurements enter this calculation.

The norm recurrence retains correlations up to a lag of 512 updates.
Equation \eqref{eq:predicted-nesterov-correlation} gives
\(0\leq\rho_{t,r}\leq\beta^{t-r}\), and
\citet[Eq.~30]{wen2026hyperball} show \(0\leq\kappa(\rho)\leq\rho\).
At \(\beta=0.95\), this bound is below \(4\times10^{-12}\) at lag 512.

\subsubsection{Rotational profiles and bounds}
\paragraph{Predictions.}
To compare our bound with validation loss, we need the rotational
coefficients produced by each learning-rate schedule. We predict these
coefficients from \(P_t\), \(\Gamma_{t,r}\), and the norm recurrence
above.

We want the bound prediction to be free of training measurements and
empirically tuned peak learning rates, i.e.\ to be fully predictive.
However, the peak learning rate affects the predicted norm trajectory
and hence the rotational profile used to evaluate the bound. After
optimizing \(M\) (\Cref{prop:optimal-mass}), the bound comparison depends
only on the normalized rotational profile \(\omega_t\). We therefore
ask how this profile can be predicted from the learning-rate shape alone.
The approximation derived below removes this scale dependence when
accumulated norm growth dominates initialization. As in
\(s_t=M\omega_t\), we separate the full learning-rate schedule,
including warmup, into its sum and normalized shape:
\[
 M_{\rm LR}:=\sum_{t=1}^{20000}\eta_t,\qquad
 \omega_t^{\rm LR}:=\frac{\eta_t}{M_{\rm LR}},\qquad
 \eta_t=M_{\rm LR}\omega_t^{\rm LR},\qquad
 \sum_{t=1}^{20000}\omega_t^{\rm LR}=1.
\]
For a fixed learning-rate shape, the nominal peak selected by training
is proportional to \(M_{\rm LR}\).

Approximating the optimizer update as tangent and replacing the update
and parameter norms by their RMS values gives
\citep[Apps.~B.2 and~E.2.1]{anonymous2026how}
\[
 \tan\Delta\phi_t\approx
 \eta_t\frac{\sqrt{C_a(0)P_t}}{R_t}.
\]
The numerator is the update displacement; the denominator is the
parameter norm. With zero decay and
\(\mathbb E\langle\mathbf X_1,\mathbf U_r\rangle_F=0\), unrolling the
update and substituting the learning-rate decomposition separates the
initial norm from accumulated growth:
\begin{equation}
 \begin{aligned}
 R_t^2
 &=R_1^2+
 \mathbb E\left\|\sum_{r=1}^{t-1}\eta_r\mathbf U_r\right\|_F^2\\
 &=R_1^2+M_{\rm LR}^2
 \mathbb E\left\|\sum_{r=1}^{t-1}\omega_r^{\rm LR}\mathbf U_r\right\|_F^2.
 \end{aligned}
 \label{eq:predicted-norm-expansion}
\end{equation}
Here \(R_1^2=\mathbb E\|\mathbf X_1\|_F^2\) is the expected squared
parameter norm at initialization. To see how the learning-rate scale
affects rotation, substitute the decomposition into the predicted
coefficient and divide its numerator and denominator by \(M_{\rm LR}>0\):
\[
 \frac{\eta_t\sqrt{C_a(0)P_t}}{R_t}
 =\frac{\omega_t^{\rm LR}\sqrt{C_a(0)P_t}}
 {\sqrt{R_1^2/M_{\rm LR}^2+
 \mathbb E\left\|
 \sum_{r=1}^{t-1}\omega_r^{\rm LR}\mathbf U_r
 \right\|_F^2}}.
\]
For a fixed learning-rate shape and fixed modeled update statistics,
the second term in the denominator is independent of \(M_{\rm LR}\).
The initialization term retains the scale dependence because changing
the learning rates leaves the initial weights unchanged. Dividing the
rotational coefficients by their sum therefore does not generally
remove this dependence.

To predict the rotational shape without the tuned nominal peak, consider
a step \(t\) at which accumulated updates dominate initialization:
\[
 R_1^2\ll R_t^2-R_1^2
 =M_{\rm LR}^2
 \mathbb E\left\|
 \sum_{r=1}^{t-1}\omega_r^{\rm LR}\mathbf U_r
 \right\|_F^2.
\]
The term \(R_t^2-R_1^2\) is the expected squared accumulated
displacement from all earlier updates, including warmup. Under this
condition, the predicted radius is close to the square root of that
contribution. Dividing the condition by \(M_{\rm LR}^2>0\) shows that
\(R_1^2/M_{\rm LR}^2\) is small relative to the second term in the
denominator above. Dropping it gives
\[
 \frac{\eta_t\sqrt{C_a(0)P_t}}{R_t}
 \approx
 \frac{\omega_t^{\rm LR}\sqrt{C_a(0)P_t}}
 {\sqrt{\mathbb E\left\|
 \sum_{r=1}^{t-1}\omega_r^{\rm LR}\mathbf U_r
 \right\|_F^2}}.
\]
The overall learning-rate scale no longer appears: scaling the rates
scales both the current displacement and the norm of the accumulated
displacement.

The norm-growth identity \eqref{eq:predicted-norm-expansion} also gives
\[
 \mathbb E\left\|
 \sum_{r=1}^{t-1}\omega_r^{\rm LR}\mathbf U_r
 \right\|_F^2
 =\frac{R_t^2-R_1^2}{M_{\rm LR}^2}.
\]
Substituting this identity into the approximate coefficient and using
\(M_{\rm LR}>0\) and \(\eta_t=M_{\rm LR}\omega_t^{\rm LR}\) gives
\begin{equation}
 \begin{aligned}
 \widehat s_t
 &:=\frac{\omega_t^{\rm LR}\sqrt{C_a(0)P_t}}
 {\sqrt{\mathbb E\left\|
 \sum_{r=1}^{t-1}\omega_r^{\rm LR}\mathbf U_r
 \right\|_F^2}}\\
 &=\frac{\omega_t^{\rm LR}\sqrt{C_a(0)P_t}}
 {\sqrt{(R_t^2-R_1^2)/M_{\rm LR}^2}}\\
 &=\frac{M_{\rm LR}\omega_t^{\rm LR}\sqrt{C_a(0)P_t}}
 {\sqrt{R_t^2-R_1^2}}\\
 &=\eta_t\sqrt{\frac{C_a(0)P_t}{R_t^2-R_1^2}}.
 \end{aligned}
 \label{eq:predicted-rotation}
\end{equation}
The finite-history calculation of \(R_t\) retains the initial norm and
the full warmup history; subtracting \(R_1^2\) isolates the accumulated
update contribution for this approximation. The comparison starts
after warmup, where the growth term is positive. The approximation is
less accurate near warmup when the initial norm remains appreciable.
For Hyperball,
\(s_t^{\rm HB}=\eta_t\), so its normalized rotational profile follows
the nominal learning-rate shape directly.

At each post-warmup update, we take the median predicted coefficient
across the 60 attention and MLP matrices. Each of the twelve layers
contains five matrix types:
fused QKV (\(2304\times768\)), attention output (\(768\times768\)),
MLP gate and up (each \(2048\times768\)), and MLP down
(\(768\times2048\)). The model assigns the same dimensions and
initialization scale to each type across layers, so the median over the
five predictions equals the median over all 60 matrices. Below,
\(\widehat s_t\) denotes this median. Dividing by the sum over the
\(N=18{,}000\) post-warmup updates gives
\[
 \omega_t=\frac{\widehat s_t}{\sum_{i=2001}^{20000}\widehat s_i},
 \qquad 2001\leq t\leq20000,\qquad
 \sum_{t=2001}^{20000}\omega_t=1.
\]

The norm plot checks the norm growth used to predict rotation. For each
matrix type, the predicted norm-growth curve is
\[
 \frac{\sqrt{R_{t+1}^2-R_1^2}}{R_1}.
\]
We take the median prediction across matrix types and divide the
resulting curve by its final value. Each predicted curve then ends at
one, so the comparison shows how norms change over training.

The bound curves are calculated from the predicted rotational profiles.
For each full profile, we choose \(M\) using \eqref{eq:optimal-mass}
and retain it when evaluating every prefix. As in
\Cref{app:schedule-comparison-protocol}, division by the common scale
\(DG\sqrt{1+\thetacap}\) gives the normalized final bound
\(\sqrt{C_N(\omega)}\). We evaluate prefixes every 250 updates and at
the first post-warmup update. The bound curves use the fraction of all
20,000 updates and begin at the end of warmup.

\paragraph{Measurements.}
We use the three evaluation seeds at the selected peak rates described
in \Cref{app:hyperball-protocol}. For each seed, we interpolate the
logged median turning angles across the 60 matrices, recorded every
50 updates, to every post-warmup update. We convert these measured
angles into rotational coefficients and normalize:
\[
 s_t^{\rm measured}=\tan\Delta\phi_t,\qquad
 \omega_t^{\rm measured}
 =\frac{s_t^{\rm measured}}{\sum_{i=2001}^{20000}s_i^{\rm measured}}.
\]
For norms, we divide each matrix's measured norm by its initial norm,
take the median across matrices, interpolate it to the same updates,
and divide the measured curve by its final value. Each measured norm
curve therefore ends at one. We perform both normalizations separately
for each seed before taking the mean and sample standard deviation
across the three seeds. We use these measured profiles to check the
predictions.

\begin{figure}[!htbp]
 \centering
 \includegraphics[width=.95\linewidth]{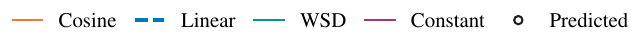}
 \includegraphics[width=.9\linewidth]{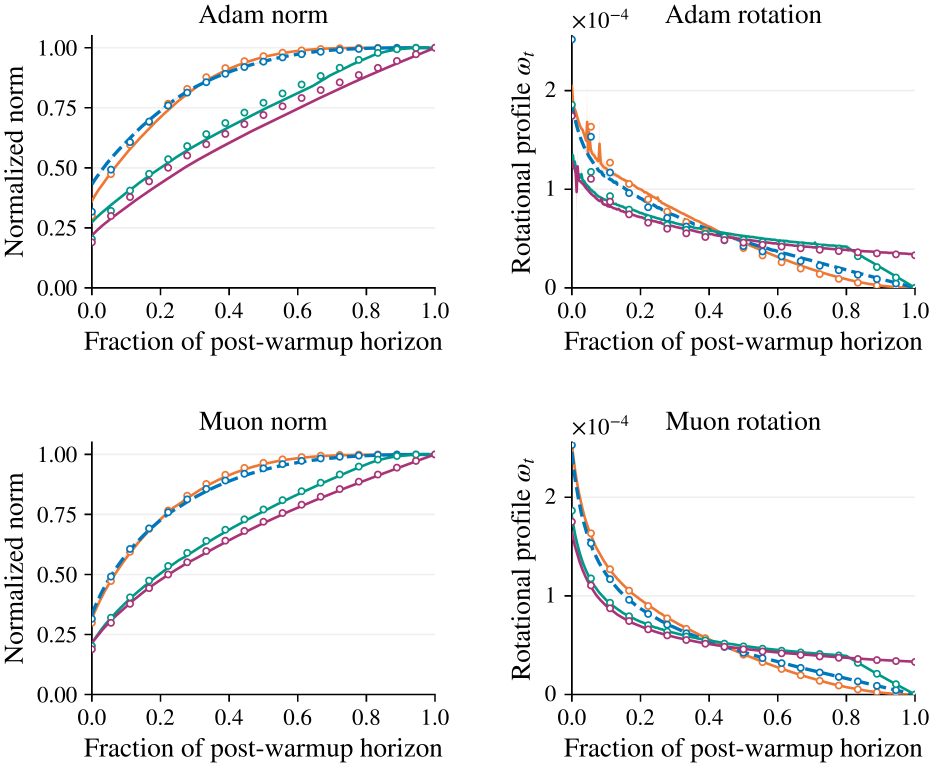}
 \caption{Predicted and measured profiles for Adam (top) and Muon
 (bottom). Open circles show model predictions; curves and bands show
 the measured mean and one sample standard deviation across three
 evaluation seeds. Left: predicted norm growth and measured norm ratios
 relative to initialization, each scaled to one at its endpoint.
 Right: unit-sum rotational profiles over updates 2,001--20,000.}
 \label{fig:unconstrained-rotation-shapes}
\end{figure}

\paragraph{Results.}
\Cref{fig:unconstrained-rotation-shapes} compares predicted and measured
norm growth and rotational profiles. The predicted norm curves follow
the changes over training in the measured curves. During WSD's
constant-rate phase, the measured norms keep growing and rotation
decreases before the nominal learning rate begins to decay. The
predicted rotational profile shows the same decrease.

\begin{figure}[!htbp]
 \centering
 \includegraphics[width=.95\linewidth]{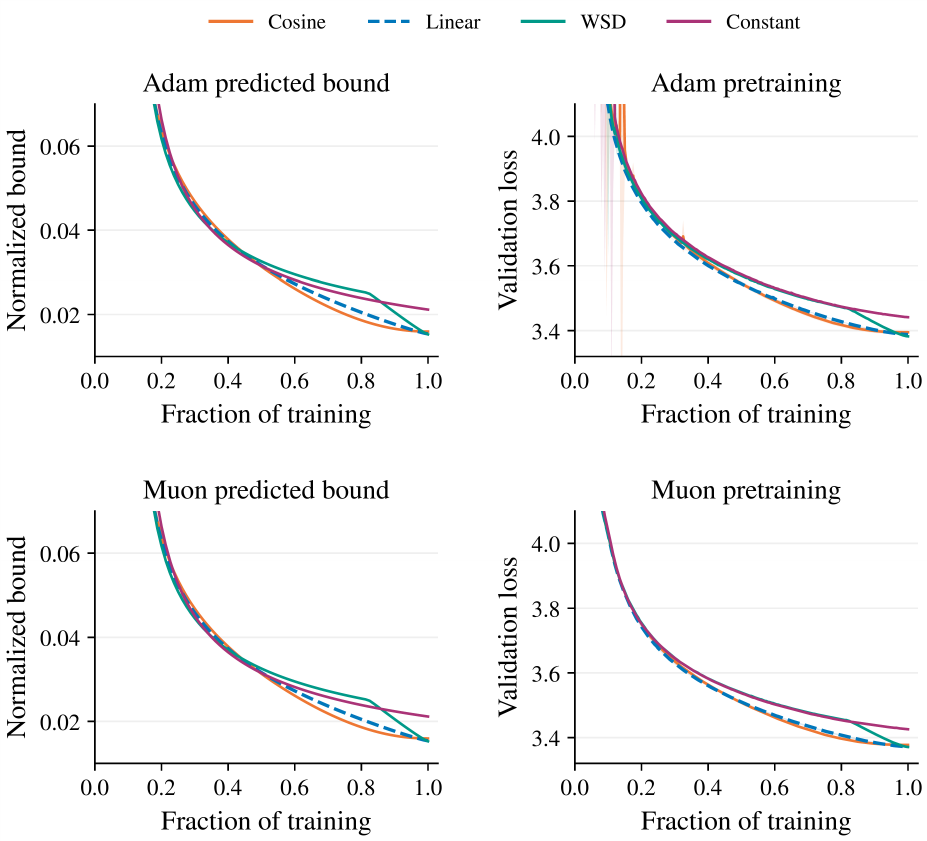}
 \caption{Predicted bounds (left) and validation loss (right)
 for Adam (top) and Muon (bottom). Bounds use the predicted rotational
 profiles, begin after warmup, and retain each profile's full-horizon
 optimal coefficient sum. Loss curves and bands show the measured mean
 and one sample standard deviation across three evaluation seeds.
 Both columns use the fraction of all 20,000 updates.}
 \label{fig:unconstrained-bound-loss}
\end{figure}

\Cref{fig:unconstrained-bound-loss} compares the predicted bounds with
measured validation loss. WSD's late improvement places it below cosine
in both the predicted bound and measured loss for Adam and Muon.

\Cref{fig:bound-contrast} compares the bound evaluated on nominal learning
rates, as in \citet{schaipp2025surprising}, with our bound evaluated on
predicted rotational profiles. Both are shown beside measured final validation
loss, relative to cosine. Under Hyperball, the two predictions coincide
and place WSD above cosine, matching its higher validation loss.
Without Hyperball, norm growth cools rotation during WSD's stable phase.
Accounting for this growth places WSD below cosine for both Adam and
Muon, recovering the observed loss ordering. The bound based on nominal
learning rates predicts the wrong ordering.

\FloatBarrier

\subsubsection{Weight decay}
\label[appendix]{app:weight-decay}
Weight decay limits norm growth and can therefore change the rotation
induced by a learning-rate schedule.

The finite-history norm recurrence \eqref{eq:predicted-norm-recursion}
accounts for decay through \(\delta_t=1-\eta_t\lambda_t\) and
the accumulated contractions \(W_{i,t}\). Factoring the update and
then normalizing gives
\[
 \begin{aligned}
 \frac{\mathbf X_{t+1}}{\|\mathbf X_{t+1}\|_F}
 &=\frac{\delta_t\left[
 \mathbf X_t-\dfrac{\eta_t}{\delta_t}\mathbf U_t\right]}
 {\left\|\delta_t\left[
 \mathbf X_t-\dfrac{\eta_t}{\delta_t}\mathbf U_t\right]\right\|_F}\\
 &=\frac{\mathbf X_t-\dfrac{\eta_t}{\delta_t}\mathbf U_t}
 {\left\|\mathbf X_t-\dfrac{\eta_t}{\delta_t}\mathbf U_t\right\|_F}.
 \end{aligned}
\]
The positive factor \(\delta_t\) cancels: the direction is that of a
step with effective displacement \((\eta_t/\delta_t)\mathbf U_t\).
If \(\langle\mathbf X_t,\mathbf U_t\rangle_F=0\), the two components
are perpendicular, so their lengths give
\[
 \tan\Delta\phi_t
 =\frac{\eta_t\|\mathbf U_t\|_F}
 {\delta_t\|\mathbf X_t\|_F}.
\]
Approximating the update as tangent and replacing the two norms by
their RMS values gives the proxy
\begin{equation}
 \widehat s_t\approx
 \frac{\eta_t\sqrt{C_a(0)P_t}}{\delta_t R_t}.
 \label{eq:weight-decay-rotation}
\end{equation}

Unlike the zero-decay growth approximation above, this prediction
depends on the learning-rate amplitude. Rescaling the rates also changes
\(\delta_t=1-\eta_t\lambda_t\) and hence the products \(W_{i,t}\)
in the norm recurrence. The prediction therefore requires the full \(R_t\),
retaining initialization and all warmup updates.

\FloatBarrier

\subsection{\texorpdfstring{\textcolor{black}{Larger-Scale Pretraining}}{Larger-Scale Pretraining}}
\begin{figure}[!htbp]
 \centering
 \includegraphics[width=.9\linewidth]{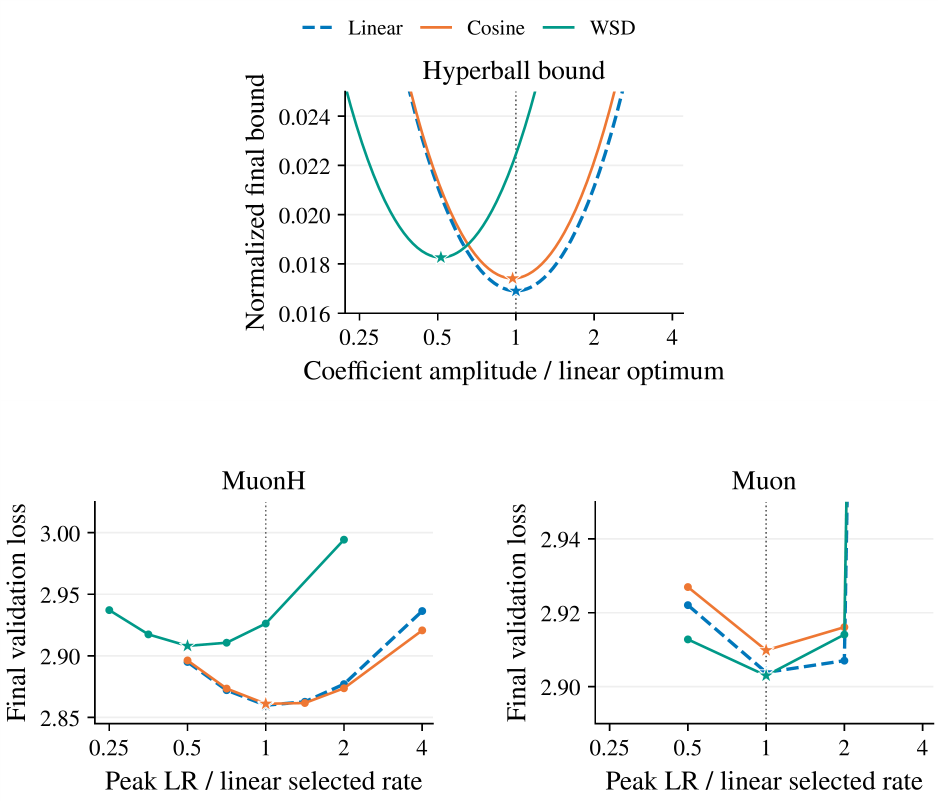}
 \caption{\textcolor{black}{Learning-rate sensitivity at 720M. Top: the
 Hyperball bound over the 14,000 post-warmup updates. Bottom: final
 validation loss for MuonH and Muon. Horizontal axes are relative to
 linear's analytical optimum (top) or selected peak rate (bottom).
 Stars mark the minima.}}
 \label{fig:720m-sensitivity}
\end{figure}

\label[appendix]{app:hyperball-720m}
{\color{black}%
To verify how our predictions hold up in a larger-scale setting, we
repeat the linear, cosine, and WSD comparison with MuonH and Muon on
FineWeb on a 720M configuration with 12 layers, hidden width 2048, MLP width 5632,
and 16 attention heads. We train for 16,000 updates with sequence
length 512 and global batch size 1,984, corresponding to approximately
16.25 billion tokens. Warmup lasts 2,000 updates, and WSD uses a linear
cooldown over the final 2,800 updates. We use gradient clipping at 0.1
and auxiliary Adam momentum $\beta_1=0.95$, with zero weight decay.
We follow the tuning and three-seed evaluation procedure of
\Cref{app:hyperball-protocol}.

\Cref{fig:720m-sensitivity} shows the tested learning rates and selected
minima. The larger runs reproduce the comparison at 124M: under
Hyperball, linear and cosine reach similar final losses, while WSD
ends at a higher loss. Without Hyperball, WSD ends at a lower loss than cosine
and a slightly lower loss than linear after cooldown.
}%

\begin{figure}[!htbp]
 \centering
 \includegraphics[width=\linewidth]{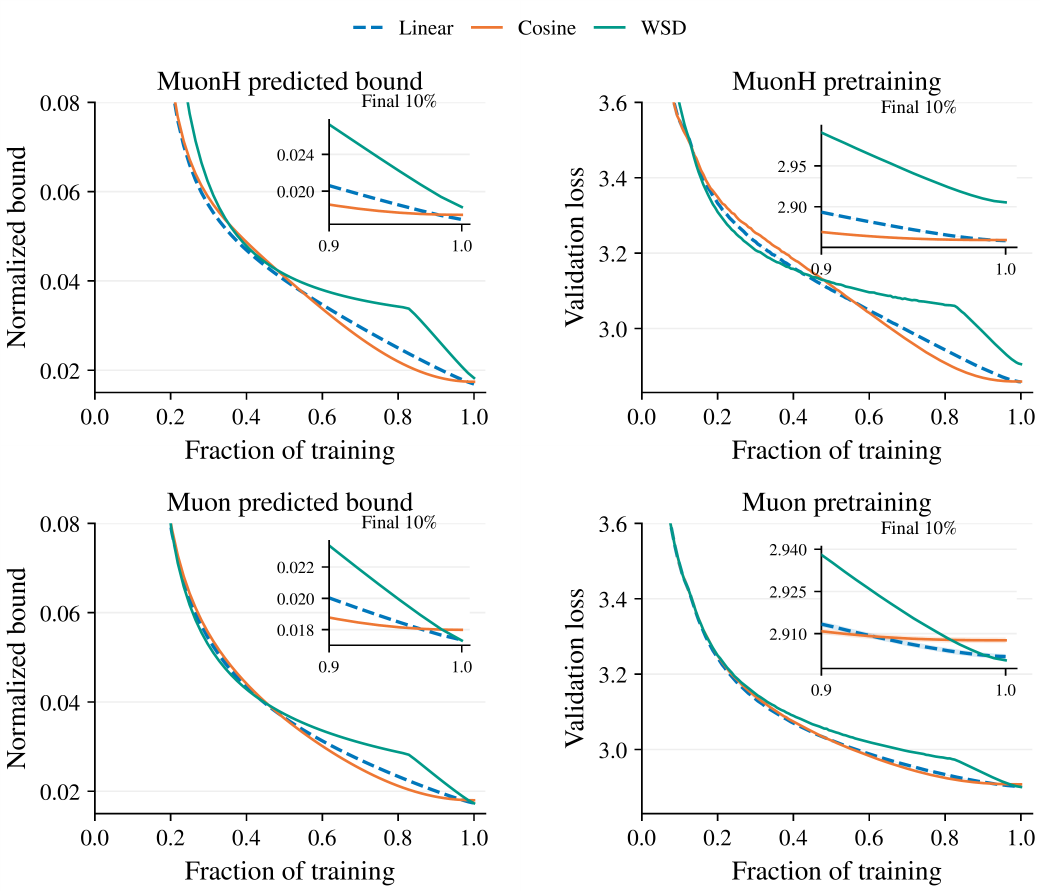}
 \caption{\textcolor{black}{Predicted bounds (left) and  validation loss
 (right) at 720M, with MuonH above and Muon below. Each predicted
 rotational profile over updates 2,001--16,000 retains its full-horizon
 optimal coefficient sum for every prefix. Loss curves and bands show
 the mean and one sample standard deviation across three evaluation
 seeds.
 Zooms show the final 10\% of training.}}
 \label{fig:720m-bound-loss}
\end{figure}

\begin{figure}[!htbp]
 \centering
 \includegraphics[width=\linewidth]{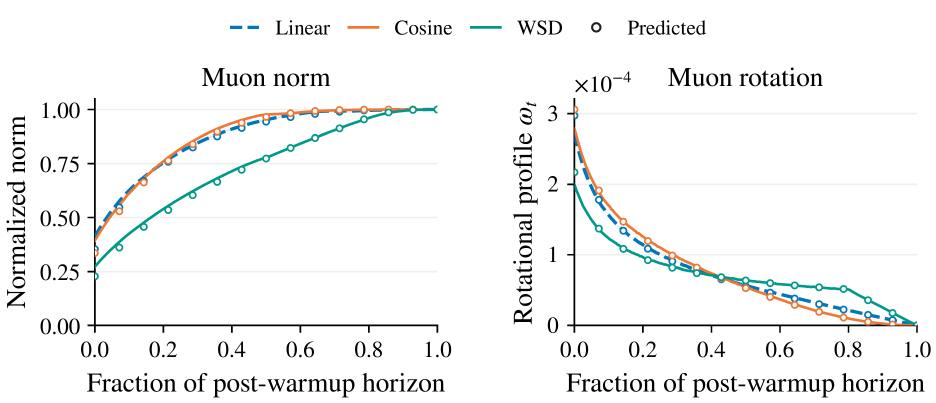}
 \caption{\textcolor{black}{Predicted and measured Muon profiles at 720M.
 Left: norm growth normalized by its endpoint. Right: unit-sum
 rotational profiles over updates 2,001--16,000. Open circles show
 predictions; curves and bands show the measured mean and one sample
 standard deviation across three evaluation seeds.}}
 \label{fig:720m-profiles}
\end{figure}

{\color{black}%
We use the prediction method of \Cref{app:rotation-diagnostic} with
the larger model's matrix dimensions and training horizon.
\Cref{fig:720m-bound-loss} compares the resulting bounds with validation
loss, and \Cref{fig:720m-profiles} compares the predicted norms and
rotational profiles with measurements.
The bounds reproduce the WSD--cosine reversal between the two settings.
With Muon, WSD and linear have close final bounds, both below cosine.
The predicted and measured norm growth cools rotation during WSD's
constant-rate phase, before its explicit cooldown begins.
}%

\FloatBarrier

\subsection{From Rotational Profiles to Learning Rates}
\label[appendix]{app:norm-aware-schedule}
The norm model of \Cref{app:rotation-diagnostic} also lets us construct
learning-rate schedules from a prescribed rotational profile. We use
it to target the bound-optimal profile for Adam and Muon without
Hyperball or weight decay. As a matrix's norm grows, a larger learning
rate is needed to produce the same turn at a given update norm. The
constructed learning rates can therefore keep rising after warmup
while the target rotation decreases.

\paragraph{Constructing the learning rates.}
The target is \(M\omega_{t-2000}^{\star}\), where
\(\omega_{1:N}^{\star}\) is the exact optimal profile of
\Cref{cor:optimal-allocation} over \(N=18{,}000\) post-warmup updates.
Both this shape and its amplitude \(M\) are common across matrices.
Using the tangent-update and RMS norm approximation of
\Cref{app:rotation-diagnostic}, we match each matrix's predicted
coefficient \(\eta_t\sqrt{C_a(0)P_t}/R_t\) to the target. This gives
\begin{equation}
 \eta_t=\frac{M\omega_{t-2000}^{\star}R_t}{\sqrt{C_a(0)P_t}},
 \qquad t=2001,\ldots,20000.
 \label{eq:norm-aware-learning-rate}
\end{equation}
The norm \(R_t\) and update power \(C_a(0)P_t\) here belong to the
matrix being updated. A larger norm requires a larger learning rate
to achieve the same coefficient; a larger update norm requires a smaller
rate.

To obtain actual learning rates, we retain initialization and warmup
in \eqref{eq:predicted-norm-recursion}; the growth-dominated comparison
in \Cref{app:rotation-diagnostic} neglected the initial contribution.

For pretraining, we tune one overall scale per optimizer and schedule
family by final validation loss. As for the standard schedules, we
express it through the \emph{overall peak}: the largest learning rate
across all matrices and updates. For each candidate peak, we find
\(M\) numerically and construct the matrix schedules, so their individual
peaks follow from the model.

Each candidate peak requires reconstructing the rates and norms
together. The initial norm stays fixed, whereas the accumulated growth
in \eqref{eq:predicted-norm-expansion} depends on the preceding rates.
Rescaling a completed learning-rate curve therefore changes the balance
between initialization and growth, and generally changes its rotational
profile. Reconstruction allows the learning-rate curves to change in
shape as well as height while retaining the target rotational profile.

To start this calculation, each matrix's warmup must meet its first
prescribed post-warmup coefficient \(M\omega_1^\star\). For a fixed
\(M\), the warmup ramp has relative heights
\[
 \frac{\eta_r}{\eta_{2001}}
 =0.01+0.99\frac{r-1}{2000},
 \qquad 1\leq r\leq2000.
\]
The initial expected squared norm is the number of entries times their
initialization variance. Equation \eqref{eq:predicted-norm-expansion}
then gives the norm at the end of warmup:
\[
 R_{2001}^2=R_1^2+\eta_{2001}^2
 \mathbb E\left\|
 \sum_{r=1}^{2000}\frac{\eta_r}{\eta_{2001}}\mathbf U_r
 \right\|_F^2.
\]
Since the rate ratios are fixed, the expectation depends only on the
warmup shape and modeled update statistics. Substituting this norm into
the matching condition
\(\eta_{2001}^2C_a(0)P_{2001}
 =(M\omega_1^\star)^2R_{2001}^2\) gives the positive warmup height
\begin{equation}
 \eta_{2001}
 =\frac{M\omega_1^\star R_1}
 {\sqrt{C_a(0)P_{2001}
 -(M\omega_1^\star)^2
 \mathbb E\left\|
 \sum_{r=1}^{2000}\frac{\eta_r}{\eta_{2001}}\mathbf U_r
 \right\|_F^2}}.
 \label{eq:norm-aware-warmup-height}
\end{equation}
For a candidate peak, we first try an amplitude \(M\) and construct all
matrix schedules. Equation \eqref{eq:norm-aware-warmup-height} sets each
matrix's warmup ramp, provided its denominator is positive. From the
norm reached during warmup, we calculate the first post-warmup rate
with \eqref{eq:norm-aware-learning-rate}, then use the norm recurrence
\eqref{eq:predicted-norm-recursion} to obtain the next norm. Repeating
these two steps gives all remaining rates. We then take the largest
rate across matrices and updates. Bisection adjusts \(M\) and repeats
this calculation until that maximum equals the candidate peak.

In this model, the update power \(P_t\) and correlations
\(\Gamma_{t,r}\) depend on optimizer settings and matrix dimensions,
so they remain the same when \(M\) changes. We calculate them once,
using the analytic Adam expressions \eqref{eq:predicted-adam-moments}
and independent Gaussian matrices for Muon
\eqref{eq:predicted-sign-correlation}, and reuse them for every proposed
\(M\) and candidate peak. The rates and predicted norms are recalculated
using these fixed statistics.

The twelve blocks repeat the same five matrix types and initialization
scales. One predicted norm history for each type therefore supplies the
schedules for all 60 matrices. The calculation uses matrix dimensions,
initialization, and
optimizer settings, without measured gradients, norms, or losses from
training. We construct all candidate schedules before training and
apply the selected schedules unchanged during the runs.

\paragraph{Matrix-wise and shared learning rates.}
In the Adam model, \(P_t\) and \(\Gamma_{t,r}\) are the same per entry
across matrix types.
Both the squared norm and the update power contain the factor
\(d_{\rm out}d_{\rm in}\), which cancels in
\eqref{eq:norm-aware-learning-rate}. Doubling the initial entry standard
deviation and all earlier rates doubles the modeled norm, so the
inversion also gives twice the rate at the next update. This scaling
holds from warmup onward, making the schedules proportional to the
initialization standard deviation.
QKV, MLP gate, and MLP up use entry standard deviation \(0.02\);
attention output and MLP down use
\(0.02/\sqrt{2\cdot12}=0.02/\sqrt{24}\). Their rates are therefore
\(1/\sqrt{24}\) times the other three rates throughout training.

\begin{figure}[!p]
 \centering
 \includegraphics[width=\linewidth]{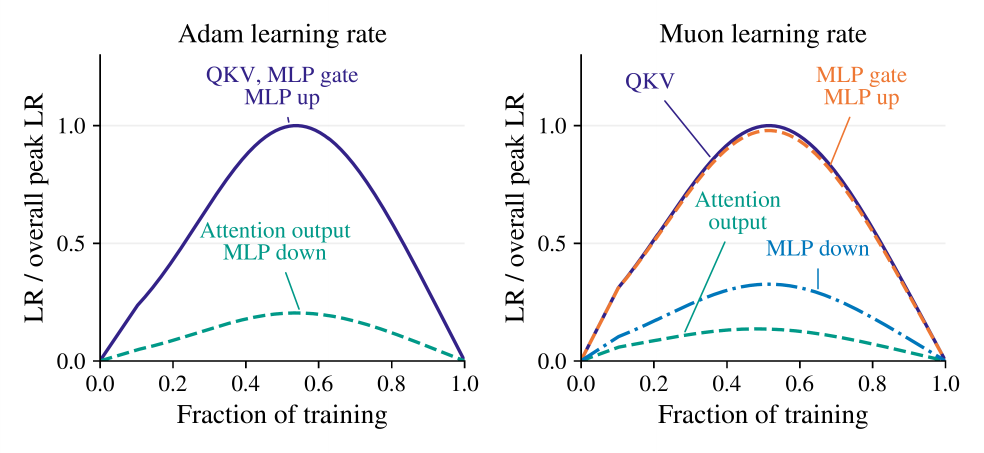}
 \caption{Constructed learning-rate schedules at the selected overall
 peaks for Adam (left) and Muon (right). Labels identify the matrix types
 using each curve across all twelve blocks; coincident schedules are
 drawn once. All rates in each panel are divided by its overall peak,
 preserving their relative heights.}
 \label{fig:matrix-wise-learning-rates}
\end{figure}

Muon's update power and correlations depend on matrix dimensions
through \eqref{eq:predicted-muon-moments}, so its learning-rate curves
can differ in shape as well as height
(\Cref{fig:matrix-wise-learning-rates}).

To test the role of these matrix-dependent rates, we also construct a
shared-rate ablation. One rate generally cannot match the target for
every matrix, so we match the median predicted coefficient:
\begin{equation}
 \eta_t=
 \frac{M\omega_{t-2000}^{\star}}
 {\operatorname{median}\!\left[\sqrt{C_a(0)P_t}/R_t\right]}.
 \label{eq:shared-norm-aware-learning-rate}
\end{equation}
The median is over the 60 attention and MLP matrices. A trial shared
warmup height determines all five norms at update 2,001 and hence
\(M\) through this equation. We then calculate the shared rates and
matrix norms successively, adjusting the warmup height by bisection
to obtain the requested peak.

\begin{figure}[!htbp]
 \centering
 \includegraphics[width=.95\linewidth]{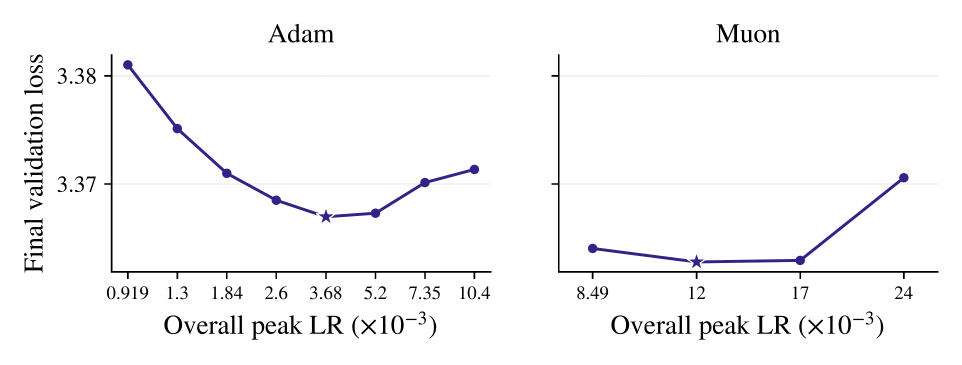}
 \caption{Learning-rate sensitivity of the matrix-wise schedules for Adam
 (left) and Muon (right). Points show final validation loss for completed
 runs on the tuning seed; stars mark the best tested overall peaks.}
 \label{fig:matrix-wise-sensitivity}
\end{figure}

\paragraph{Pretraining comparison.}
We use the training recipe of \Cref{app:hyperball-protocol}, with one
tuning-seed run per candidate overall peak for each family.
The matrix-wise grids have successive peaks separated by \(\sqrt2\):
eight values from \(0.0013/\sqrt2\) to \(0.0104\) for Adam and four from
\(0.012/\sqrt2\) to \(0.024\) for Muon
(\Cref{fig:matrix-wise-sensitivity}).
\Cref{tab:hyperball-winners} reports the selected rates. For the shared
Adam schedule, we reduced the peak from \(0.0016\) to \(0.0013\) to
improve training stability.

\begin{figure}[!htbp]
 \centering
 \includegraphics[width=\linewidth]{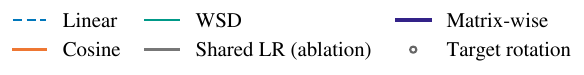}
 \includegraphics[width=\linewidth]{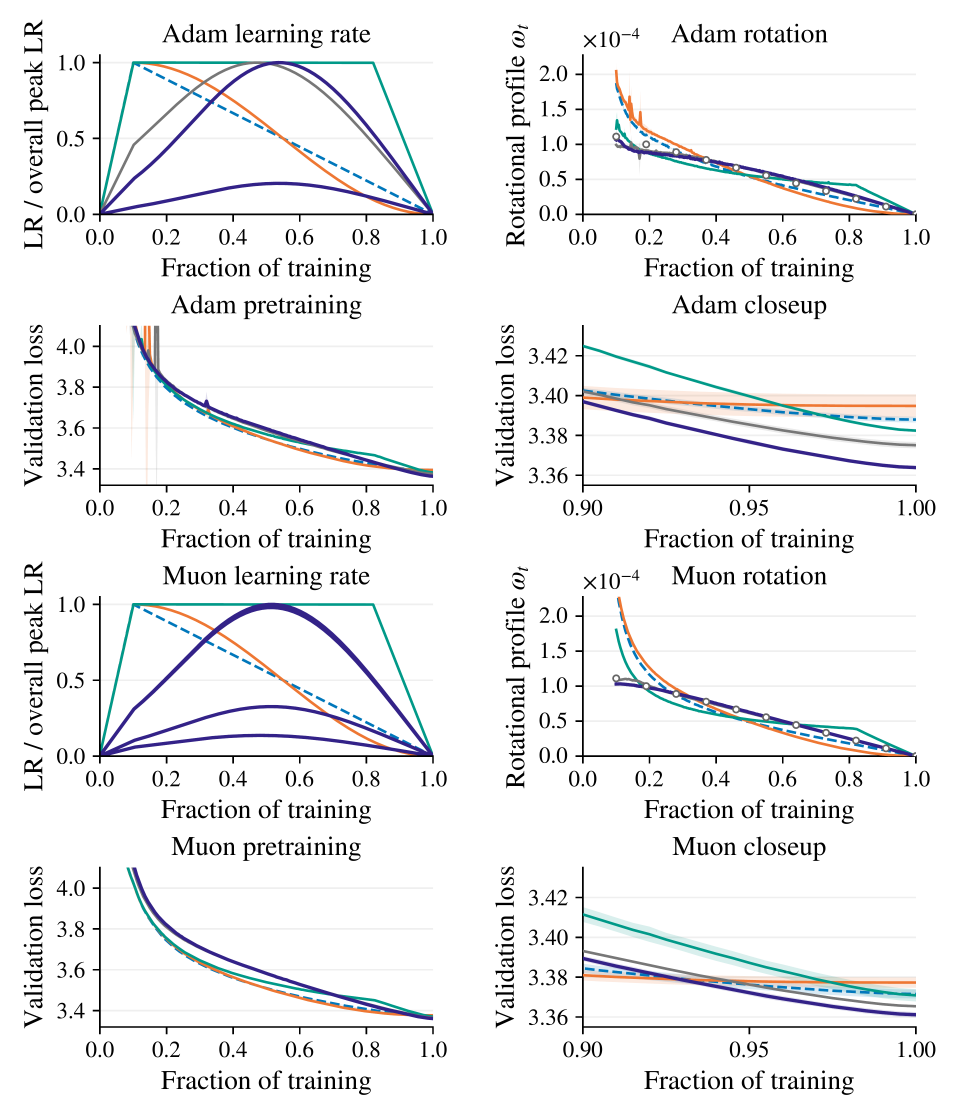}
 \caption{Matrix-wise scheduling and its shared-LR ablation for Adam
 (top block) and Muon (bottom block), compared with independently tuned
 linear, cosine, and WSD. Within each \(2\times2\) block, top left:
 precomputed learning rates divided by each schedule's overall peak,
 using one common peak for all purple matrix-class curves; coincident
 curves are drawn once. Top right: measured unit-sum rotational profiles
 derived from the median turning angle across the 60 matrices; open
 circles show the theoretical target. Bottom left: measured
 validation loss. Bottom right: the final 10\% of the same loss curves.
 Measured rotation and loss curves show means across three evaluation
 seeds; bands show one sample standard deviation. Final-loss comparisons
 use the separate final validation measurements.}
 \label{fig:norm-aware-schedule}
\end{figure}

We hold the selected schedules fixed for three fresh evaluation seeds
and convert measured turning angles to normalized rotational profiles
using the procedure of \Cref{app:rotation-diagnostic}.
\Cref{fig:norm-aware-schedule} compares both constructions with linear,
cosine, and WSD at their selected rates. The measured median rotational
profiles follow the prescribed decay. Matrix-wise scheduling reaches
lower final validation loss than the shared-LR ablation and each of the
three standard schedules in every seed.
The prolonged learning-rate increase is consistent with the findings
of \citet{topollai2026warmup}: without
weight decay, longer warmup permits larger competitive peak rates and
improves final loss at those rates for both Adam and Muon.

\FloatBarrier